\documentclass[11pt]{article}

\usepackage[utf8]{inputenc}
\usepackage{amsmath}
\usepackage{amsfonts}
\usepackage{mathtools}
\usepackage{graphicx}
\usepackage{color}
\usepackage[hidelinks]{hyperref}
\usepackage{tabularx}
\usepackage{float}
\usepackage{array}
\usepackage{multirow}
\usepackage{makecell}
\usepackage{url}
\usepackage{colortbl}
\usepackage{dsfont}
\usepackage{stmaryrd}
\usepackage{natbib}
\usepackage{caption}
\usepackage{color}
\usepackage{xcolor}
\usepackage{titlesec}
\usepackage{subcaption}
\usepackage{algorithm}
\usepackage{algpseudocode}
\usepackage{url}
\usepackage{extarrows}
\usepackage{booktabs}
\usepackage{multirow}

\renewcommand{\baselinestretch}{1.3}

\makeatletter
\@addtoreset{equation}{section}

\makeatother

\newcounter{Fig}[figure]

\newcounter{Tab}[table]

  {\refstepcounter{Tab}
   \addtocounter{table}{1}
   \samepage\vspace{0.2cm}
   \centerline{#1} \nobreak
   \begin{verse}{Table \thesection.\arabic{Tab}:}
  }%
  {\end{verse} \vspace{0.2cm}}

\relax

\newcommand{\bqa}{\begin{eqnarray*}}
\newcommand{\eqa}{\end{eqnarray*}}
\newcommand{\bqan}{\begin{eqnarray}}
\newcommand{\eqan}{\end{eqnarray}}
\newcommand{\bqt}{\begin{quote}}
\newcommand{\eqt}{\end{quote}}
\newcommand{\bt}{\begin{tabbing}}
\newcommand{\et}{\end{tabbing}}
\newcommand{\bit}{\begin{itemize}}
\newcommand{\eit}{\end{itemize}}
\newcommand{\ben}{\begin{enumerate}}
\newcommand{\een}{\end{enumerate}}
\newcommand{\beq}{\begin{equation}}
\newcommand{\eeq}{\end{equation}}
\newcommand{\bdefi}{\begin{definition}}
\newcommand{\edefi}{\end{definition}}
\newcommand{\bpro}{\begin{proposition}}
\newcommand{\epro}{\end{proposition}}
\newcommand{\blem}{\begin{lemma}}
\newcommand{\elem}{\end{lemma}}
\newcommand{\bth}{\begin{theorem}}
\newcommand{\eth}{\end{theorem}}
\newcommand{\bco}{\begin{corollary}}
\newcommand{\eco}{\end{corollary}}
\newcommand{\bdes}{\begin{description}}
\newcommand{\edes}{\end{description}}
\newcommand{\bre}{\begin{remark}}
\newcommand{\ere}{\end{remark}}

\newtheorem{definition}{Definition}[section]
\newtheorem{proposition}[definition]{Proposition}
\newtheorem{lemma}[definition]{Lemma}
\newtheorem{theorem}[definition]{Theorem}
\newtheorem{corollary}[definition]{Corollary}
\newtheorem{remark}[definition]{Remark}

\begin{document}

\begin{titlepage}

\title{A Wasserstein GAN-based climate scenario generator for risk management and insurance: the case of soil subsidence}

\author{{\large Antoine H\textsc{eranval}$^1$},
{\large Olivier L\textsc{opez}$^2$},
{\large Didier N\textsc{gatcha}$^{3}$},
{\large Daniel N\textsc{kameni}$^{2,4}$}  }

\date{\today}
\maketitle

\renewcommand{\baselinestretch}{1.1}

\begin{abstract}
According to the United Nations Office for Disaster Risk Reduction (2025), the average annual cost of natural catastrophes increased from 70–80 billion USD between 1970 and 2000 to 180–200 billion USD between 2001 and 2020. Reports from organizations such as the IFOA and the WWF highlight the need for the insurance sector to adapt to this rapidly evolving context by developing medium- to long-term strategies that go beyond the one-year horizon of prudential regulations such as Solvency II. This paper introduces an artificial intelligence framework based on Conditional Generative Adversarial Networks (Conditional GANs) to generate future spatio-temporal trajectories of climatic indices. The approach focuses on the Soil Wetness Index (SWI), a key indicator used in France to assess drought severity. Drought accounts for approximately 30\% of the indemnities paid under the French natural catastrophe insurance scheme. The proposed model, SwiGAN, simulates plausible drought propagation patterns up to 2050 for a region of France particularly exposed to this hazard. By generating realistic sequences of SWI maps, SwiGAN provides insights into drought dynamics under climate change scenarios and supports the design of adaptive risk management and insurance strategies. The methodology is also generalizable to other climate-related perils and actuarial applications such as economic scenario generation. 
\end{abstract}

\vspace*{0.5cm}


\vspace*{0.5cm}


\vspace*{0.5cm}

\noindent{\bf Short title:} GANs for climate risk management.

\vspace*{.5cm}

{\small
\parindent 0cm
$^1$ Biostatistiques et Processus Spatiaux (BioSP), INRAE, Avignon, France\\
$^2$ CREST, CNRS, Ecole polytechnique, Groupe ENSAE-ENSAI, ENSAE Paris, Institut Polytechnique de Paris, Palaiseau, France\\
$^3$ Fondation du Risque, Institut Louis Bachelier, Paris, France \\ 
$^4$ Detralytics, Paris La Défense Cedex, France. \\

E-mails: antoine.heranval@inrae.fr, olivier.lopez@ensae.fr, bakouen@gmail.com, daniel.nkameni@ensae.fr}

\end{titlepage}

\small
\normalsize
\addtocounter{page}{1}

\section{Introduction}

Current climate evolution scenarios call into question the ability of insurance against natural catastrophes to cover a risk that represents a growing financial burden. According to \cite{undrr_gar2025}, the average yearly cost of natural catastrophes increased from 70--80 billion USD between 1970 and 2000 to 180--200 billion USD between 2001 and 2020. The European Insurance and Occupational Pensions Authority recently expressed its views on the challenge of reducing the insurance protection gap in order to offer protection adapted to the forthcoming consequences of climate change. Many reports from actuarial associations (see \cite{ifoa2025_planetary_solvency}) and non-governmental organizations (see \cite{wwf2026_insurance_protection_gap}) have pointed out the need for the sector to prepare for this evolving situation.

This adaptation requires anticipation of the future state of the climate. Conceiving a medium- to long-term strategy creates a need to project beyond the one-year horizon of current prudential regulations such as the Solvency II Directive. From this perspective, insurers cannot simply reproduce the current distribution of their losses. Climate evolution scenarios are a valuable source of information but are not suited to the specific needs of the insurance sector, either because they are not focused on the appropriate indicators and/or because they fail to capture low-probability events such as those involved in the computation of the Value-at-Risk.

In the present paper, we develop a generic methodology to simulate the evolution of weather indices related to insurance products, based on deep generative models. The case study we consider is related to geotechnical drought and the phenomenon of drought-induced subsidence on clay soils, but could be extended to any situation where an index describes the severity of a natural disaster. The context of drought-induced subsidence, especially in the French market, is particularly relevant due to the specificity of the local public-private protection regime: for this hazard, compensation is triggered by the values taken by an index produced by the national meteorological agency. This mechanism is close to a parametric insurance framework\footnote{Among the most obvious differences, compensation in the French Natural Catastrophes system is not computed from the value of the index but after expertise of each individual claim. Only the characterization of the disaster is driven by the index.}, which is another important case for which our methodology is particularly relevant.

Two key principles drive our approach:
\begin{itemize}
\item one needs to take advantage of existing climate projection models and the expert assumptions made in their construction: as already noted, models produced by climate science are not designed for insurance purposes but rely on sophisticated physical models whose outputs are valuable. On the other hand, different scenarios of climate evolution (depending on our collective ability, as an international community, to slow down the increase in average temperatures) support the possibility of adjusting the generator depending on the assumptions that are made;
\item one wishes to be able to perform fast simulations without incurring too high a cost in terms of the complexity of the phenomena to be modeled. The behavior of the weather indices we focus on depends on complex parameters (related to geographical and geophysical conditions) that are difficult to anticipate in the context of a parametric model.
\end{itemize}

Deep generative models are good candidates to address the second objective. In the present paper, we rely on Generative Adversarial Networks (GANs) as a key component of our scenario generator. GANs were introduced by \cite{goodfellow2014generative} and consist of two competing deep neural networks that mutually improve each other: one generates realistic observations, while the other detects simulated data from true observations. These methods have proven their efficiency in image generation, see e.g. \cite{lu2024empirical}. Recently, \cite{flaig2022scenario} demonstrated their relevance in the development of economic scenario generators. They are currently receiving considerable interest in financial applications, see \cite{ramzan2024_gan_finance}, \cite{jiang2024_gan_data_imbalance}, \cite{xueping2025_gan_creditrisk}. See also \cite{wuth_rich}, chapter 11, where applications of generative methods to actuarial science are discussed. In our case, we rely on Wasserstein GANs, see \cite{arjovsky2017wasserstein} and \cite{gulrajani2017improved}, which are refinements of these generative techniques that tend to avoid the so-called ``mode collapse" (see \cite{srivastava2017veegan}) observed in traditional GANs. To account for non-stationarity and complex spatio-temporal dynamics, specific network structures are considered.

The rest of the paper is organized as follows. Section \ref{sec:context} is devoted to the precise problem statement, identifying the requirements of our generator and positioning our work within the current literature. The GAN methodology and the network architectures are described in Section \ref{sec:gan_methodology}, providing a general approach that can be transferred to a wide range of climate-related risks. In Section \ref{sec:data}, we present the study area and all the variables involved in our study, including the weather index on which we focus. The results of our model, as well as their validation and interpretation, are presented in Section \ref{sec:results}. In Section \ref{sec:soil_subsidence}, we focus on an analysis of drought-induced soil subsidence, where we use the results of our model to estimate the distribution of insurance costs for this hazard at the 2050 horizon. The appendix (Section \ref{sec:appendix}) gathers technical aspects of the training process and hyperparameters. All the code and results from our analysis are freely available in this online repository: \url{https://github.com/dnkameni/SwiGAN}.

\section{Context and general framework} \label{sec:context}

In this section, we provide an overview of the generative method developed in the rest of the paper. Section \ref{subsec:context} presents the context of drought-induced soil subsidence, focusing on the French example of the Grand Est region, which shows a concerning recent evolution of risk due to climate change. Section \ref{sec:prob_desc} introduces the notations and the general objective of the generative models. Closely related approaches from the literature are discussed in Section \ref{subsec:related}.

\subsection{Context of the study}
\label{subsec:context}

Projecting the medium- to long-term evolution of a climate insurance portfolio often relies on the ability to forecast a number of climate indices. Here, we consider the case of drought-induced soil subsidence, which serves as a relevant example. Drought-induced soil subsidence is a hazard observed in areas where clay is highly present in the ground. This hazard causes the weakening of the foundations and structures of buildings through the continuous swelling of the ground after heavy rainfall, followed by its shrinking during drought episodes. These damages are generally more severe in areas where buildings and vegetation are close, as plants tend to absorb soil water, thereby accelerating the process of soil subsidence. These damages usually appear in the form of sinking foundations, weakening or disruption of structural stability, and cracks in walls and floors. This hazard has become particularly concerning in France, where it is considerably weakening the public-private regime for protection against natural catastrophes (``Catastrophes Naturelles" regime). The average yearly losses on the French insurance market increased from 400 million euros over the period 1989--2015 to 1 billion euros between 2016 and 2020. According to the French Central Reinsurance Fund (CCR), drought-induced soil subsidence caused damages exceeding 3.5 billion euros in 2022. This makes subsidence one of the major concerns of the insurance sector. Although the French example is striking in magnitude, the problem is not specific to this country. Indeed, an increasing number of cases are observed in the United Kingdom, Spain, and some U.S. states such as Texas, California, and Colorado. The situation is not necessarily identical: apart from geological similarities, the level of risk depends on building standards and exposure to severe drought episodes, which may not increase at the same rate across regions.

The evolution of drought-induced soil subsidence, in a context of climate change, requires anticipation to ensure that the market will be able to cover the risk in the medium to long term. In France, compensation related to soil subsidence claims is governed by a weather index that materializes the intensity of a drought episode. This is the Uniform Soil Wetness Index (which we abbreviate as SWI), produced by Météo France, and based on this quantity (see Section \ref{subsec:SWI} for a more precise description of the criterion), a city affected by drought may or may not be eligible for insurance compensation. Hence, the evolution of this index is key to the future of protection against subsidence. The ability to project the distribution of this weather index in the future is essential to redefine the rules of protection. Should we modify the procedure used to characterize events to make insurance against drought-induced soil subsidence sustainable? What level of effort in terms of prevention should be undertaken to contain the evolution of risk? The answers to such questions depend on our ability to gain a better understanding of how the index will behave in the coming years, when climate conditions will differ from those observed today.

Climatology may provide evolution scenarios, such as those produced by the Community Climate System Model (see \cite{drake2005overview}). These simulation tools are not specifically designed for insurance purposes, and the weather indices simulated by these tools may not be adapted to insurance analysis. Moreover, running these scenarios is computationally intensive, as they rely on complex geophysical models. For instance, in the case of SWI, Météo-France provides simulations under different RCP\footnote{Stands for Representative Concentration Pathways which are presented in Section \ref{subsec:RCP}} scenarios, with only a limited number of replications for each scenario (a few dozen) due to computational constraints. This number is too small to obtain a clear view of extreme events and high quantiles of the distribution, such as the 99.5\% Value-at-Risk used in the Solvency II regulation.

Our aim is therefore to develop a methodology capable of projecting the values of a weather index based on historical and projected data, while enabling a large number of simulations to be performed within a reasonable computational time. In the case of soil subsidence and within the context of the French public-private protection system, this projection is sufficient to deduce the evolution of the frequency of claims, since compensation is solely triggered by the values of this index. When it comes to determining the financial loss, additional assumptions are required, as shown in Section \ref{subsec:houses}. Let us note that, since the framework extends to any type of weather index, this approach is particularly appealing for the study of parametric insurance products, whose compensation mechanism entirely depends on an index.

In the following, our simulation tool must fulfill the following requirements:
\begin{itemize}
\item projection of a weather index at the scale of a given territory, taking into account information on climate evolution;
\item fast computation, since we need to rapidly produce simulations in order to estimate high quantiles of the loss distribution;
\item ability to capture geographical features, since many factors (presence of forests, mountains, etc.) may influence the weather index. Their effects are difficult to capture in a simple parametric model. An efficient scenario generator should capture these effects without the need to explicitly specify or formalize them.
\end{itemize}

\subsection{Problem description}
\label{sec:prob_desc}

Let us consider an index $I_t \in \mathbb{R}^d$ that we want to project, where $t \geq 0$ is the (discrete) time. This index is multivariate because we observe its values at $d$ locations, each coordinate of $I_t$ corresponding to a fixed geographical position. Historical data on the index are available between $0$ and $T$. In the case of drought, the SWI index is available on a monthly basis, but for other types of indices, the time scale may differ.

Let us also note that we implicitly consider a setting where $d$ is large. In the case of the SWI index, the French territory is decomposed into squares with edges of 8 kilometers. Focusing on a smaller part of the territory (namely an administrative region) corresponds to a dimension $d \approx 1{,}628$.

The data exhibit both temporal and spatial dependencies. Additionally, the data $(I_t)_{t=1,\cdots,T}$ are enriched with contextual information, namely variables $X_t \in \mathbb{R}^d$ that inform us about the state of the climate at time $t$. These covariates may be of various types and resolutions, meaning that their initial dimension is not necessarily $d$, in which case projections and interpolations need to be performed to achieve the required resolution (see \cite{yang2015spatial}). The covariates essentially need to fulfill two requirements:

\begin{enumerate}
\item they need to provide relevant information about the state of the climate at a given time;
\item projections must already be available for these variables.
\end{enumerate}

Regarding this second aspect, we need to observe $(X_t)_{t=1,\cdots,T}$ to match them with the corresponding index values $(I_t)_{t=1,\cdots,T}$. We assume that we have at our disposal $(\hat{X}_t)_{t=T+1,\cdots,T+h}$, where $h$ is our projection horizon, and $\hat{X}_t$ are anticipated values. These projections are typically produced by high-level climate models under RCP scenarios, or by statistical projections coupled with stochastic weather generators such as \cite{obakrim2025multivariate}. These projections do not impact the training phase of our generator; they only play a role in the forecasting step. Once the generator is trained, the idea is to be able to plug in any type of projections for the variables $X_t$, so that one can compare and adapt to different assumptions about climate evolution. Therefore, we do not focus on how these projections are obtained.

The aim of our generator is to reproduce the conditional distribution of $I_t$ given the past values of the covariates and of the index, that is $\mathbb{P}_t(\cdot \mid C_t)$, where $C_t = ((X_s)_{t-u \leq s \leq t}, (I_s)_{t-u \leq s < t})$. Here, $u \geq 0$ is the length of a time window used to reduce the dimension and keep the dimension of $C_t$ constant. Typically, it is chosen large enough so that we can neglect the dependence between $I_t$ and the history prior to $t-u$. Given a context $C_t$, the generator can be seen as a function $G$ transforming a random noise $Z \in \mathbb{R}^{z}$ (generated from a known distribution) into a quantity in $\mathbb{R}^d$ (note that $d$ and $z$ are not necessarily of the same dimension), such that
$$\hat{I}_t=G(Z;C_t)\in \mathbb{R}^d.$$ The conditional distribution of the simulated variable $\hat{I}_t$, denoted $\mathbb{P}^*_t(\cdot \mid C_t = c)$, should ideally be close to $\mathbb{P}_t(\cdot \mid C_t = c)$ for all values of $c$.

The idea is then to rely on this approximated distribution $\mathbb{P}^*_t(\cdot \mid C_t = c)$ to generate future values of the index. The following iterative procedure can be used to project the index from $T+1$ to $T+h$ ($T$ being the latest time of observation):

\begin{itemize}
\item Simulate i.i.d. replications $Z_1, \cdots, Z_h$ under distribution $\mathbb{P}_Z$;
\item For $k = 1, \cdots, h$,
let \begin{eqnarray*}
\hat{C}_{T+k} &=& (\tilde{X}_{T+k}, \cdots, \tilde{X}_{T+k-u}, \tilde{I}_{T+k-1}, \cdots, \tilde{I}_{T+k-u}), \\
\hat{I}_{T+k} &=& G(Z_k, \hat{C}_{T+k}),
\end{eqnarray*}
where $\tilde{X}_t = X_t$ and $\tilde{I}_t = I_t$ if $t \leq T$, and $\tilde{X}_t = \hat{X}_t$ if $t > T$, and $\tilde{I}_t = \hat{I}_t$ if $T < t \leq T+k-1$.
\end{itemize}

The aim is not to forecast a single value, but to obtain an estimated distribution for $(I_{t})_{T+1 \leq t \leq T+h}$. In the insurance applications we have in mind, there is a link between the index and a financial cost of losses (in addition to our example in Section \ref{sec:soil_subsidence}, see for example \cite{rahman2021remote} in agriculture). Therefore, obtaining a distribution is convenient for estimating high quantiles and for gaining insight into the volume of reserves that will need to be accumulated in the future to comply with prudential requirements.

\begin{figure}[h!]
    \centering
    \includegraphics[width=1\linewidth]{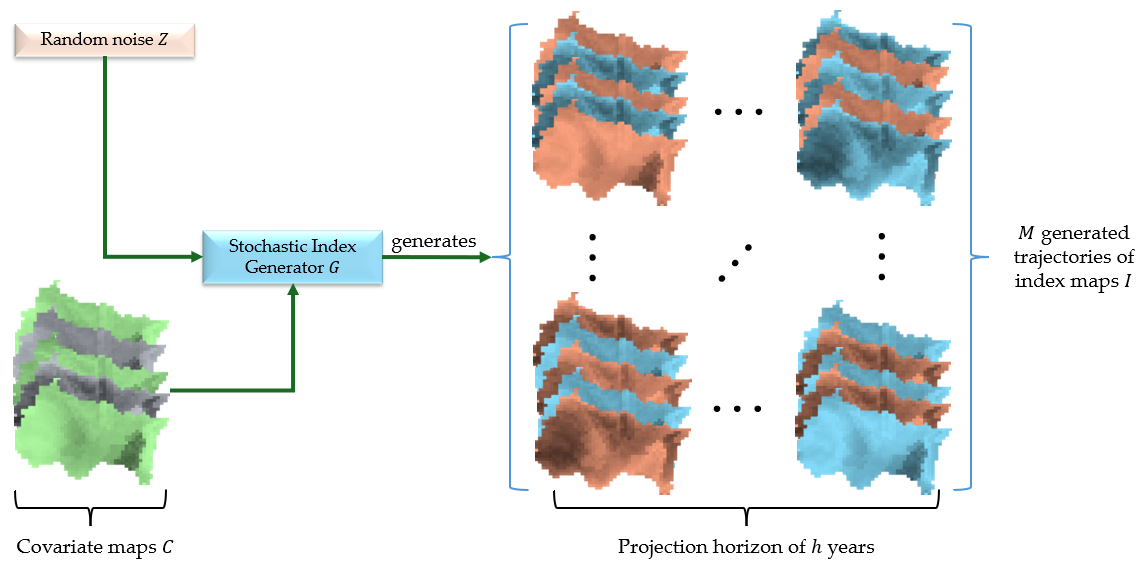}
    \caption{Description of the functioning of the proposed stochastic weather generator.}
    \label{fig:gen_functioning}
\end{figure}

The methodology is quite generic, in the sense that the function $G$ can take many forms, ranging from a parametric multivariate time series model to a deep learning generator. We explore the latter direction essentially for two reasons: the dimensionality of the problem and the complexity of spatial and temporal dependencies. We summarize the functioning of $G$ in the generation of $M$ trajectories of the index in Figure \ref{fig:gen_functioning}.

\subsection{Related work}
\label{subsec:related}

For such a problem, it is relatively natural to turn towards deep learning generative techniques such as GANs. These techniques have produced spectacular results in image generation (see for example \cite{lu2024empirical}), and our problem can be reformulated as an image generation problem, as illustrated in Figure \ref{fig:gen_functioning}. As mentioned above, the coordinates of $I_t$ are related to specific geographical locations. $I_t$ can therefore be represented as a map, with each pixel corresponding to the value of the index at that location. The idea is to benefit from advances in the field of image generation and apply them to our weather index projection problem.

Moreover, a growing number of works explore the possibility of generating or forecasting weather variables using deep learning models in general, and GANs in particular. \cite{belhajjam2024climate} use a CNN-powered GAN to generate extreme rainfall events in the cities of Tetouan and Martil in Morocco. \cite{chakraborty2025cv} propose a GAN-based model to forecast worldwide sea surface temperatures over long horizons. \cite{sha2024spatial} propose a robust methodology to increase the resolution of spatial climate data in order to assess the impacts of future climate evolutions on local environments. \cite{foroumandi2024generative} propose a GAN-based model to monitor flash droughts in real time based on the Standardized Soil Moisture Index, while \cite{shukla2023deep} rely on a similar methodology for vegetative drought prediction in India. Still in the field of drought, \cite{ferchichi2024spatio} rely on a GAN model to study the impact of climate change on drought in Africa.

In addition, in insurance and financial applications, generative methods such as GANs have also received considerable attention as potential economic scenario generators (see \cite{flaig2022scenario} and \cite{buch2023estimating}). The present work positions itself at the crossroads of these two applications, since our aim is to develop weather index generators that allow us to project financial losses (and risk measures such as Value-at-Risk) at a given horizon.

\section{Generative Adversarial Network methodology} \label{sec:gan_methodology}

This section is devoted to the description of the generative model, based on conditional Wasserstein Generative Adversarial Networks. The general description of Wasserstein GANs is presented in Section \ref{subsec:wgan}. This generative technique is based on two competing deep neural networks, the generator and the discriminator. We describe the main characteristics of the generator (which relies, in our case, on a UNet architecture) in Section \ref{subsec:gen}. The structure of the discriminator is given in Section \ref{subsec:discriminator}.

\subsection{Wasserstein GANs}
\label{subsec:wgan}

Generative Adversarial Networks, as introduced by \cite{goodfellow2014generative}, combine two neural networks: one transforms white noise into a random vector whose distribution aims to mimic a target distribution from which we observe $n$ replications. These observations are mixed with the generated ones and passed to the second network (the discriminator), which attempts to distinguish true observations from simulated ones. The two networks learn simultaneously in a competitive manner: the generator tries to improve its ability to fool the discriminator, while the discriminator gradually enhances its capacity to detect fake observations. The principle can easily be extended to conditional GANs (see \cite{jung2021conditional}), where context variables are associated with observations, with the aim of approximating a conditional distribution.

Formally, the procedure works as follows. The generator can be defined as an over-parameterized map
$(Z,C)\rightarrow G_{\theta}(Z,C),$
where $Z$ is a random noise generated from a known arbitrary distribution, $C$ corresponds to the context as defined in Section \ref{sec:prob_desc}, and $\theta \in \Theta$ gathers all the weights in the different units of the neural network. Similarly, the discriminator is a map $(I,C)\rightarrow D_{\alpha}(I,C)$, where $\alpha \in \mathcal{A}$ corresponds to the different weights or parameters to be tuned during the training of this second network.

In the initial version of GANs, the map $D_{\alpha}$ takes values in $[0,1]$, with the idea that $D_{\alpha}(I_t, C_t)$ should be close to 1 for a true observation $(I_t, C_t)$, and close to 0 for generated ones. The two-network structure of Wasserstein GANs (WGANs) is identical to that of GANs, but the objective function used to train these networks is based on the Wasserstein distance between two distributions.

The Wasserstein distance between two distributions $\mathbb{P}$ and $\mathbb{Q}$ with finite first-order moments is defined as
\begin{equation}
\begin{split}
  W(\mathbb{P}, \mathbb{Q}) & = \sup_{f \in Lip_1} \mathbb{E}_{x \sim \mathbb{P}}[f(x)]  -  \mathbb{E}_{x \sim \mathbb{Q}}[f(x)], \\
                & = \sup_{f \in Lip_1} \int f(x) d\mathbb{P}(x) - \int f(x) d\mathbb{Q}(x),
\end{split}
\end{equation}

where $Lip_1$ denotes the set of $1$-Lipschitz functions. If computing this distance were feasible, we would aim to minimize the Wasserstein distance between the empirical distribution of the observations $P_r$ and the distribution produced by the generator $P_g$. The idea of WGANs is to approximate the supremum over $Lip_1$ by a supremum over the class $\mathcal{D}_c = \{D_{\alpha}(\cdot \mid c) : \alpha \in \mathcal{A}\} \subset Lip_1$. Ideally, for a given value of the context $c$, the class $\mathcal{D}_c$ approximates the set $Lip_1$. The error of this approximation typically decreases as the size of the network increases (and also with additional constraints on activation functions to ensure that $D_{\alpha}(\cdot \mid c)$ is $1$-Lipschitz).

In our conditional WGAN framework, for a given generator $G_{\theta}$ associated with distribution $\mathbb{P}_{\theta}(\cdot \mid c)$, the discriminator computes the pseudo-Wasserstein distance
\begin{equation}\label{eq:equation_critic}
\begin{split}
W_{\mathcal{D}}(\theta) = \max_{\alpha \in \mathcal{A}} \mathbb{E}_{x \sim P_r}[D_{\alpha}(x \mid c)]  -  \mathbb{E}_{z \sim \mathcal{N}(0, \mathbb{I})}[D_{\alpha}(G_{\theta}(z \mid c))]
\end{split}
\end{equation}
where $x$ is drawn from the distribution of real maps $P_r$ and $z$ is drawn from a standard normal distribution. This loss is approximated over the training samples. Equation \ref{eq:equation_critic} shows that, similarly to the case of classical GANs, our discriminator's goal is to maximize the distance between the distributions of real and generated maps. Unlike standard GANs, the discriminator in WGANs can output values outside the interval $[0,1]$. This resolves the issue of vanishing gradients—and thus mitigates mode collapse—that arises from the optimization of the binary cross-entropy loss in GAN networks (\cite{goodfellow2014generative}).

The generator's objective is to minimize $W_{\mathcal{D}}(\theta)$, leading to the following double optimization problem:
\begin{equation}\label{eq:adv_loss_wgan}
\begin{split}
    \min_{\theta\in \Theta} \max_{\alpha \in \mathcal{A}} \mathbb{E}_{x \sim P_r}[D_{\alpha}(x \mid c)]  -  \mathbb{E}_{z \sim \mathcal{N}(0, \mathbb{I})}[D_{\alpha}(G_{\theta}(z \mid c))].
\end{split}    
\end{equation}

The quantity 
$\mathcal{L}_{adv} = \mathbb{E}_{x \sim P_r}[D_{\alpha}(x \mid c)] - \mathbb{E}_{z \sim \mathcal{N}(0, \mathbb{I})}[D_{\alpha}(G_{\theta}(z \mid c))]$
is known as the adversarial loss, which reflects the competition between the discriminator and the generator and enables the model to achieve more accurate results than classical deep learning approaches.

To enforce the 1-Lipschitz constraint on $D_{\alpha}(\cdot \mid c)$, \cite{arjovsky2017wasserstein} proposed weight clipping, which consists of limiting the range of the discriminator's weights by clamping them to a fixed interval over $\mathcal{A}$ after each gradient update. However, \cite{gulrajani2017improved} showed that this approach is suboptimal, as it increases optimization and training difficulties. They instead proposed adding a penalty term on the norm of the gradients to the adversarial loss. Since the gradients of a 1-Lipschitz function have a norm of at most 1 everywhere, this penalty enforces the 1-Lipschitz condition. This version of WGAN, which we use in our study, is referred to as WGAN-GP. The optimization problem in Equation \ref{eq:adv_loss_wgan} therefore becomes:
\begin{equation}\label{eq:adv_loss_wgan_gp}
\begin{split}
    \min_{\theta\in \Theta} \max_{\alpha \in \mathcal{A}} \mathbb{E}_{x \sim P_r}[D_{\alpha}(x \mid c)]  -  \mathbb{E}_{z \sim \mathcal{N}(0, \mathbb{I})}[D_{\alpha}(G_{\theta}(z \mid c))] + \lambda_{pen}\mathbb{E}_{\hat{x} \sim P_i}[(\| \nabla_{\hat{x}} D_{\alpha}(\hat{x})\|_2 - 1)^2]
\end{split}    
\end{equation}

Here, $P_i$ represents the distribution of linearly interpolated maps between pairs of real and generated maps. As in \cite{gulrajani2017improved}, we choose this set of maps because enforcing the gradient norm constraint on all possible maps (real and generated) is intractable. The penalty coefficient $\lambda_{pen}$ is a tunable parameter that controls how strongly the 1-Lipschitz constraint is enforced during training. 

$\hat{\theta} \in \Theta$ and $\hat{\alpha} \in \mathcal{A}$ denote, respectively, the optimal parameters of the generator and the discriminator that solve the optimization problem in Equation \ref{eq:adv_loss_wgan_gp}.

In Sections \ref{subsec:gen} and \ref{subsec:discriminator}, we describe the specific architectures of the generator and discriminator networks used in this study.

\subsection{Architecture of the generator}
\label{subsec:gen}

\begin{figure}[h!]
    \centering
    \includegraphics[width=1\linewidth]{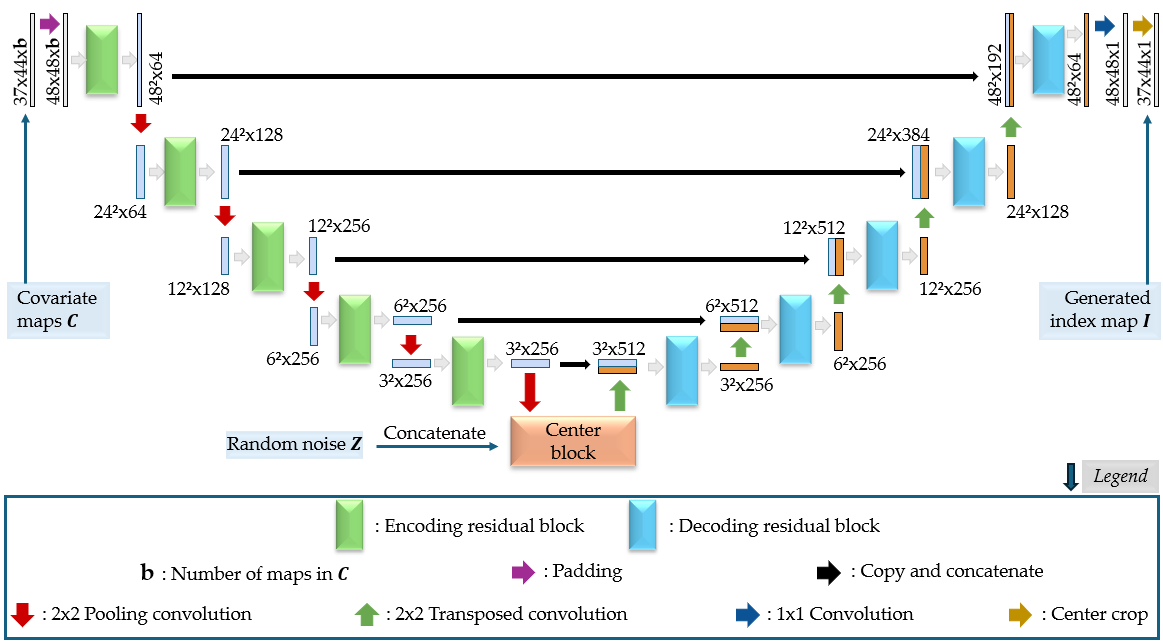}
    \caption{Architecture of the UNet generator. This generator takes covariates and a random noise $Z$ as inputs and produces an index map. The architecture of the residual blocks is illustrated in Figure \ref{fig:residual_block}.}
    \label{fig:unet_generator}
\end{figure}

As described in Figure \ref{fig:unet_generator}, we adopt a UNet architecture (see \cite{ronneberger2015unetconvolutionalnetworksbiomedical}) for the generator. Initially developed for image segmentation due to its ability to recover fine details from input images, this architecture has since been extended to various other vision tasks, including image generation with GANs (see, for example, \cite{isola2018imagetoimagetranslationconditionaladversarial}).

The idea behind the UNet structure is, first, to reduce the dimension of the input maps through an initial encoding, followed by a decoding stage at the end of the generator to retrieve an object of the appropriate dimension; second, to transfer, through skip connections and at various levels of resolution, the details learned during the encoding phase to the decoding phase. This helps improve image reconstruction.

The encoding part is composed of successive convolutional layers (CNNs) and downsampling blocks, meaning that after each layer, a map of lower resolution is obtained by aggregating the values of the pixels within a window of a given size. The decoding part, conversely, transforms these low-resolution maps back into maps of the initial resolution by applying upsampling layers after each convolutional block.

The convolutional layers from both the encoding and decoding phases are organized into residual blocks, meaning that the output of the layers is added to their inputs. This method was introduced in the ResNet model of \cite{he2015deepresiduallearningimage} to counteract the unexpected loss in performance observed in very deep neural network architectures \cite{pascanu2013difficultytrainingrecurrentneural} \cite{bianchini2014complexity}, which makes them harder to train. This phenomenon is known as the vanishing gradient problem, i.e., the progressive shrinkage of gradients during backpropagation. The authors showed that by introducing skip connections between the input and the output of a convolutional block (see Figure \ref{fig:residual_block} in the appendix), the overall flow of gradients is preserved without degrading model learning. This, in turn, facilitates the training of much deeper networks, with a reduced risk of vanishing gradients.

A more precise description of the different blocks and hyperparameters used in our application can be found in Section \ref{subsec:sup_generator}.

\subsection{Architecture of the discriminator}
\label{subsec:discriminator}

The key aspects of the discriminator architecture on which we rely are essentially of two types:

\begin{itemize}
\item the need to preserve the $1$-Lipschitz constraint, in line with the general principle of WGANs. This is achieved via instance normalization, as described in \cite{gulrajani2017improved};
\item the need to ensure that the generated maps are consistent both locally and globally.
\end{itemize}

\begin{figure}[h!]
    \centering
    \includegraphics[width=1\linewidth]{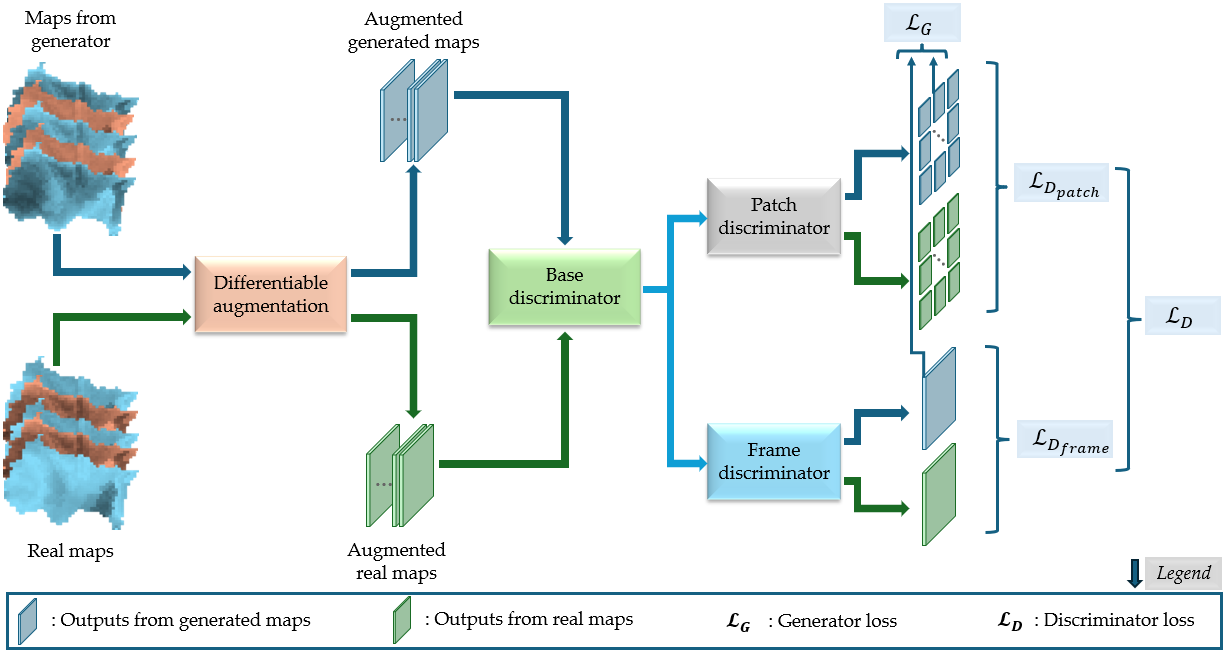}
    \caption{Architecture of the discriminator. Both the generated maps and the real maps are passed through this discriminator, which is composed of a patch discriminator and a frame discriminator sharing common layers corresponding to a base discriminator. The maps are augmented beforehand using a differentiable augmentation procedure (see Section \ref{sec:model_training} for more details).}
    \label{fig:patch_gan_discrim}
\end{figure}

Regarding this latter aspect, we decompose the discriminator into two blocks (see Figure \ref{fig:patch_gan_discrim}): a frame discriminator and a patch discriminator, sharing a set of layers that we refer to as the base discriminator. The ``frame" discriminator aims to determine whether the entire map is realistic. It takes a generated or real map and outputs a single value. The ``patch" discriminator, however, focuses on smaller regions of the map. It takes a generated or real map of dimension $D_1 \times D_2$ and outputs multiple values in the form of a $d_1 \times d_2$ grid, which is smaller than the original map. Each value of the grid is a score representing the ``realness" of the corresponding tile obtained after decomposing the input map into $N = \frac{D_1 D_2}{d_1 d_2}$ tiles. Using the notations of Section \ref{subsec:wgan}, this implies that the function $D_{\alpha}$ can be decomposed as

\begin{eqnarray}
\begin{split}
    i_{base} & = D^{(b)}_{\alpha^{(b)}}(i \mid c), \\
    D_{\alpha}(i \mid c) & = D^{(f)}_{\alpha^{(f)}}(i_{base} \mid c) + \frac{1}{N} \sum_{l=1}^N D^{(p), l}_{\alpha^{(p)}}(i_{base} \mid c)
\end{split}
\end{eqnarray}

where $i$ is the map being tested, $D^{(f)}_{\alpha^{(f)}}$ (resp. $D^{(p),l}_{\alpha^{(p)}}$) refers to the frame discriminator (resp. the $l^{th}$ output of the patch discriminator) and $D^{(b)}_{\alpha^{(b)}}$ is the base discriminator. 

The hyperparameters and specifications of the network on our data (presented in Section \ref{sec:data}) are provided in Section \ref{subsec:sup_discriminators}.

\section{The data}\label{sec:data}

We now focus on the data used to train our generative model. Some general characteristics of the Grand Est region, which is the perimeter of the following study, are given in Section \ref{subsec:study}. The SWI index, already mentioned as the weather index that is key in the analysis of drought, is described more precisely in Section \ref{subsec:descSWI}. The climate scenarios that are considered are described in Section \ref{subsec:RCP}, and the covariates used to project the evolution of the climate are presented in Section \ref{subsec:covariates}. Elements regarding data preprocessing are given in Section \ref{sec:data_preprocess}.

\subsection{Study area}
\label{subsec:study}
\begin{figure}[h!]
    \centering
    \begin{subfigure}{0.44\textwidth}
        \centering
        \includegraphics[width=0.8\linewidth]{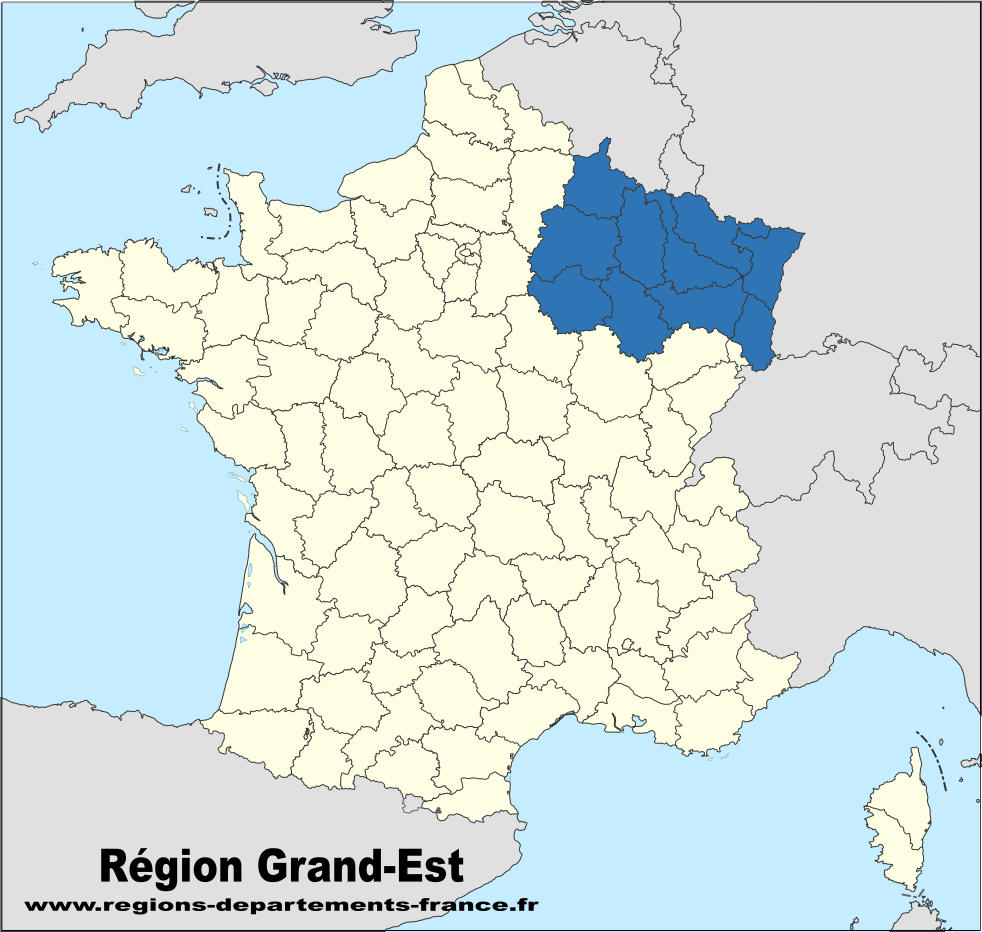}
        \captionsetup{font=small, justification=centering}
        \caption{The Grand Est region of France (Source : \href{https://www.regions-departements-france.fr/region-grand-est.html}{Regions-departements-france.fr})}
    \end{subfigure}
    \hfill
    \\
    \begin{subfigure}{0.44\textwidth}
        \centering
        \includegraphics[width=0.9\linewidth]{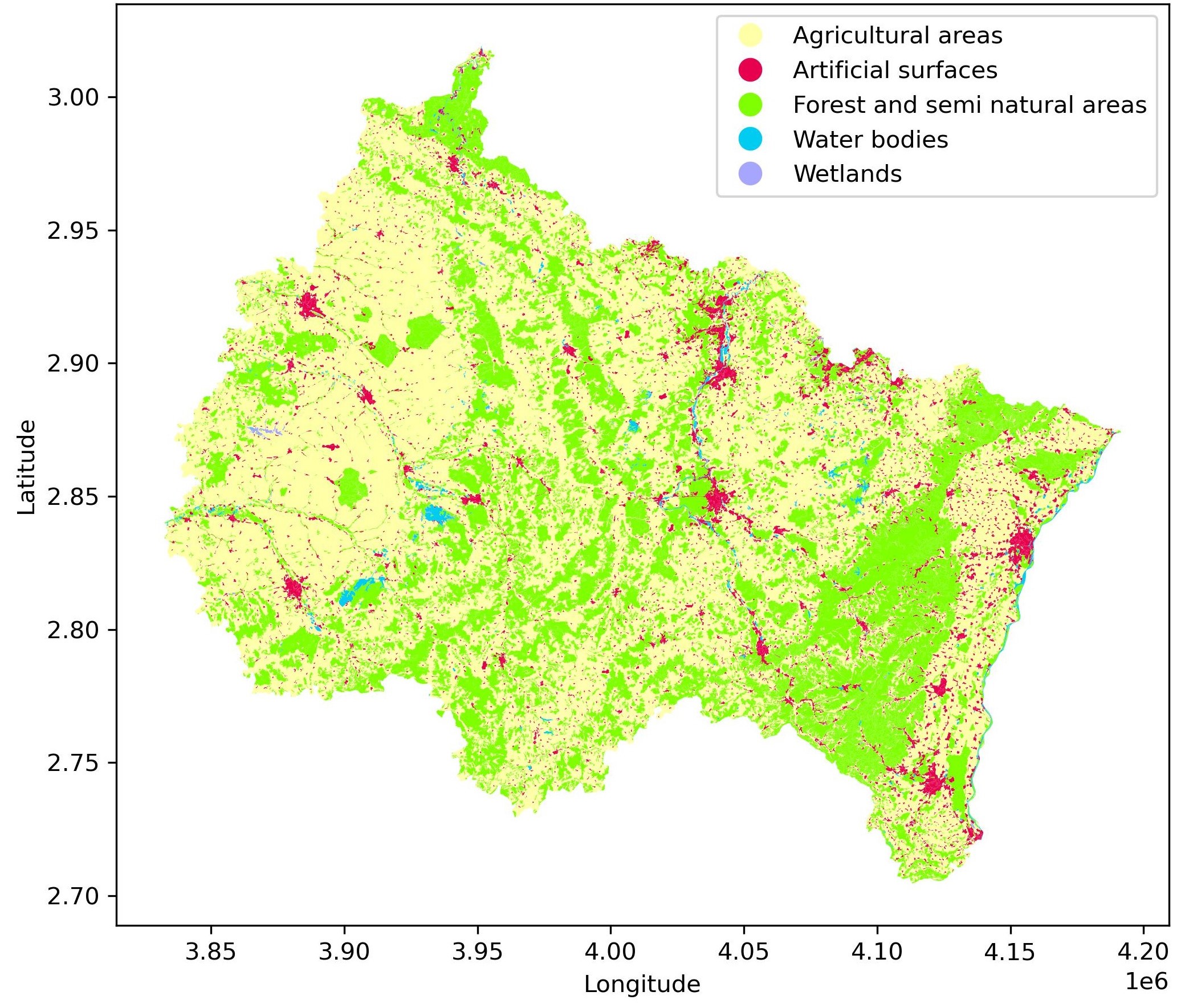}
        \captionsetup{font=small, justification=centering}
        \caption{Land cover (Source : \href{https://www.data.gouv.fr/datasets/corine-land-cover-occupation-des-sols-en-france/}{CORINE Land Cover})}
    \end{subfigure}
    \begin{subfigure}{0.44\textwidth}
        \centering
        \includegraphics[width=0.95\linewidth]{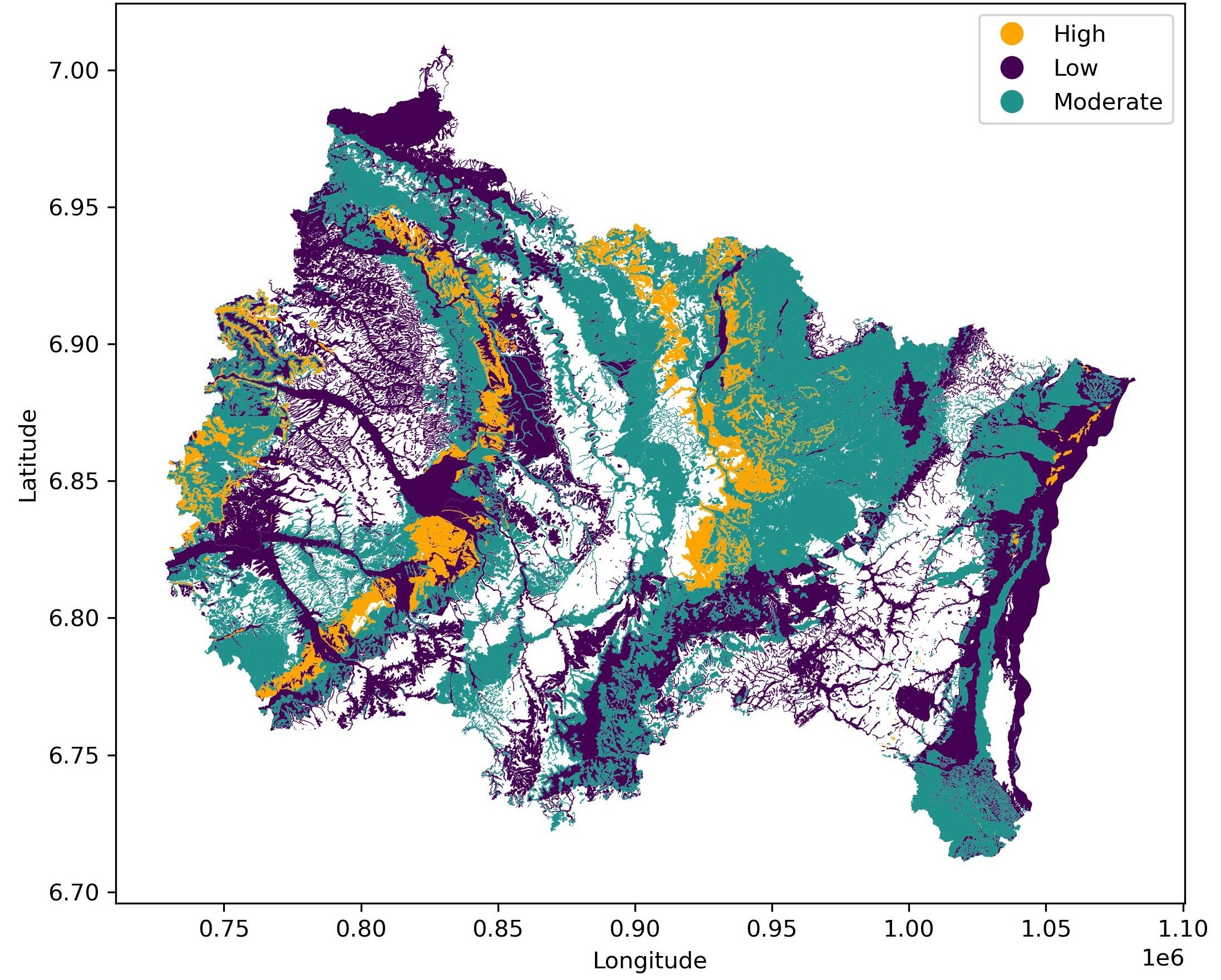}
        \captionsetup{font=small, justification=centering}
        \caption{Exposure to soil subsidence (Source : \href{https://www.georisques.gouv.fr/donnees/bases-de-donnees/retrait-gonflement-des-argiles}{Géorisques.gouv.fr}).}
    \end{subfigure}
    \caption{Socio-demographic and economic characteristics of the Grand Est region of France.}
    \label{fig:grand_est}
\end{figure}

The study area to which our model (which we will call SwiGAN henceforth) is applied is the \textbf{Grand Est} region of France (see Figure \ref{fig:grand_est}). This region has a surface area of approximately 57,430 km$^2$ (about 10\% of metropolitan France) and a population of about 5.55 million\footnote{\url{https://www.prefectures-regions.gouv.fr/grand-est/Region-et-institutions/Portrait-de-la-region/Chiffres-cles/Les-chiffres-cles-en-region-Grand-Est}}. Approximately 59.3\% of this region is agricultural land, and 33.8\% is covered by forests and semi-natural environments. The remaining urban area is particularly exposed to soil subsidence due to the significant proportion of argillaceous soils\footnote{\url{https://www.grand-est.developpement-durable.gouv.fr/IMG/pdf/25110_retrait-gonflement_argile_finale.pdf}}, which are very sensitive to variations in soil moisture. The Grand Est region is therefore particularly vulnerable to droughts, as these may impact both agriculture and buildings in urban areas. 

Moreover, by 2028, the region is expected to experience an increase in the probability of occurrence of heatwaves and drier summers\footnote{\url{https://www.grandest.fr/wp-content/uploads/2023/08/ge-brochure-changement-climatique-v2-pap-bd.pdf}}. This justifies the need for more efficient drought risk management tools and motivates our choice to focus on this region for the application of SwiGAN. 

The historical data for the Grand Est region were collected for the period 1960 to 2024. SWI, which is our target variable, and the explanatory variables are presented and described in the following sections. 

\subsection{Description of the uniform Soil Wetness Index (SWI)}
\label{subsec:descSWI}

The uniform Soil Wetness Index (uniform SWI) is a hydrometeorological indicator that quantifies the relative level of soil moisture over a depth of about 2 meters. It measures the state of soil water reserves relative to the useful water reserve level (the level required for plant growth). Its values generally range from 0 to 1, with 0 indicating extremely dry soils and 1 indicating saturated soils. Physical measurement of SWI is possible, but according to Météo-France, the process is complex and costly. Therefore, the uniform SWI is derived from land surface models that simulate the soil water balance, integrating information on precipitation, evapotranspiration, runoff, and soil characteristics. 

The term ``uniform" associated with the Soil Wetness Index studied in this paper refers to a specific configuration of the SIM model\footnote{For more information on the uniform SWI and the SIM model, see: \url{https://donneespubliques.meteofrance.fr/client/document/fiche_swi_uniforme_vf_409.pdf}} which ensures that soil geological characteristics and vegetation cover are homogeneous across the entire French territory. This configuration was specifically designed in 2009 for the monitoring and analysis of drought-induced soil subsidence hazards within France's natural catastrophes regime. It was chosen because it is best suited to isolate and reproduce the meteorological effects driving drought-induced soil subsidence in clay soils.

Because it can be computed consistently over long historical periods, the SWI is widely used to assess soil moisture deficits, monitor drought conditions, and analyze the impacts of climate change on hydrological processes. Beyond its institutional role, soil-moisture–based indices such as the SWI have also been shown to be highly relevant for economic impact modeling. In particular, \cite{Charpentier2022} demonstrate that a smoothed SWI (aggregated over three months) is a strong predictor of both the frequency and severity of soil subsidence claims in France, significantly improving the statistical performance of models for drought-related insurance costs. Similarly, \cite{EcotoChambaz2022} use hydrometeorological indicators, including soil-moisture indices, within a Super Learner framework to forecast the cost of drought events in France, highlighting the predictive value of soil-moisture information for climate-related financial risk.

In the rest of this paper, the term SWI will refer to the uniform SWI, which is computed and used by Météo-France in the reports prepared for the interministerial CatNat commission as part of the institution’s contribution to France's natural catastrophes regime\footnote{More details are available here: \url{https://meteofrance.fr/missions/contribuer-la-securite/meteo-france-dans-le-dispositif-catnat}}.

\subsection{Description of the Representative Concentration Pathways (RCP)}
\label{subsec:RCP}

Representative Concentration Pathways (RCPs) are standardized greenhouse gas concentration trajectories developed for climate change assessments. Each RCP corresponds to a specific level of radiative forcing reached by the end of the 21st century (e.g., RCP 2.6, RCP 4.5, RCP 6.0, and RCP 8.5), reflecting different assumptions regarding socio-economic development, energy use, and mitigation policies. These scenarios provide a coherent framework for comparing climate model projections and for assessing the potential impacts of climate change under contrasted emission pathways.

In this study, we focus on RCP 4.5 and RCP 8.5, which represent intermediate and high-emission pathways, respectively. RCP 4.5 assumes the implementation of mitigation policies leading to a stabilization of radiative forcing during the 21st century, whereas RCP 8.5 corresponds to a scenario of continued increases in greenhouse gas emissions with limited mitigation efforts. Considering these two contrasted scenarios allows us to explore a range of plausible future evolutions of the SWI under moderate and extreme climate change conditions.

\subsection{Description of the covariates}
\label{subsec:covariates}

The modeling framework relies on a set of physically based meteorological covariates that describe the main components of the surface water and energy balance. These variables are commonly used in hydrological and land-surface modeling, and they directly influence soil moisture dynamics, making them relevant predictors of the SWI and its projected evolution under future climate scenarios. Table \ref{tab:covariates} summarizes the covariates considered in this study.

\begin{table}[h!]
\centering
\begin{tabular}{lll}
\hline
\textbf{Variable} & \textbf{Description} & \textbf{Unit} \\
\hline
\texttt{huss} & Specific humidity near the surface & kg/kg \\
\texttt{prtot} & Total precipitation & kg/m$^{2}$/s \\
\texttt{rlds} & Incoming longwave radiation & W/m$^{2}$ \\
\texttt{rsds} & Incoming shortwave (visible) radiation & W/m$^{2}$ \\
\texttt{tas} & Daily mean near-surface air temperature & K \\
\texttt{tasmax} & Daily maximum near-surface air temperature & K \\
\texttt{tasmin} & Daily minimum near-surface air temperature & K \\
\texttt{evspsblpot} & Potential evapotranspiration (Penman--Monteith, SICLIMA) & -- \\
\hline
\end{tabular}
\caption{Meteorological covariates used to model the SWI and its evolution.}
\label{tab:covariates}
\end{table}

These variables collectively describe the atmospheric drivers of soil moisture. Precipitation (\texttt{prtot}) and specific humidity (\texttt{huss}) provide direct inputs to the soil water balance, determining water availability. Air temperature (\texttt{tas}, \texttt{tasmax}, \texttt{tasmin}) and radiative fluxes (\texttt{rsds}, \texttt{rlds}) control evaporative demand and soil drying processes, while wind speed (\texttt{sfcWind}) influences the soil’s drying capacity.

Potential evapotranspiration (\texttt{evspsblpot}), computed following the Penman--Monteith formulation in the SICLIMA framework, integrates several of these drivers into a physically consistent representation of atmospheric evaporative demand. Because evapotranspiration and precipitation jointly determine the temporal evolution of soil moisture, this variable is particularly important for explaining and predicting the SWI.

All covariates originate from global climate model outputs and are available across historical and future periods under various RCP scenarios. Their physical relevance and availability in climate projections make them suitable for estimating the SWI and for analyzing its expected evolution under climate change. We show the evolution of three key covariates under both emission scenarios (RCP 4.5 and RCP 8.5) in Figures \ref{fig:temp_rcp45} to \ref{fig:precip_rcp85} in the Appendix. The spatial maps clearly indicate more extreme variations in covariate values under RCP 8.5 compared to RCP 4.5, consistently across the different time horizons considered. This contrast highlights the amplified climatic signal associated with the high-emission scenario, particularly in the later decades.

\subsection{Data preprocessing}
\label{sec:data_preprocess}

Data preprocessing is a crucial step in spatio-temporal data modeling with deep learning. This includes all steps related to data preparation, normalization, and splitting. All weather covariates were downloaded in NetCDF file format from the DRIAS website\footnote{To access the historical data and RCP projections of weather covariates, create a free account here: \url{https://www.drias-climat.fr/commande}}. The SWI was downloaded in CSV format from Météo France's website\footnote{SWI data is available free of charge on this website: \url{https://donneespubliques.meteofrance.fr/?fond=produit&id_produit=301&id_rubrique=40}}. Since the SWI dataset is only available at a monthly time step, all weather covariates originally available at a daily time step were aggregated to a monthly time step. The aggregation of the covariates presented in Table \ref{tab:covariates} was performed by considering monthly maxima. For precipitation, both total and average monthly precipitation were also considered, resulting in a total of 11 weather covariates. Regarding the spatial resolution of the data, all datasets (covariates and SWI) are available on an identical regular grid of 8 km $\times$ 8 km, called the SAFRAN\footnote{Système d’analyse fournissant des renseignements atmosphériques à la neige} grid. This grid divides metropolitan France into 9\,892 pixels.

During preprocessing, all variables were converted into raster format, which is suitable for ingestion by deep learning models. Each raster in the Grand Est region has dimensions of 37 $\times$ 44, corresponding to 1,628 pixels (see panel (a) of Figure \ref{fig:grand_est_add_info} in the Appendix). There were no missing data in the variables used in this study. Over the period 1960 to 2024, a total of 760 monthly rasters were produced for each variable. The period 1960 to 2020 was used for training, 2021 to 2022 for validation, and 2023 to 2024 for testing. All variables were normalized using the means and standard deviations computed on the training set.

\section{Forecasting results} \label{sec:results}

\begin{figure}[h!]
    \centering
    \includegraphics[width=1\linewidth]{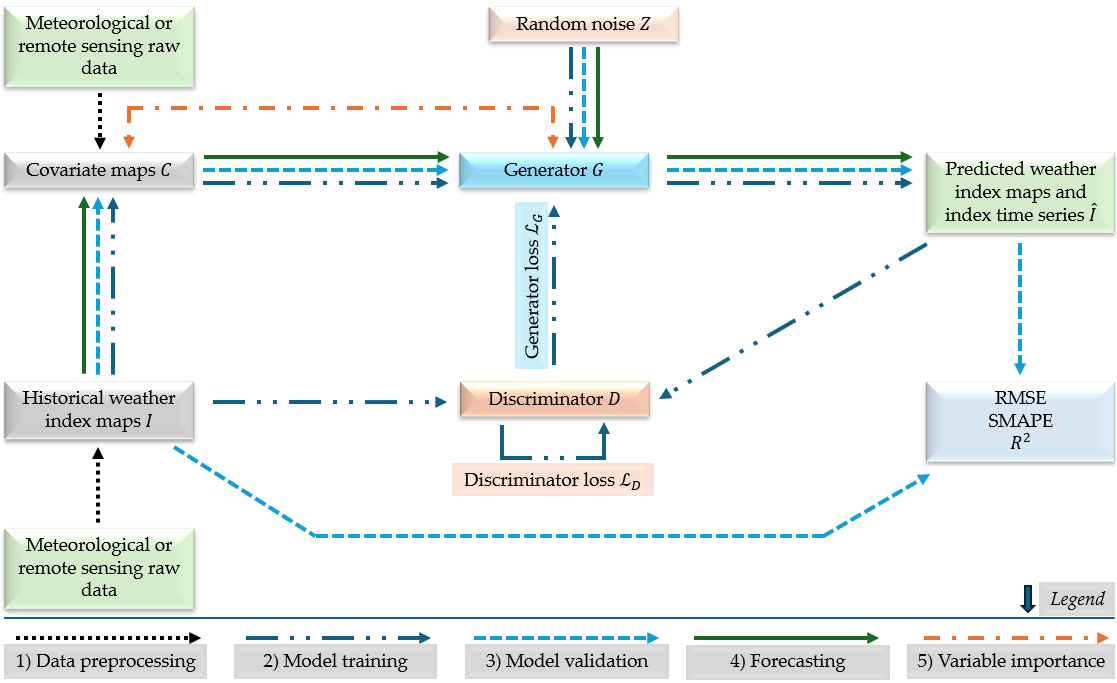}
    \caption{General architecture of the proposed forecasting methodology.}
    \label{fig:forecast_arch}
\end{figure}

To obtain the results presented in this section using the data described in Section \ref{sec:data}, we follow the steps of the architecture presented in Figure \ref{fig:forecast_arch}. Model validation, model analysis, and variable importance are presented in Sections \ref{sec:model_validation}, \ref{sec:results_analysis}, and \ref{sec:var_importance}, respectively.

\subsection{Validation of the final model}\label{sec:model_validation}

To evaluate the quality of the proposed model on the test period, we use a series of metrics proposed by related work (see Section \ref{subsec:related}). Let $y_i$ denote the observed SWI at a given pixel for month $i$ and $\hat{y}^j_i$ the generated SWI at that pixel for trajectory $j$ and month $i$. $i$ ranges from 1 (January 2023) to $n=24$ (December 2024). These metrics are defined for each trajectory in Table \ref{tab:eval_metrics}.

\begin{table}[h!]
\centering
\begin{tabular}{lll}
\hline
\textbf{Metric}                             & \textbf{Abbreviation} & \textbf{Formula} \\ \hline
Mean Square Error                           & $MSE_j$                & $\frac{1}{n}\sum_{i=1}^{n}(y_i-\hat{y}_i^j)^2$                 \\
Root Mean Square Error                      & $RMSE_j$               & $\sqrt{MSE_j}$                 \\
Mean Absolute Error                         & $MAE_j$                & $\frac{1}{n}\sum_{i=1}^{n}|y_i-\hat{y}_i^j|$                 \\
Standerdized Mean Absolute Percentage Error & $SMAPE_j$              & $\frac{1}{n}\sum_{i=1}^{n}\frac{|y_i-\hat{y}_i^j|}{(|y_i| + |\hat{y}_i^j|)/2} \times 100\%$                 \\
Coefficient of determination                & $R^2_j$                  & $1 - \frac{\sum_{i=1}^{n}(y_i-\hat{y}_i^j)^2}{\sum_{i=1}^{n}(y_i-\Bar{y}^j)^2}$                 \\
Correlation coefficient                     & $\rho_j$                 & $1 - \frac{\sum_{i=1}^{n}(y_i-\Bar{y}^j)(\hat{y}_i-\Bar{\hat{y}}^j)}{\sqrt{\sum_{i=1}^{n}(y_i-\Bar{y}^j)^2}\sqrt{\sum_{i=1}^{n}(\hat{y}_i-\Bar{\hat{y}}^j)^2}}$                 \\ \hline
\end{tabular}
\caption{Model evaluation metrics. For a given sequence $(x_i)_{1\leq i \leq n}$, $\Bar{x} = n^{-1}\sum_{i=1}^{n}x_i$.}
\label{tab:eval_metrics}
\end{table}

The final values of the metrics are obtained by averaging over all $M$ generated scenarios. The MSE, RMSE, MAE, and SMAPE are used to assess the ability of the model to generate accurate SWI trajectories. The $R^2$ and $\rho$ metrics assess the ability of the model to reproduce the temporal dynamics observed in the real data. Taken together, these metrics provide an effective evaluation of the model’s ability to generate realistic scenarios. 

\begin{figure}[h!]
    \centering
    \includegraphics[width=1\linewidth]{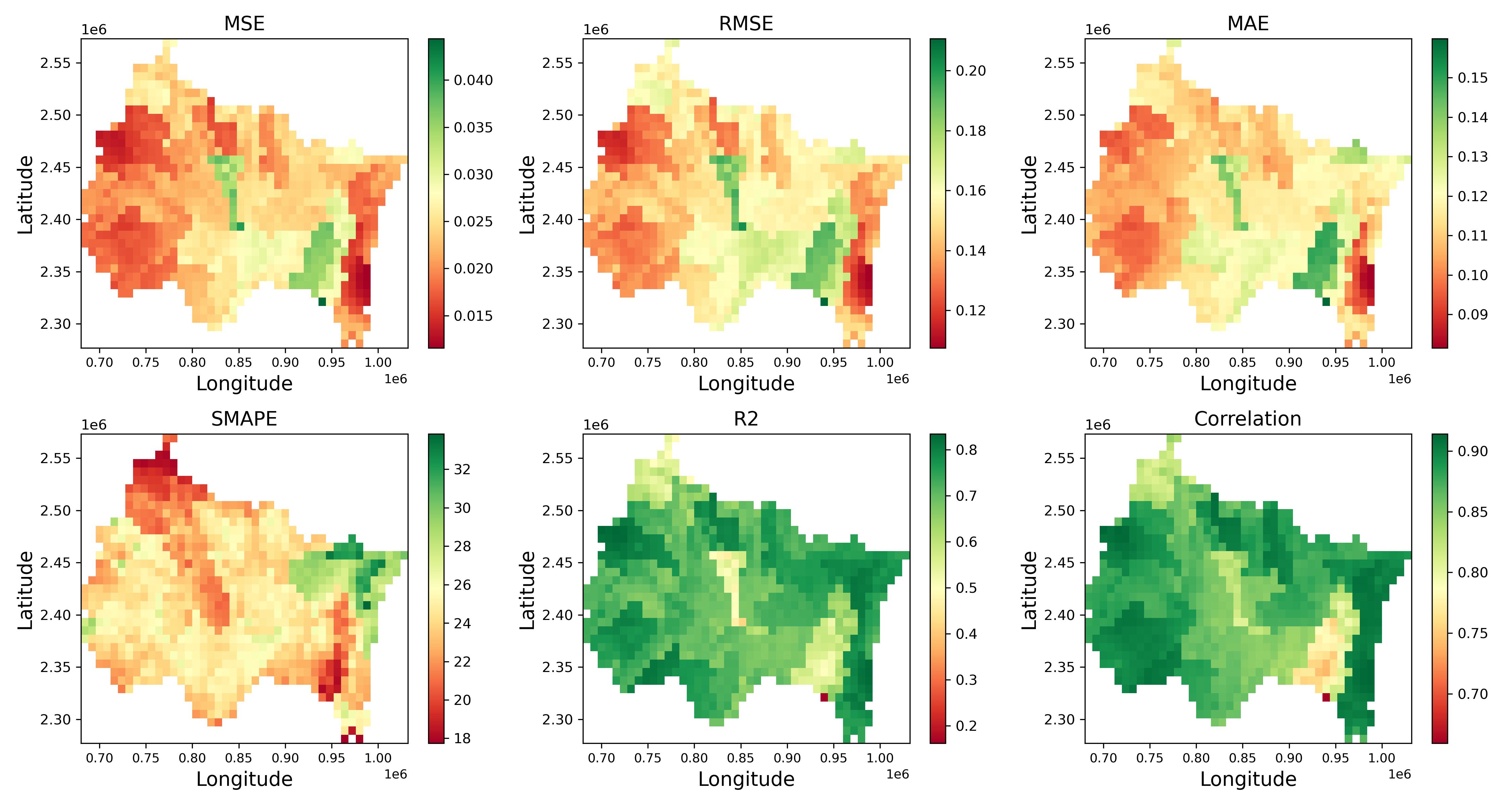}
    \caption{Values of the validation metrics in Table \ref{tab:eval_metrics} for the final SwiGAN model.}
    \label{fig:test_metrics}
\end{figure}

We compute the metrics defined in Table \ref{tab:eval_metrics} on the test period for each pixel in the Grand Est region. These metrics are presented in Figure \ref{fig:test_metrics}. 

Recall that the objective of this work is not to predict SWI values with perfect accuracy, but rather to generate realistic trajectories of SWI conditioned on covariates. Overall, the performance of the SwiGAN model is satisfactory. Across the region, the model exhibits a maximum RMSE of 0.21, with approximately 80\% of pixels having an RMSE below 0.16. Regarding SMAPE, the model reaches a maximum value of 33.83\%, with more than 80\% of pixels exhibiting a SMAPE below 25.59\%.

The SwiGAN model globally succeeds in capturing the trend in the SWI data. This conclusion is supported by the values of the coefficient of determination and the correlation coefficient. Indeed, $R^2$ and $\rho$ values are generally very high across the region, with a maximum $R^2$ of 0.83 and a maximum $\rho$ of 0.91. Approximately 80\% of pixels have $R^2$ values above 0.67, and 80\% of pixels have $\rho$ values above 0.85. 

\begin{figure}[h!]
    \centering
    \includegraphics[width=1\linewidth]{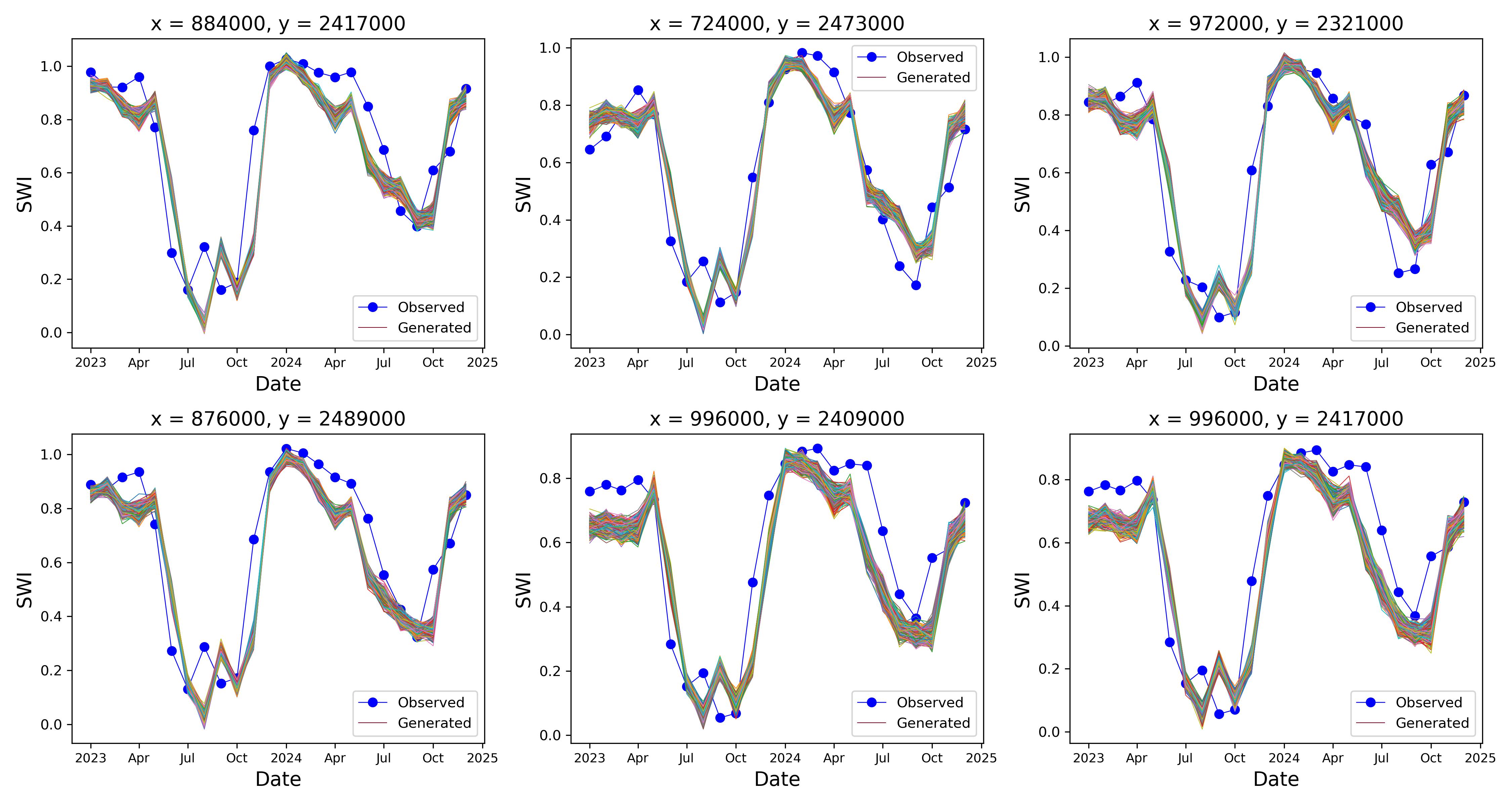}
    \caption{Observed SWI trajectory and generated SWI trajectories over the test period. The generated trajectories follow the same trend as the observed trajectory, as suggested by the values of $R^2$ and $\rho$.}
    \label{fig:test_swi_trajectories}
\end{figure}

In preparation for the use case developed in Section \ref{sec:soil_subsidence}, we further assess the validity of SwiGAN by comparing the observed SWI trajectory with the generated SWI trajectories over the test period for the six pixels with the highest number of buildings. Note that the higher the number of buildings in a pixel or town, the greater the potential financial damages that could be caused by drought-induced soil subsidence (see Section \ref{subsec:houses}). These trajectories are presented in Figure \ref{fig:test_swi_trajectories}. The figure shows that all trajectories generated by SwiGAN follow the trend of the real data, as suggested by the values of $\rho$ and $R^2$. In addition, in some cases, SwiGAN generates SWI values that are more extreme than those observed, while remaining realistic. This indicates that mode collapse was avoided during the training process and that the proposed model exhibits good generalization properties. These extreme but plausible scenarios may be useful for studying and preparing coping and risk management strategies for future catastrophic events.

\begin{figure}[H]
    \centering
    \begin{subfigure}{0.44\textwidth}
        \centering
        \includegraphics[width=0.875\linewidth]{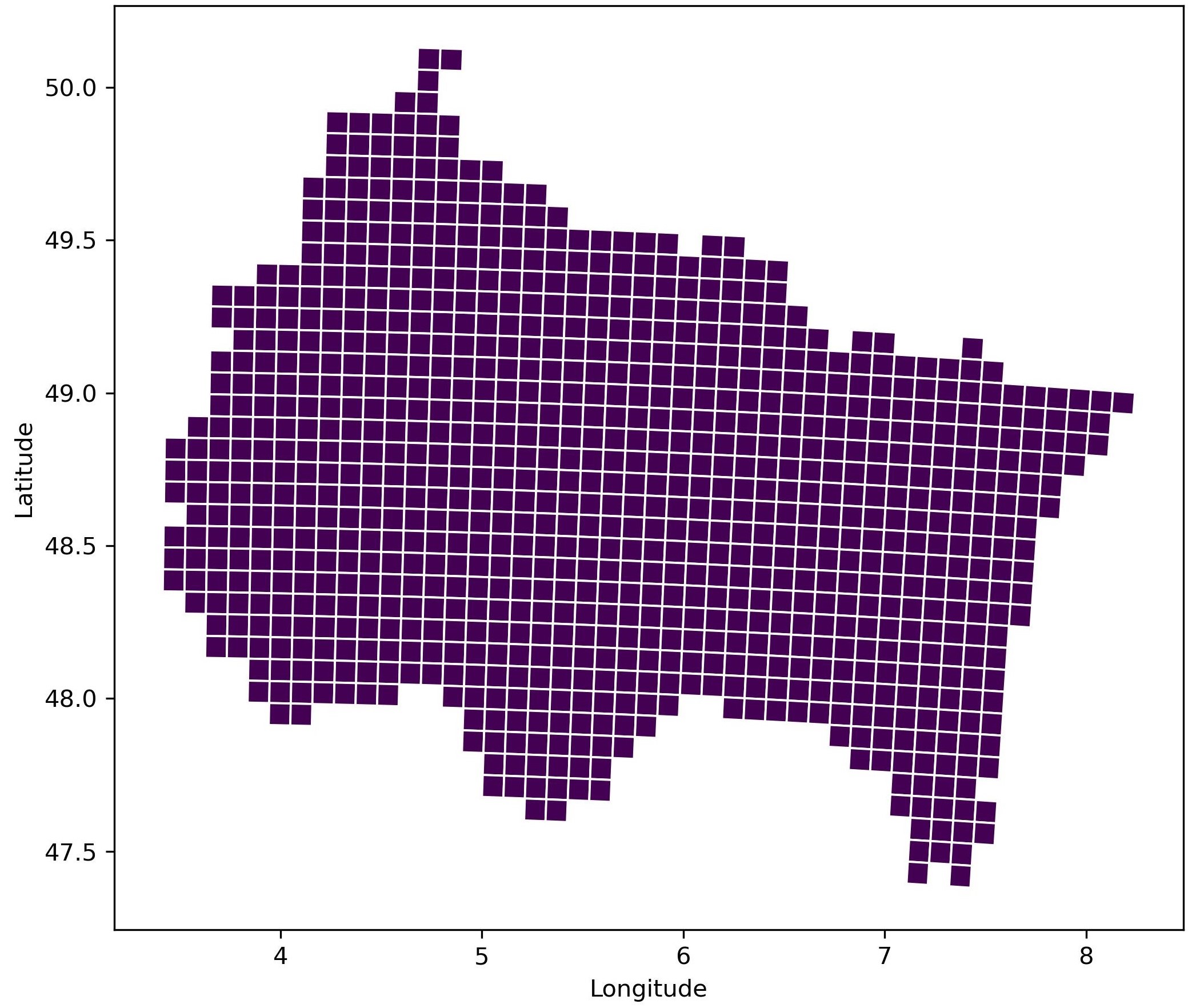}
        \captionsetup{font=small, justification=centering}
        \caption{8km $\times$ 8km safran grid}
    \end{subfigure}
    \begin{subfigure}{0.44\textwidth}
        \centering
        \includegraphics[width=1\linewidth]{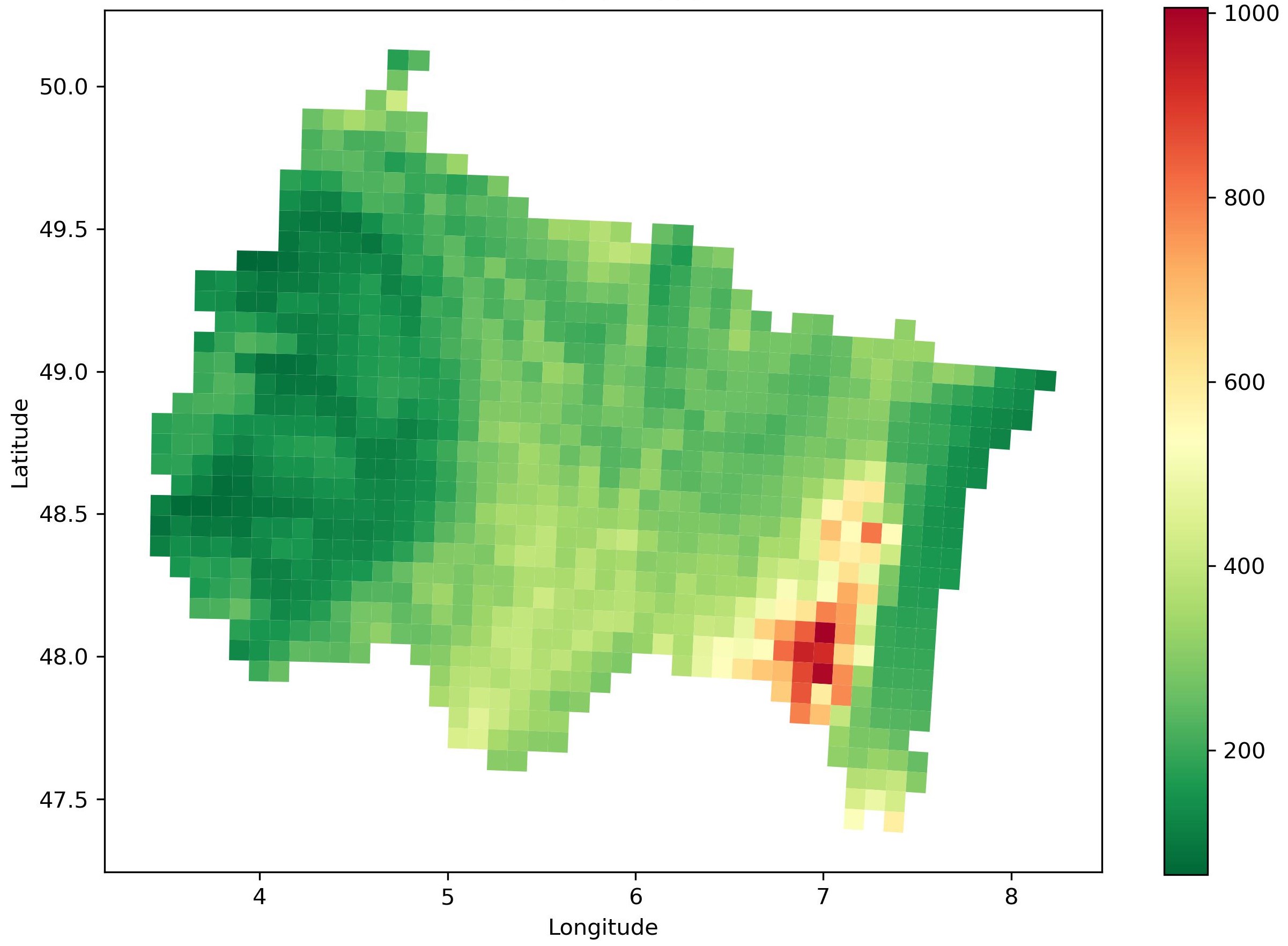}
        \captionsetup{font=small, justification=centering}
        \caption{Altitude of each grid cell}
    \end{subfigure}
    \caption{Additional information on the Grand Est region of France}
    \label{fig:grand_est_add_info}
\end{figure}

The lowest values of the validation metrics are observed in the center and south-eastern parts of the region (see Figure \ref{fig:test_metrics}). These areas are almost entirely covered by forests and semi-natural environments (see panel (b) of Figure \ref{fig:grand_est}). This observation is less concerning, as our primary focus for drought forecasting lies in agricultural and residential areas affected by drought-induced soil subsidence. When comparing Figure \ref{fig:test_metrics} with panel (b) of Figure \ref{fig:grand_est_add_info}, we observe that the pixels where SwiGAN performs worst are mainly located at high altitudes and in mountainous areas, such as the Vosges and Ardennes mountain ranges. The relatively lower performance of SwiGAN in mountainous regions is not an isolated case. Indeed, several empirical studies, such as \cite{cai2014assessment}, \cite{markstrom2016towards}, \cite{mazrooei2019improving}, \cite{wrzesien2019characterizing}, and \cite{foroumandi2024generative}, have highlighted the difficulty of weather models in achieving strong performance in snow-dominated and mountainous regions.

According to \cite{foroumandi2024generative}, the reduced performance of drought forecasting models in such regions can be explained by the scarcity of data representative of these zones. Indeed, these areas are limited within the Grand Est region, resulting in datasets that are insufficient to capture the environmental dynamics specific to these zones. These dynamics include interactions between temperature and elevation, which can lead to atypical effects on the occurrence and intensity of drought. Another specificity of mountainous areas lies in soil characteristics, which influence infiltration, groundwater storage capacity, discharge rates, and soil moisture (\cite{strachan2017testing}). Finally, certain weather covariates that strongly impact soil moisture—and consequently drought conditions—such as precipitation, temperature, and evapotranspiration, exhibit different behaviors in mountainous areas \cite{foroumandi2024generative}. 

To assess the relative performance of our model, we reviewed the literature for similar studies. \cite{wang2024soilmoistureprediction} compares the performance of multiple classical machine learning algorithms as well as LSTM-based networks for predicting the soil moisture index. Similarly to the present work, the authors use a combination of meteorological inputs related to soil moisture, such as precipitation, atmospheric temperature, and humidity, along with historical soil moisture data to construct their covariates. However, unlike SwiGAN, these models are trained on 30 different stations considered independently, without accounting for spatial interactions. Other works, such as \cite{foroumandi2024generative}, are specifically tailored to the prediction of other climate variables like precipitation, for which observations are more frequent. Under these conditions, a fair comparison between these previous studies and the present work is not possible. To the best of our knowledge, there is no prior study that uses deep generative models to produce realistic trajectories of the SWI over a contiguous geographic region.

The analysis of performance metrics and the explanation of reduced model performance in certain areas lead to the conclusion that the proposed SwiGAN model exhibits good overall performance and very good performance in agricultural and residential areas, which are the main focus of Section \ref{sec:soil_subsidence}. The SwiGAN model also demonstrates strong generalization capabilities while efficiently reproducing the trends observed in the real data. 

\subsection{Results analysis and discussions}
\label{sec:results_analysis}

Using the proposed SwiGAN model, we generate SWI data over the period 2025 to 2050. This generation is based on projections of weather covariates under the RCP 4.5 and RCP 8.5 scenarios. We generate $M = 1\,000$ trajectories of SWI over this period. To facilitate discussion and interpretation, the results presented in this section correspond to the mean trajectory, i.e., the average SWI values over the $1\,000$ trajectories. This mean trajectory is often used in empirical studies as the best estimate of the forecast of the weather index for a given period (see \cite{ferchichi2024spatio} and \cite{foroumandi2024generative}). To access all simulations from 2025 to 2050, please visit this repository: \url{https://github.com/dnkameni/SwiGAN}.

\begin{figure}[h!]
    \centering
    \includegraphics[width=1\linewidth]{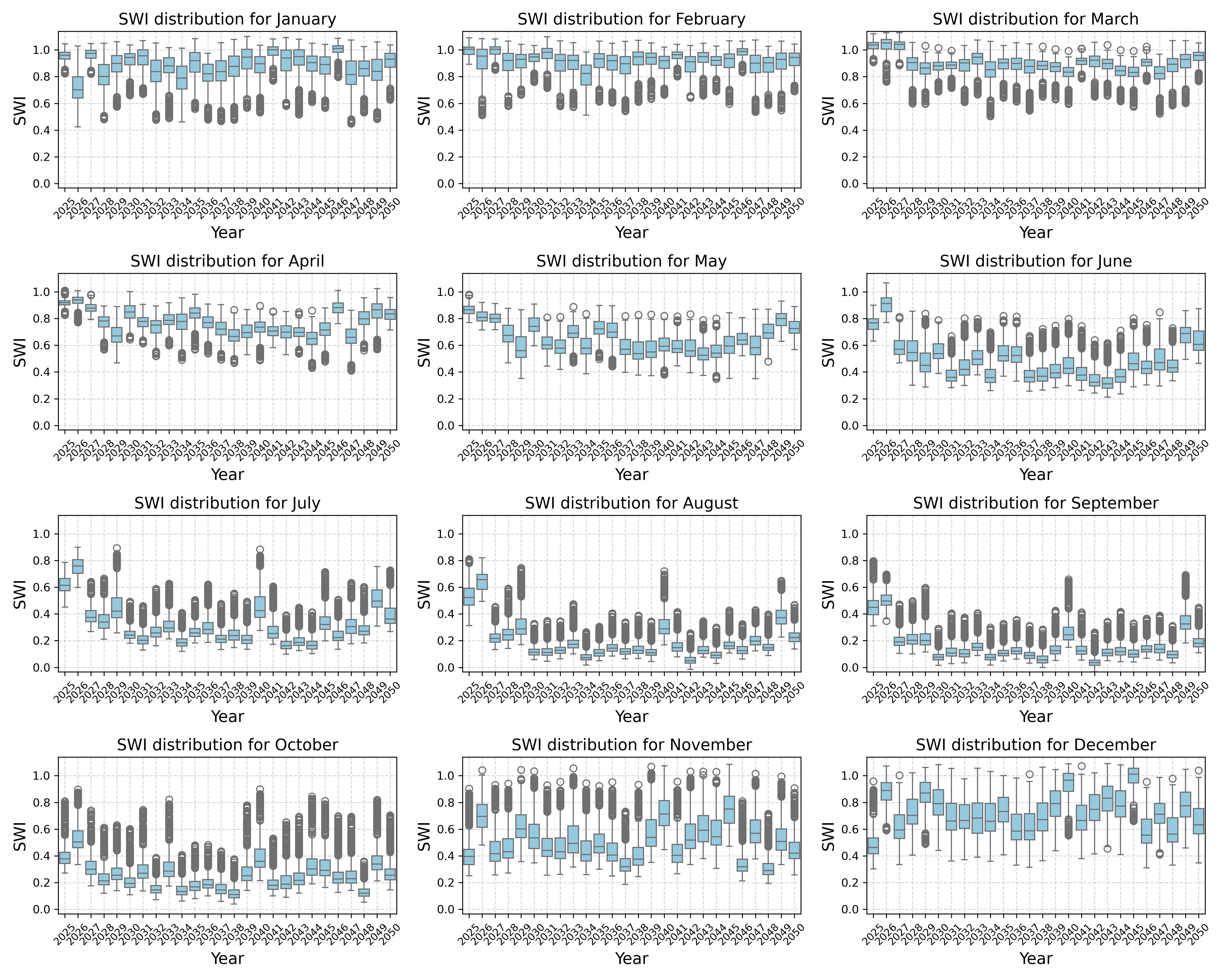}
    \caption{Spatial distribution of the mean trajectory of SWI values generated under RCP 4.5 by SwiGAN.}
    \label{fig:dist_swi_rcp45}
\end{figure}

Figures \ref{fig:dist_swi_rcp45} and \ref{fig:dist_swi_rcp85} (in the appendix) show the evolution of the spatial distribution of SWI under the RCP 4.5 and RCP 8.5 scenarios, respectively, over the period 2025 to 2050. For both RCP scenarios, the SwiGAN model generates relatively stable and high values of SWI during wet months (December, January, February, March, April, and May) over the considered period. During these months, precipitation, temperature, and evapotranspiration conditions result in higher soil water content (hence higher SWI values) and 
lower variability (hence more stable SWI values) than during drier months of the year. For drier months (June, July, August, September, October, and November), which include summer months, average SWI values are much lower and exhibit a decreasing trend over the period 2025 to 2050. Over this period, we also observe greater variability in generated SWI for these dry months than for wetter months. The decreasing trend in SWI values during these months may be a direct consequence of global warming, which is incorporated into the construction of the two RCP scenarios.

As expected, SWI values generated by SwiGAN under the RCP 8.5 scenario are generally lower than those generated under the RCP 4.5 scenario. As discussed in Section \ref{subsec:RCP}, the RCP 8.5 scenario is more extreme and pessimistic, assuming an average increase in global temperatures of 3.2$^\circ$C to 5.4$^\circ$C, compared to 1.7$^\circ$C to 3.2$^\circ$C for the RCP 4.5 scenario, which is considered an intermediate scenario (\cite{soubeyroux2020nouvelles}). These results are also consistent with empirical studies predicting an increase in drought episodes under the RCP 8.5 scenario (see \cite{lin2020drought}). The RCP 4.5 scenario is generally considered more probable, whereas the RCP 8.5 scenario is typically used as a stress-test scenario in insurance, finance, and infrastructure studies. However, a recent report from Météo-France suggests that the evolution of weather in France over the period 2005 to 2024 appears closer to the RCP 8.5 scenario than to the RCP 4.5 scenario (see \cite{soubeyroux2020nouvelles}).

\begin{figure}[h!]
    \centering
    \begin{subfigure}{0.735\textwidth}
        \centering
        \includegraphics[width=1\linewidth]{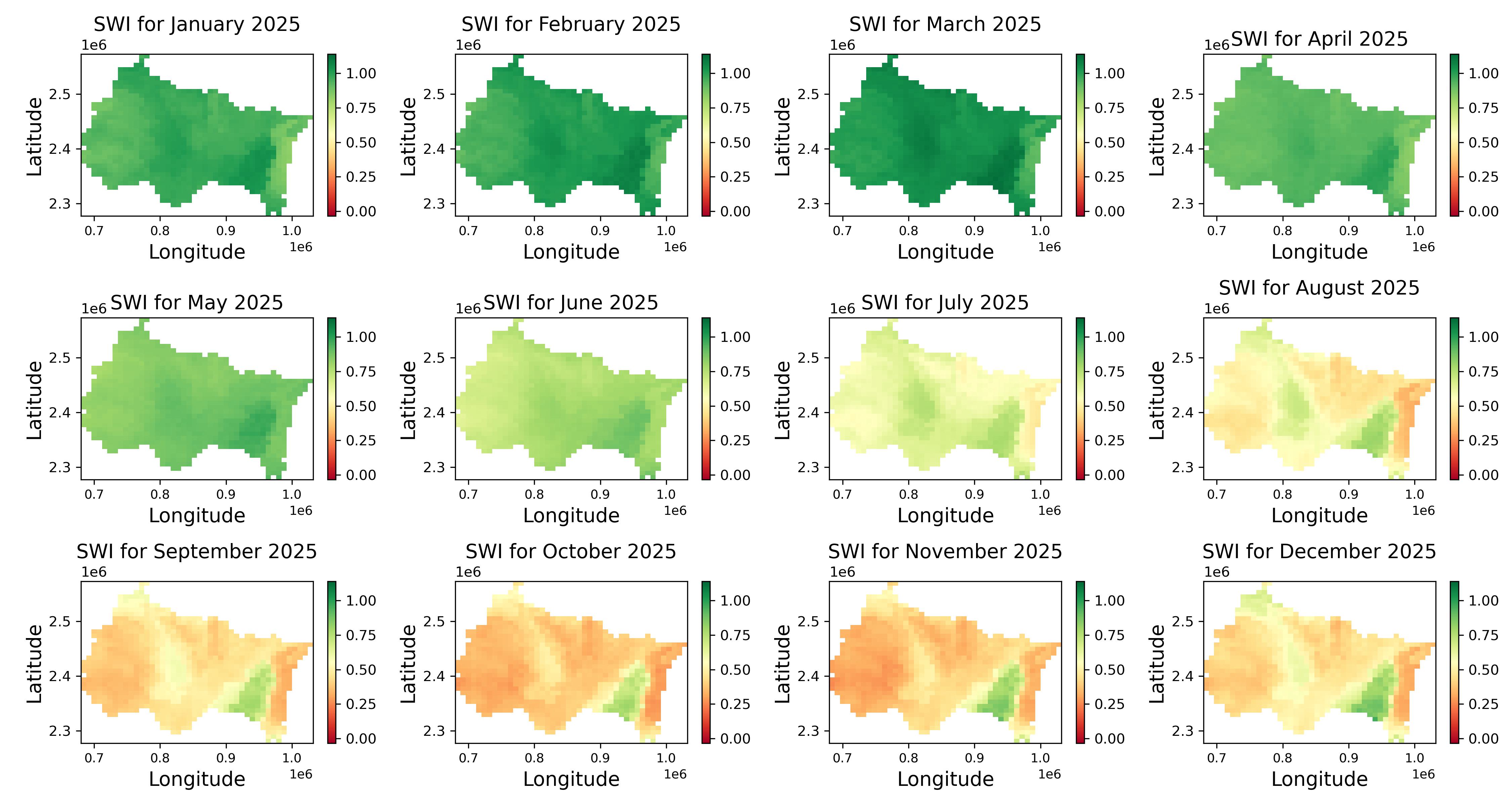}
        \captionsetup{font=small, justification=centering}
        \caption{2025}
    \end{subfigure}
    \hfill
    \\
    \begin{subfigure}{0.735\textwidth}
        \centering
        \includegraphics[width=1\linewidth]{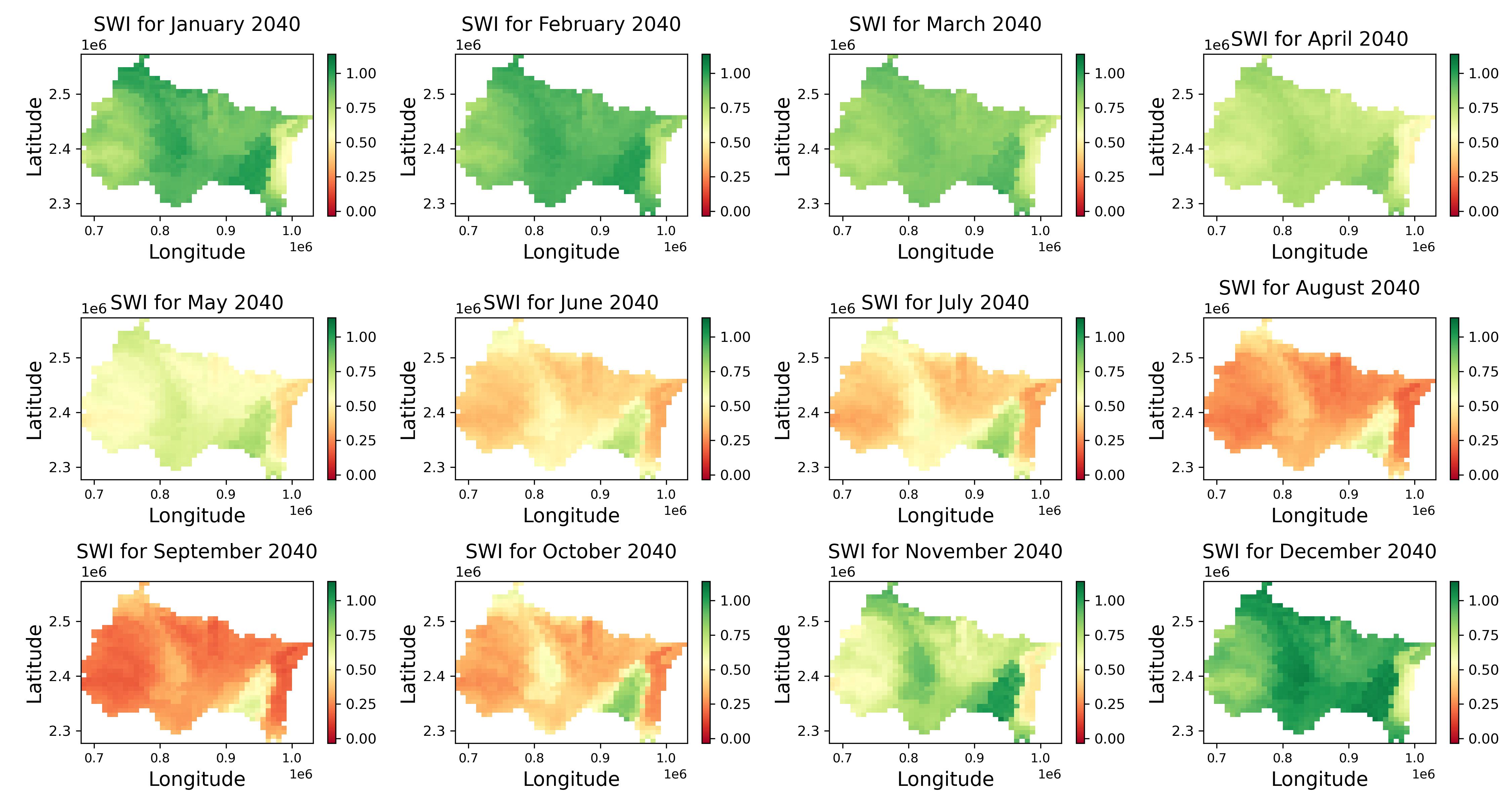}
        \captionsetup{font=small, justification=centering}
        \caption{2040}
    \end{subfigure}
    \hfill
    \\
    \begin{subfigure}{0.735\textwidth}
        \centering
        \includegraphics[width=1\linewidth]{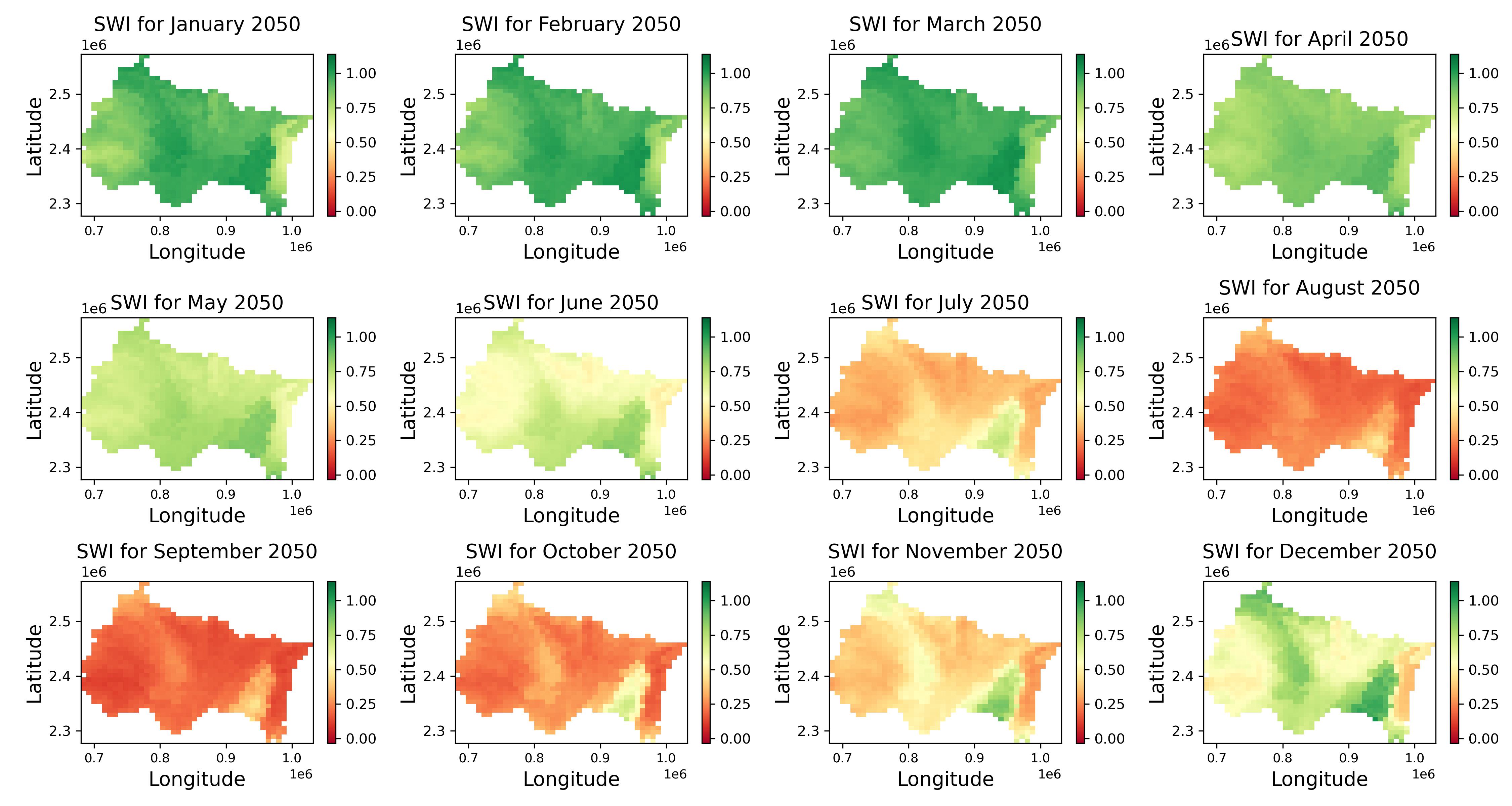}
        \captionsetup{font=small, justification=centering}
        \caption{2050}
    \end{subfigure}
    \caption{Monthly SWI maps generated by SwiGAN under RCP 4.5.}
    \label{fig:swi_maps_month_rcp45}
\end{figure}

The previous results can also be observed using SWI maps generated by SwiGAN. Figures \ref{fig:swi_maps_month_rcp45} and \ref{fig:swi_maps_month_rcp85} (in the appendix) show the generated SWI maps for the average trajectory under RCP 4.5 and RCP 8.5, respectively. From these maps, we observe a general decrease in soil wetness for all months (both wet and dry) between 2025 and 2050. This decrease in soil wetness is more rapid and pronounced under RCP 8.5 than under RCP 4.5. The fact that weather evolutions observed between 2005 and 2024 appear closer to the RCP 8.5 scenario than to the RCP 4.5 scenario makes these results particularly concerning and raises questions about how this reduction in soil wetness due to global warming will impact future losses related to drought, particularly agricultural losses and damages to buildings caused by drought-induced soil subsidence. These results also raise questions about the future insurability of these risks. One of the aims of this paper is to use the proposed SwiGAN model to provide insurers with a preview of what future losses might look like (at least in terms of trends), in order to help them adjust their risk management strategies. This is the motivation behind the application to soil subsidence developed in Section \ref{sec:soil_subsidence}.

\subsection{Variable importance} \label{sec:var_importance}

\begin{figure}[h!]
    \centering
    \includegraphics[width=0.7\linewidth]{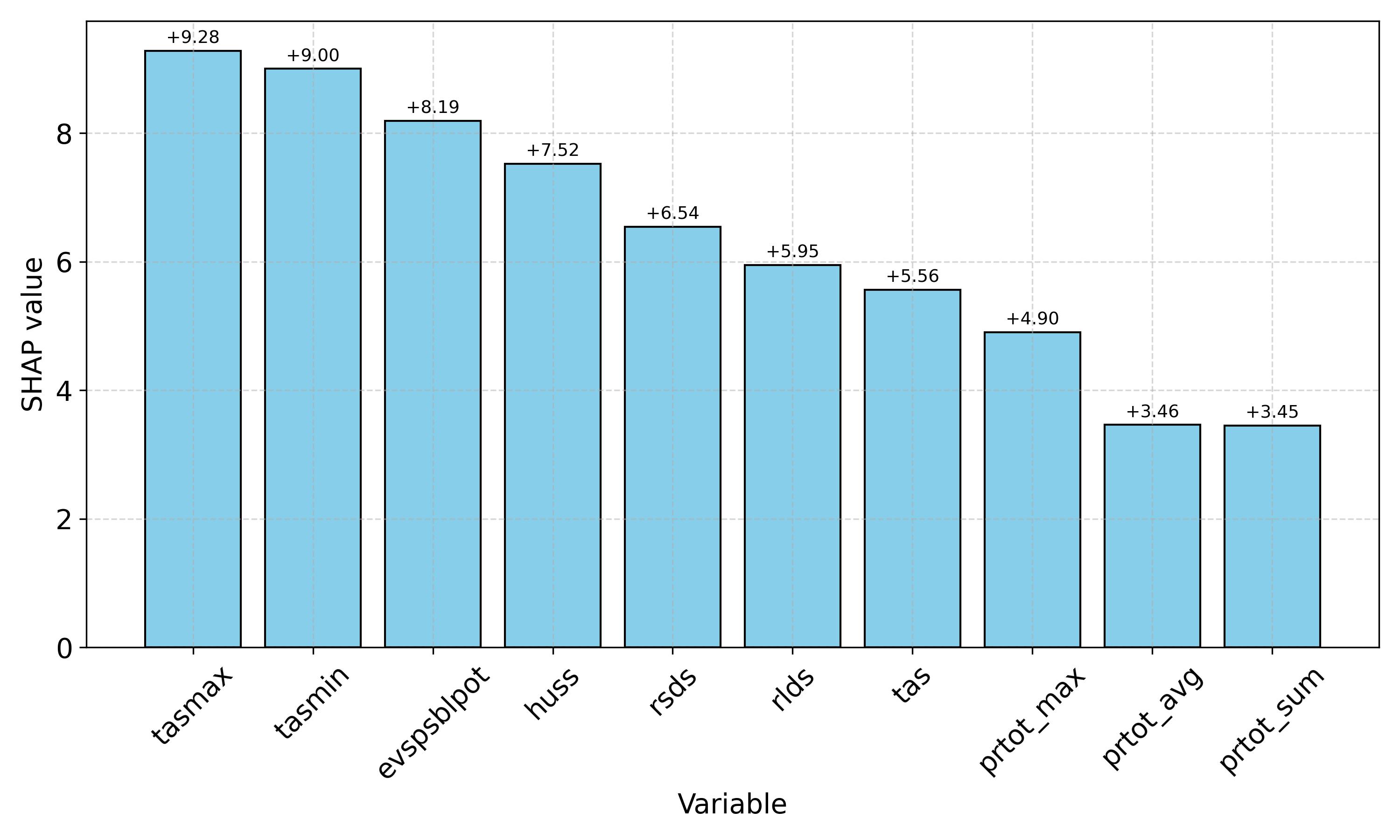}
    \caption{Shapley values used to assess the importance of covariates in explaining the variability of SWI in our SwiGAN model. For more details on Shapley values, see Section \ref{subsec:var_importance_sup}.}
    \label{fig:var_importance}
\end{figure}

To assess the importance of each covariate in explaining the variability of SWI in the proposed SwiGAN model, we use the Shapley additive explanations method. Theoretical details on this method are provided in Section \ref{subsec:var_importance_sup} in the appendix. The final Shapley values presented in Figure \ref{fig:var_importance} are obtained by averaging the values computed for all trajectories over the test set. Details on the covariates presented in this figure are given in Table \ref{tab:covariates}. Note that \texttt{prtot\_max}, \texttt{prtot\_avg}, and \texttt{prtot\_sum} represent the maximum, average, and total monthly precipitation, respectively.

This analysis reveals that temperature and evapotranspiration have the greatest impact on the variability of SWI. This result is expected and is consistent with the conclusions of numerous studies on soil moisture and drought (\cite{ferchichi2024spatio}, \cite{xue2025study}, \cite{li2025analysis}). According to these studies, the higher the level of evapotranspiration, the greater the amount of water lost by the soil, and consequently, the lower the level of soil moisture. The cycle of soil water accumulation through precipitation followed by water loss through evapotranspiration is driven by temperature and other variables such as humidity and wind speed. Indeed, high temperatures accelerate the soil water cycle by increasing the rate at which soils lose water through evapotranspiration. Therefore, high temperatures during a given month, or during the preceding months, lead to lower soil water content and thus lower SWI values for the considered month. 

Precipitation was expected to be among the most influential variables, as it represents the replenishment of water lost by soils through evapotranspiration and deep infiltration. However, our results suggest a different outcome, which can be explained by the inclusion of past SWI values among the covariates during the training process. Indeed, the SWI value of the month immediately preceding the considered month contains information about the ``memory" of the soil, i.e., the amount of water accumulated due to precipitation over previous months. This accumulated precipitation, captured through the SWI of the previous month, appears to be more informative than the precipitation occurring during the current month. Further analysis of variable importance confirms the strong contribution of the SWI of the previous month in explaining the variability of SWI in a given month.

Moreover, \cite{xue2025study} highlight a non-linear and sometimes unexpected relationship between soil moisture and precipitation. They argue that this may be due to geographical location, precipitation seasonality, and temperature fluctuations, which can lead to asynchronous variations between soil moisture and precipitation. The authors also emphasize the impact of long-term precipitation behavior on current soil moisture dynamics. In particular, long-term changes in precipitation and temperature can affect vegetation and surface properties, thereby altering soil water storage capacity.

Despite these observations, it is important to note that the differences in explanatory power between the variables are not very large. This suggests that the contribution of each variable is significant and plays an essential role in ensuring the performance presented in Section \ref{sec:model_validation}.

\begin{figure}[h!]
    \centering
    \includegraphics[width=1\linewidth]{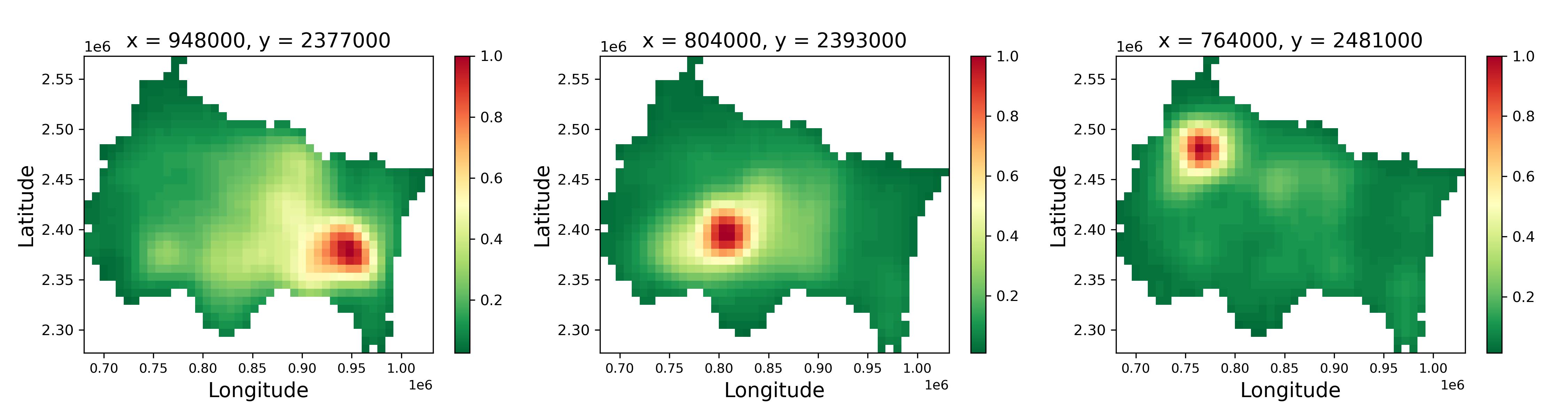}
    \caption{Spatial importance of pixels in the prediction of SWI values for three selected pixels. The first pixel (from left to right) is located in the Vosges and Ardennes mountain ranges, the second in a forest area, and the third in an urban area (see panel (b) of Figure \ref{fig:grand_est}).}
    \label{fig:spatial_importance}
\end{figure}

To gain further insight into how the SwiGAN model generates SWI trajectories, we study spatial importance for a given predicted map. Specifically, we aim to determine, for a given predicted pixel, which regions of the maps are most influential in the model's output for that pixel. To this end, we compute Shapley values for each pixel with respect to the prediction of the pixel of interest and average each pixel's contribution over the entire test period and across all covariates. We then calculate the proportion of these contributions relative to that of the pixel of interest. Figure \ref{fig:spatial_importance} shows the spatial importance (or dependence) plots at three different positions. We observe that the model is able to capture spatial information from a relatively large region surrounding the predicted pixel. This confirms our intuition that the model successfully captures spatial dependencies in the data. This avoids situations where the model generates outputs in which two neighboring geographical areas exhibit drastically different drought behaviors, which would be unrealistic. 

Moreover, in terms of spatial dependence, we observe that the model adapts to the nature of the geographical location being predicted. Indeed, the first map (from left to right) in Figure \ref{fig:spatial_importance} shows the spatial importance for a pixel located in the Vosges and Ardennes mountain ranges. For this pixel, the model, having difficulties in capturing relevant information in the immediate vicinity (as discussed in Section \ref{sec:model_validation}), instead extracts information from a wider geographical area. This behavior is not observed in the third figure, where the pixel of interest is located in an urban area.

It is worth noting that CNNs are highly effective at capturing local spatial patterns in natural images, but they often struggle to detect more global features. In our model design, we introduced attention-based layers to enhance the global sensitivity (see \cite{roy2018concurrentspatialchannelsqueeze} and \cite{woo2018cbamconvolutionalblockattention}) of the CNN components of SwiGAN. This enables the model to extract as much relevant information as possible from the pixels surrounding the pixel of interest, thereby improving both model accuracy and the quality of the generated outputs. This design also improves the geographical interpretability of the results.

\section{Application to drought-induced soil subsidence and insurance losses} \label{sec:soil_subsidence}

This section presents an application of SwiGAN to the estimation of future drought-related insurance costs associated with soil subsidence. Our approach builds on the work of \citet{heranval2023application}, who propose statistical and machine learning models to predict drought costs in France based on climatic indicators and exposure variables.

Section \ref{subsec:SWI} describes how the SWI is used to determine NatCat eligibility. This is a specific feature of the French public-private protection system: a certain number of criteria must be met to trigger eligibility and compensation. In the absence of a real database on damages, we explain the proxy used to estimate the cost of a given drought event for a city in Section \ref{subsec:houses}. Finally, a discussion on the evolution of insurance losses related to drought-induced soil subsidence is provided in Section \ref{subsec:discussion}.

\subsection{Using SWI to determine areas potentially damaged by drought-induced soil subsidence}
\label{subsec:SWI}

When a commune (an administrative unit in France) is affected by damages due to drought-induced soil subsidence, local authorities can submit a request for the recognition of a state of natural catastrophe (NatCat recognition) to the CatNat regime. This status is granted or denied based on a thorough analysis of SWI values by the regime. Compensation by insurance companies can only begin once the state of natural catastrophe has been officially recognized.

France's CatNat regime uses a three-step methodology to determine whether a commune is eligible for the recognition of a state of natural catastrophe caused by drought-induced soil subsidence for a given year. The first step consists of aggregating SWI values for each SAFRAN pixel from monthly to yearly values by considering the driest month. Then, return periods of these yearly SWI values are used to identify abnormal pixels for that year. Finally, the pixels within the commune of interest, as well as the characteristics of surrounding communes, are analyzed to determine eligibility. More details about this methodology are available on the Météo-France website\footnote{\url{https://meteofrance.fr/sites/meteofrance.fr/files/files/editorial/brochure-catnat-meteo-france.pdf}}.

\begin{figure}[h!]
    \centering
    \includegraphics[width=1\linewidth]{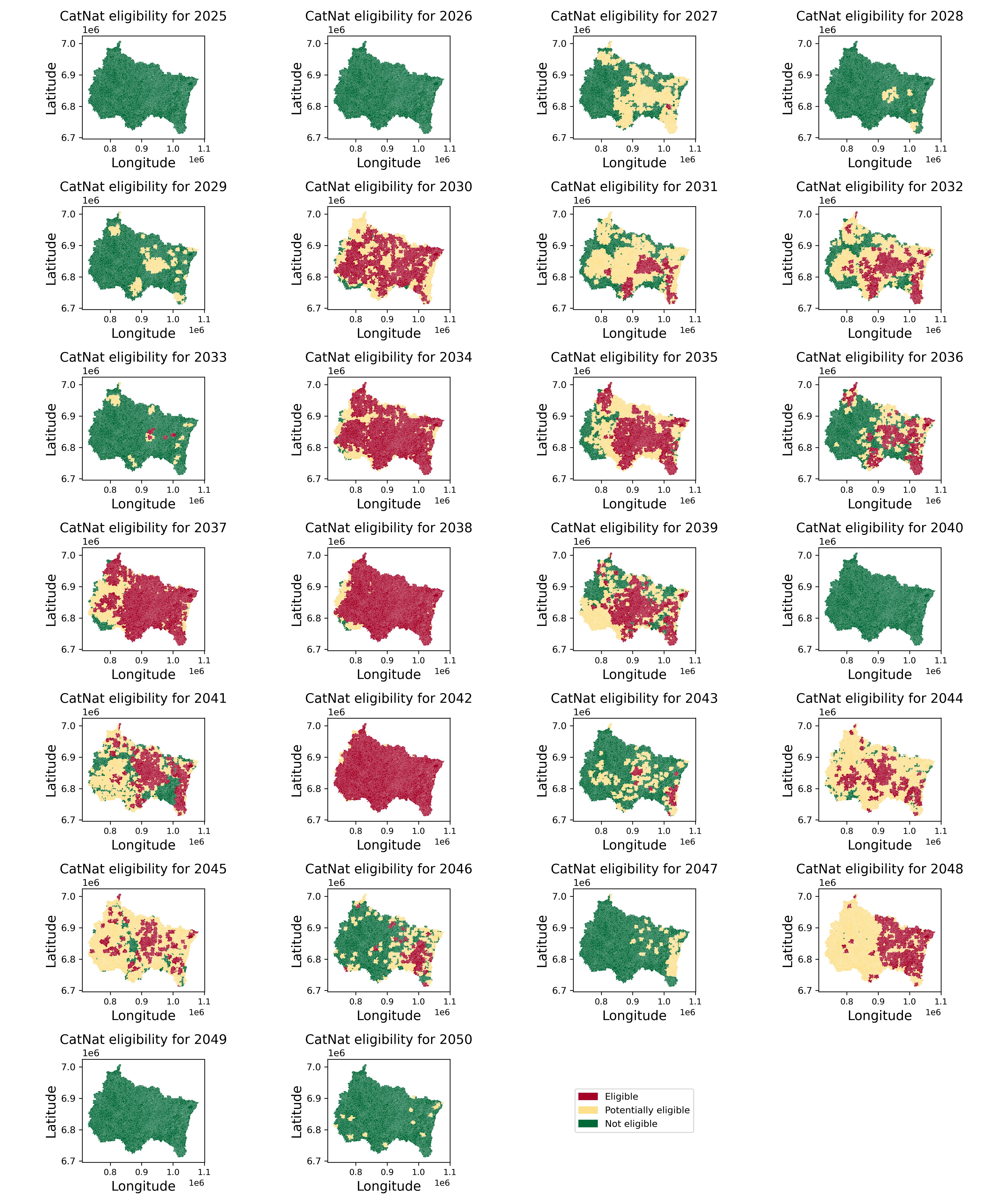}
    \caption{Evolution of commune eligibility for the recognition of a state of natural catastrophe due to drought-induced soil subsidence under RCP 4.5.}
    \label{fig:evol_recogn_rcp45}
\end{figure}

We reproduce the same methodology, with the aim of studying the evolution of damages over the period 2025 to 2050. We use the SWI values generated by SwiGAN as inputs to this methodology. Figure \ref{fig:evol_recogn_rcp45} shows the evolution of commune eligibility over the considered period under the RCP 4.5 scenario. Figure \ref{fig:evol_recogn_rcp85} in the appendix presents the same evolution for the RCP 8.5 scenario. The eligibility decisions displayed in these figures correspond to the generated SWI trajectories with the lowest SWI values.

In these figures, potentially eligible communes refer to those that are not officially eligible based on the methodology described above but are surrounded by eligible communes and could be declared eligible after deliberation by the CatNat regime. In the remainder of this section, we focus on eligible communes to study the evolution of future damages caused by drought-induced soil subsidence. To further validate the SwiGAN model, we evaluate its ability to identify communes that were eligible during the test period of our study. Figure \ref{fig:accuracy_catnat} in the appendix shows that, over the test period, SwiGAN correctly identifies eligible communes in about 90\% of cases and correctly distinguishes between eligible and non-eligible communes in about 70\% of cases. This confirms the conclusions of Section \ref{sec:model_validation} regarding the accuracy of SwiGAN. The error rate of approximately 30\% in this identification task can be explained by the fact that the criteria used to identify eligible communes were slightly modified during the test period, changes that were not accounted for in our study due to the lack of available information.

For both the RCP 4.5 and RCP 8.5 scenarios, the figures show an overall increase in the number of communes impacted by drought-induced soil subsidence and thus eligible for recognition by the CatNat regime. These results are consistent with those obtained from the analysis of generated SWI values in Section \ref{sec:results_analysis}. We also observe that even under the RCP 4.5 scenario, which is considered less extreme, a substantial number of communes become affected by drought-induced soil subsidence over time. Our conclusions are consistent with those of \cite{barthelemy2025analysis}, whose study analyzes past and future droughts leading to soil subsidence in France. 

\subsection{From number of exposed houses to financial costs}
\label{subsec:houses}

Once a commune receives recognition of a state of natural catastrophe due to drought from the CatNat regime, all buildings located in areas of this commune that contain clay soils become eligible for insurance compensation. Insurance companies then intervene to compensate damaged buildings depending on the nature of the insurance contracts. To estimate the maximum number of damaged buildings, and hence the exposure of insurers in the Grand Est region, we use data on the 2D and 3D modeling of the French territory and its infrastructures\footnote{The database is available on this website: \url{https://geoservices.ign.fr/telechargement-api/BDTOPO?zone=R44&format=SHP}} to count the number of residential buildings located in areas with clay soils in each commune of the Grand Est region of France.

Having obtained the maximum number of damaged buildings in each commune eligible for compensation following the CatNat regime methodology, we adapt a formula proposed by \cite{heranval2023application} to estimate the financial costs of damages due to drought-induced soil subsidence in the Grand Est region. This formula is given by: 

\begin{equation}
\label{eq:cost_buildings}
    C(x)=a+bx-be^{-kx}
\end{equation}

where $x$ is the maximum number of damaged buildings, $C$ is the maximum financial cost of damages in euros, $a = 464.4$ and $b = 4.121 \times 10^8$ are parameters calibrated by \cite{heranval2023application} for the whole of France, and $k=-\frac{a}{5b}$. The initial function proposed by \cite{heranval2023application} was $C_0(x)=a+bx$. We add the term $-be^{-kx}$ to eliminate negative costs while ensuring that asymptotically $C$ converges to $C_0$ when the number of buildings becomes large (which is typically the case).

\begin{figure}[h!]
    \centering
    \includegraphics[width=1\linewidth]{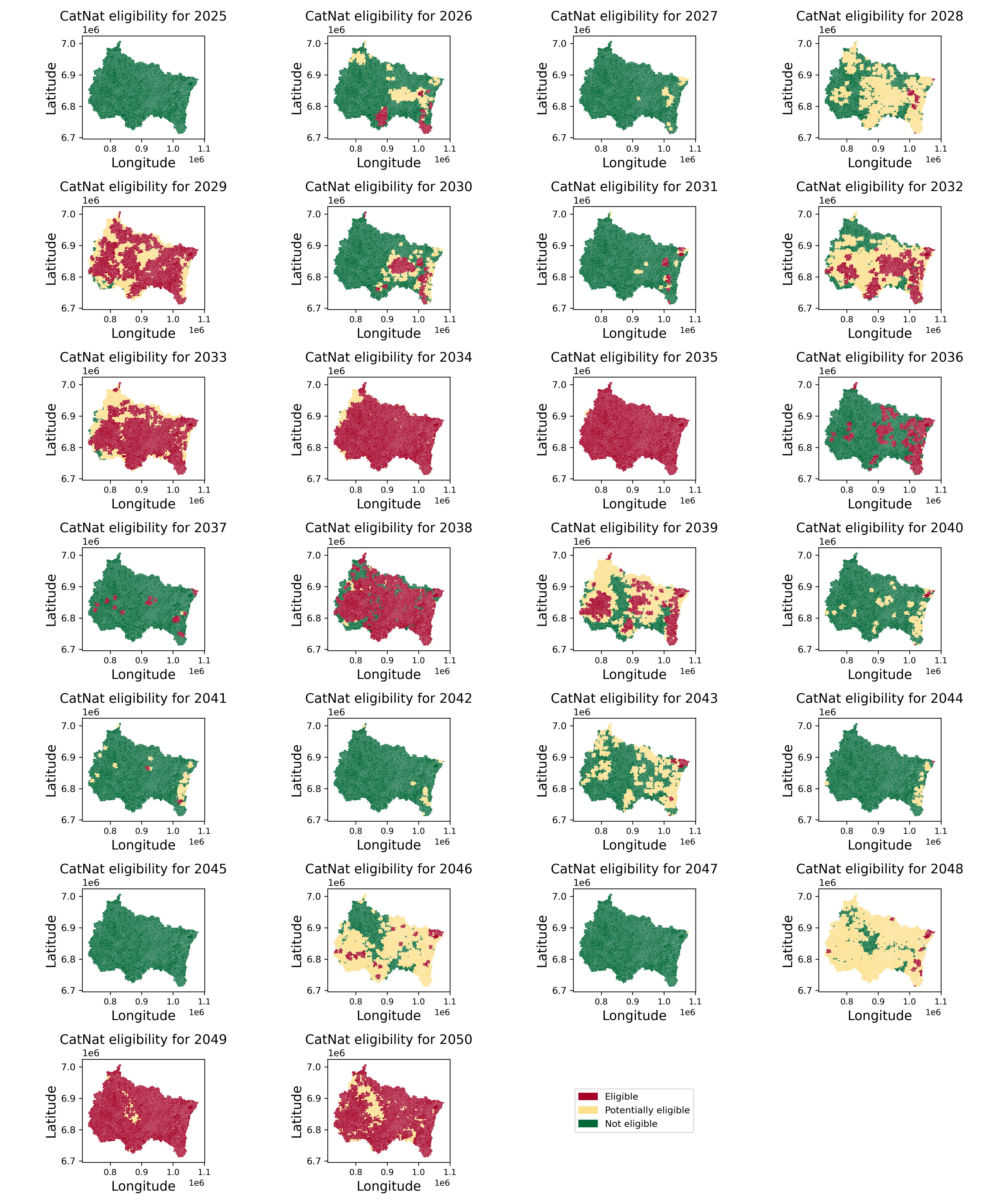}
    \caption{Evolution of commune eligibility for the recognition of a state of natural catastrophe due to drought-induced soil subsidence under RCP 8.5.}
    \label{fig:evol_recogn_rcp85}
\end{figure}

Recall that we generate 1\,000 different trajectories for the evolution of SWI between 2025 and 2050. For a given trajectory, we compute drought-related CatNat eligibility as illustrated in Figures \ref{fig:evol_recogn_rcp45} and \ref{fig:evol_recogn_rcp85}. Using Equation \ref{eq:cost_buildings}, we then compute, for that trajectory, the maximum cost of damages due to drought-induced soil subsidence in the Grand Est region over the period 2025 to 2050. This leads to a distribution of insurance costs illustrated in panel (a) of Figure \ref{fig:costs_rga_rcp45} for the RCP 4.5 scenario and in panel (a) of Figure \ref{fig:costs_rga_rcp85} in the appendix for the RCP 8.5 scenario.

\begin{figure}[h!]
    \centering
    \begin{subfigure}{0.49\textwidth}
        \centering
        \includegraphics[width=1\linewidth]{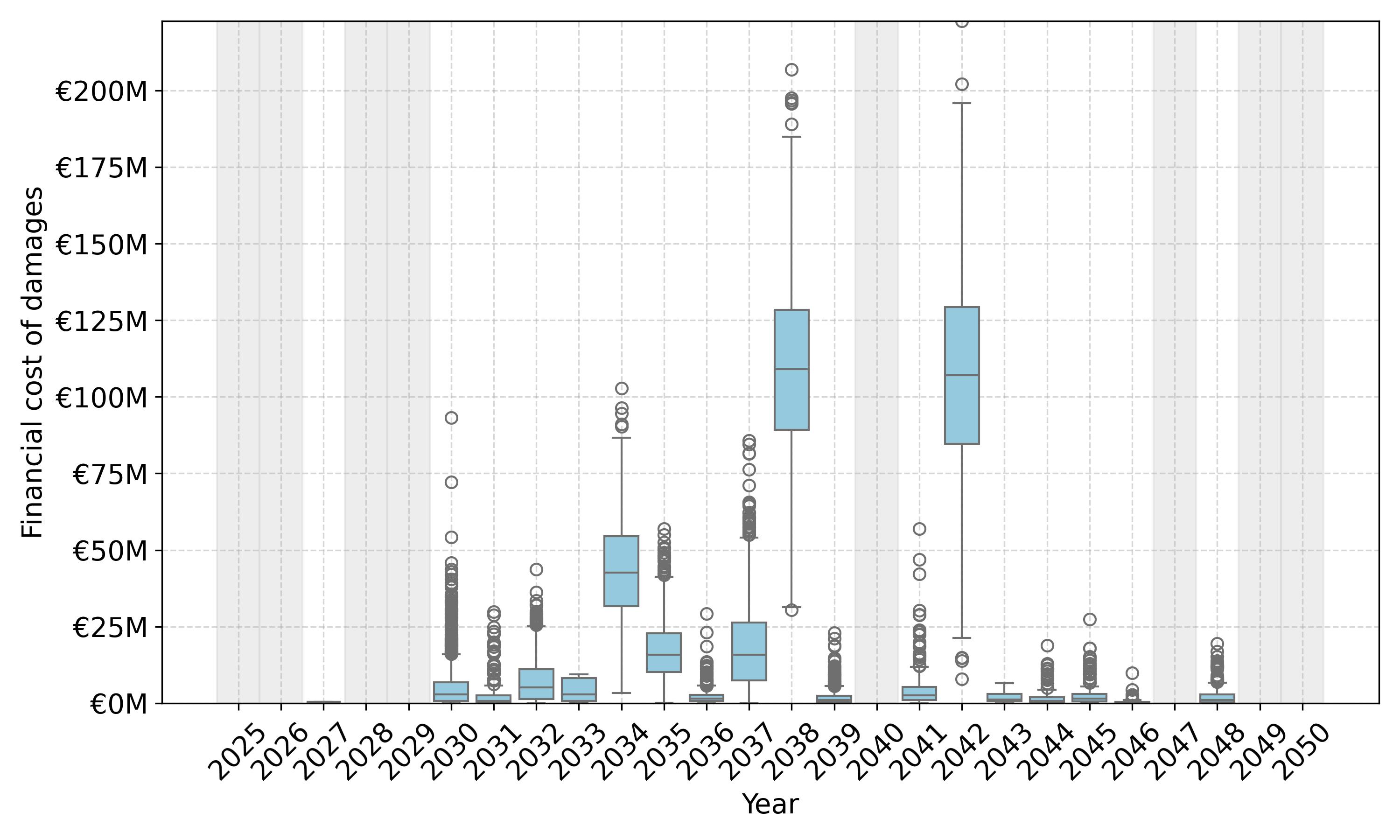}
        \captionsetup{font=small, justification=centering}
        \caption{Insurance damage costs}
    \end{subfigure}
    \begin{subfigure}{0.49\textwidth}
        \centering
        \includegraphics[width=1\linewidth]{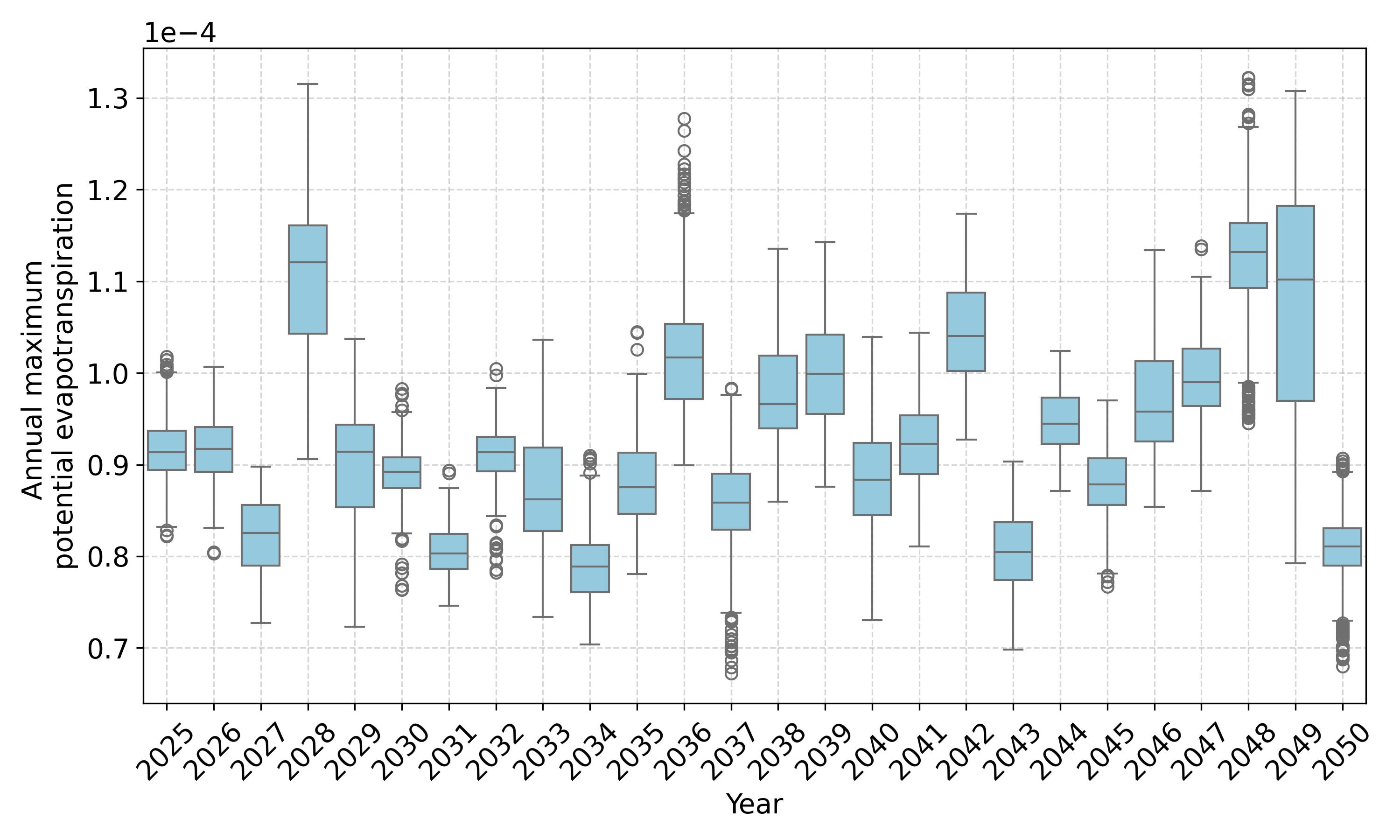}
        \captionsetup{font=small, justification=centering}
        \caption{Potential evapotranspiration}
    \end{subfigure}
    \caption{Future insurance damage costs due to drought-induced soil subsidence under RCP 4.5 (panel (a)). Panel (b) shows the annual maximum potential evapotranspiration, which, as shown in Section \ref{sec:var_importance}, is the covariate with the greatest impact on the variability of SWI.}
    \label{fig:costs_rga_rcp45}
\end{figure}

\subsection{Discussions on the evolution of subsidence}
\label{subsec:discussion}

The results of the previous section should be interpreted with caution. Many factors have not been taken into account when projecting the evolution of subsidence. First, we do not consider changes in exposure. The population of the urban areas considered is assumed to remain fixed in this analysis. An increase in population would likely lead to an increase in the number of exposed buildings. On the other hand, building standards are evolving, meaning that new constructions should (theoretically) be more resilient to damages caused by drought-induced soil subsidence. Second, in this simplified study, we do not account for adaptation strategies. In addition to improvements in new buildings, preventive measures applied to existing structures may contribute to reducing losses. Another limitation is that we do not consider that some affected houses may be removed from the exposure after being impacted, as they could be too damaged to be affected again.

The purpose of the present study is not to analyze these behavioral or structural changes, but rather to demonstrate how the Wasserstein GAN generator developed in this work can be used for risk projection. Regarding Figure \ref{fig:costs_rga_rcp45}, we observe that, under the RCP 4.5 scenario, the evolution of losses is not strictly increasing over time. This is due to the fact that the evolution of SWI depends on several complex factors. For instance, panel (b) of Figure \ref{fig:costs_rga_rcp45} shows that the evolution of potential evapotranspiration—identified as one of the most influential variables affecting SWI variability—is not monotonic. Moreover, as discussed in Section \ref{subsec:SWI}, the relationship between SWI values, eligibility, and financial compensation within the CatNat system is indirect and involves complex, non-linear transformations.

Another important point concerns the order of magnitude of the losses. The Grand Est region in France has historically been less exposed to drought than southern regions. Currently, subsidence accounts for approximately 1 to 1.5 billion euros of losses annually in France. Although no official statistics exist for Grand Est, its relatively lower exposure and demographic characteristics (8\% of the national population, with approximately 30\% living in urban areas where individual housing density is lower) suggest that the current annual burden is on the order of a few tens of millions of euros during severe drought years. In the simulations, peak losses in the worst years exceed 100 million euros, with a Value-at-Risk close to 200 million euros. This trend is particularly concerning and supports the hypothesis of a significant increase in damages due to drought-induced soil subsidence in this region.

Finally, the fact that simulated losses decrease after 2042 should not necessarily be reassuring. The complex criteria governing NatCat attribution in France tend to reduce future eligibility. Indeed, drought conditions are assessed relative to historical benchmarks, and eligibility is determined based on whether a given year is significantly worse than previous ones. As drought events become more frequent and severe, the relative threshold for eligibility may increase, leading to a reduction in the probability of future eligibility and, consequently, a reduction in insured losses from the system’s perspective. However, individual costs for households may remain high, even if they are insufficient for their commune to be declared eligible for compensation. This could ultimately contribute to an increase in the insurance protection gap.

\section{Conclusion}

The Wasserstein GAN methodology introduced in this paper highlights the benefits of such generative techniques for simulating weather indices related to insurance claims. The example developed for the Soil Wetness Index in the context of subsidence can be extended to other quantities (precipitation levels, hail severity indicators, or even complex indices used in index insurance). Since the model learns the impact of the current state of the climate on the index, it enables the projection of risk at a given horizon. Moreover, it can adapt to the assumptions underlying climate scenarios, allowing for an easy comparison of their impact on insurance outcomes. Compared to fully physical models, generative models are computationally efficient in producing simulations, which facilitates the estimation of relevant insurance metrics such as Value-at-Risk. A limitation of the approach lies in the size of the territory considered (here, a French administrative region), which is necessary to reduce dimensionality, limit computational requirements during training, and remain consistent with the relatively small dataset. Regarding the evolution of damages, we conducted a simplified study in the case of subsidence. We emphasize that these projections do not account for adaptation strategies. We believe that the generative methodology developed in this work can serve as a valuable tool for testing the impact of various adaptation strategies and for enhancing prevention efforts in the field of natural disasters.

\textbf{Acknowledgment: } Olivier Lopez acknowledges funding from the Excellence Chair CARE (Allianz, Ensae, Risk Fundation). Antoine Heranval acknowledge the support of the Chaire Geolearning funded by Andra, BNP-Paribas, CCR and SCOR Foundation.

\section{Appendix} \label{sec:appendix}

\subsection{Additional illustrations of the data and results}

\begin{figure}[h!]
    \centering
    \includegraphics[width=1\linewidth]{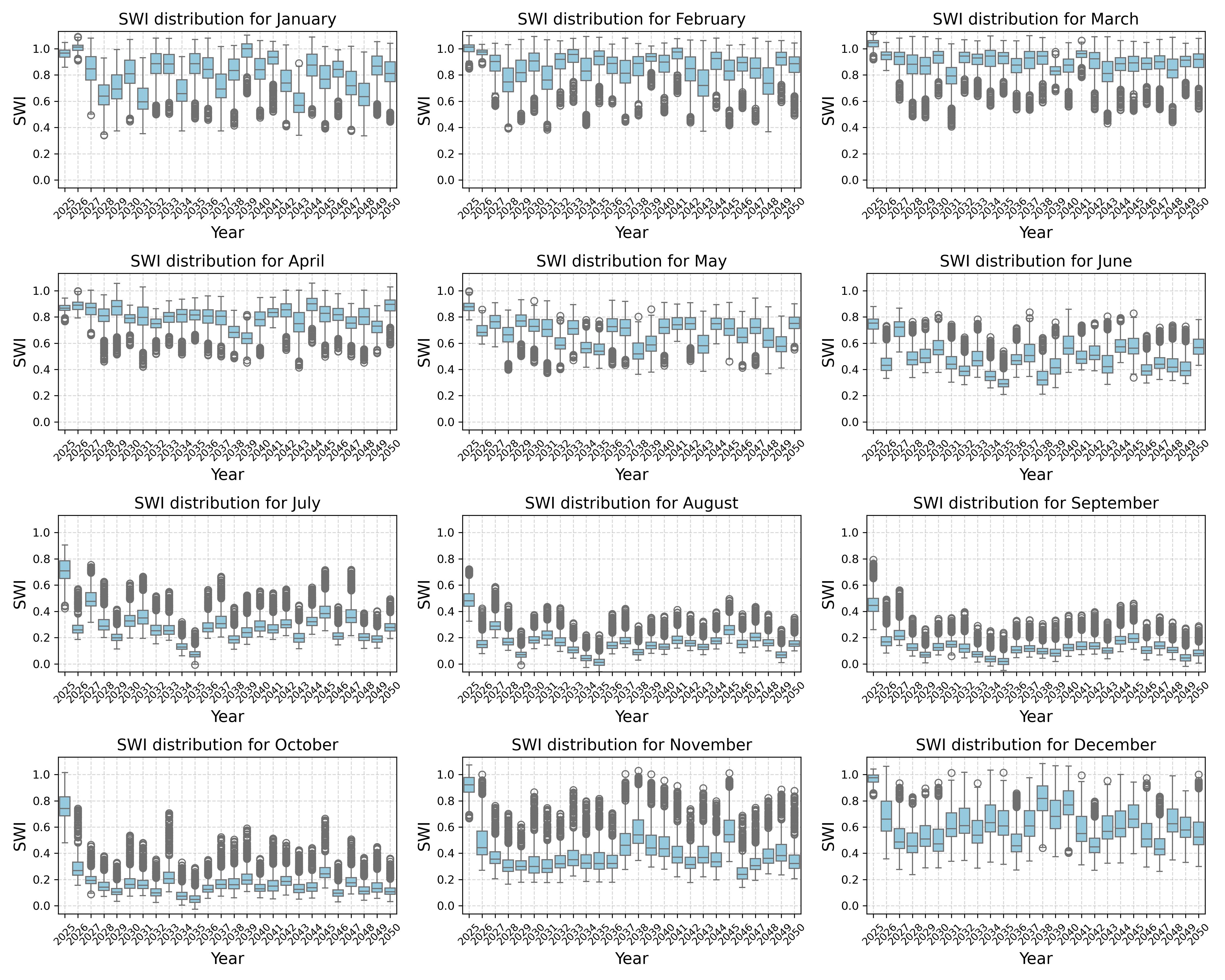}
    \caption{Spatial distribution of the mean trajectory of SWI values generated under RCP 8.5 by SwiGAN.}
    \label{fig:dist_swi_rcp85}
\end{figure}

\begin{figure}[h!]
    \centering
    \begin{subfigure}{0.737\textwidth}
        \centering
        \includegraphics[width=1\linewidth]{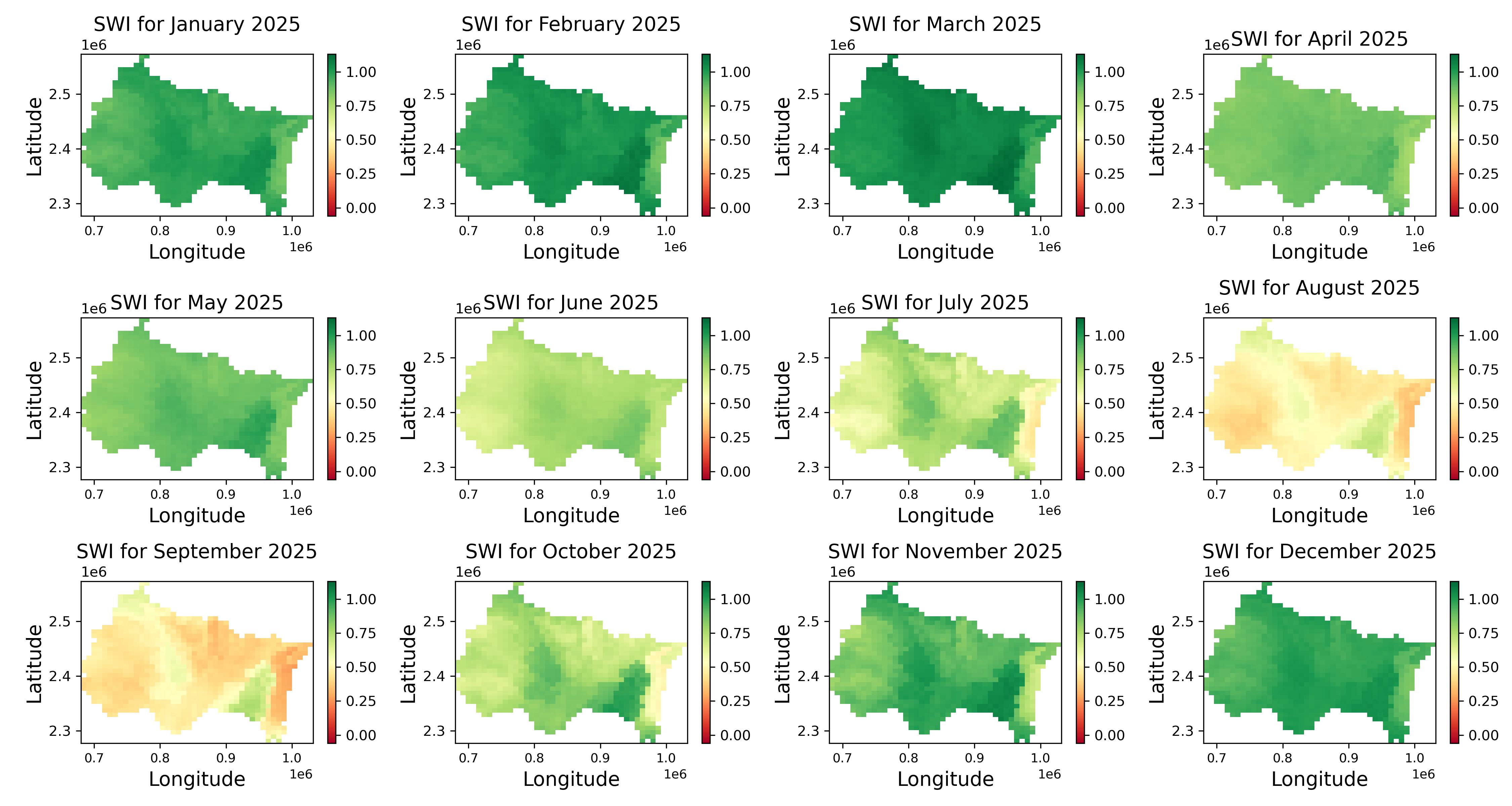}
        \captionsetup{font=small, justification=centering}
        \caption{2025}
    \end{subfigure}
    \hfill
    \\
    \begin{subfigure}{0.737\textwidth}
        \centering
        \includegraphics[width=1\linewidth]{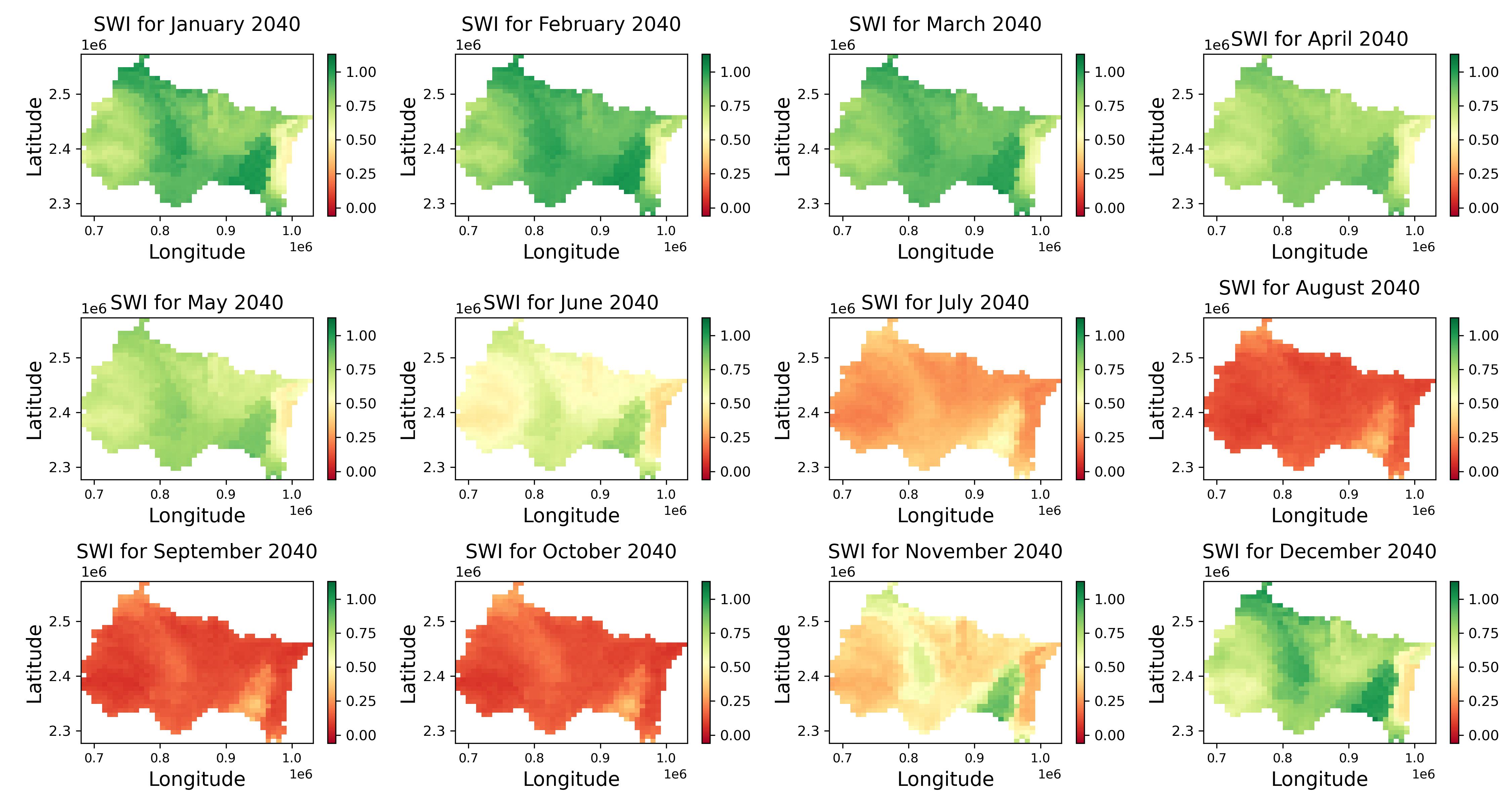}
        \captionsetup{font=small, justification=centering}
        \caption{2040}
    \end{subfigure}
    \hfill
    \\
    \begin{subfigure}{0.737\textwidth}
        \centering
        \includegraphics[width=1\linewidth]{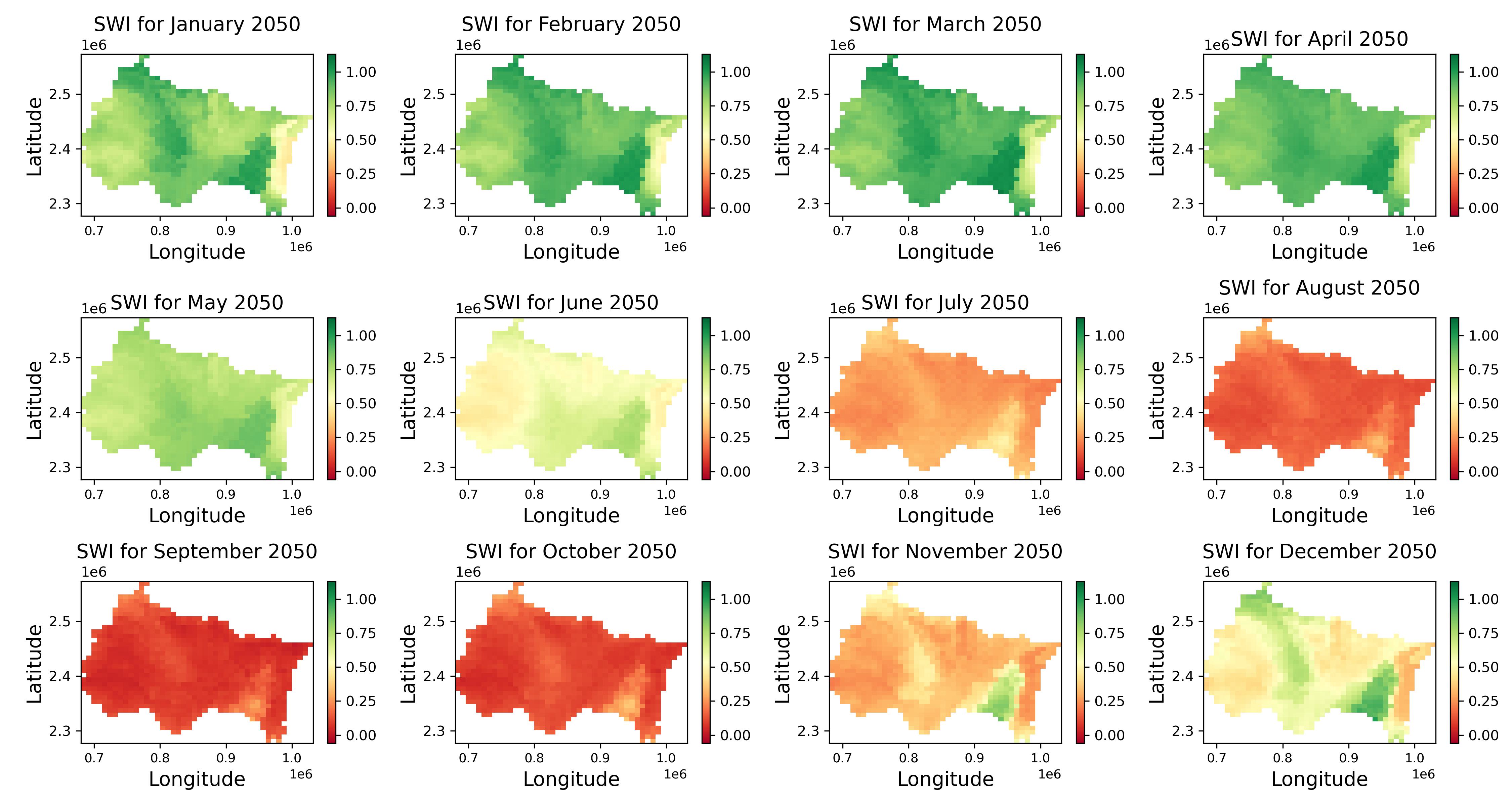}
        \captionsetup{font=small, justification=centering}
        \caption{2050}
    \end{subfigure}
    \caption{Monthly SWI maps generated by SwiGAN under RCP 8.5.}
    \label{fig:swi_maps_month_rcp85}
\end{figure}

\begin{figure}[h!]
    \centering
    \includegraphics[width=0.7\linewidth]{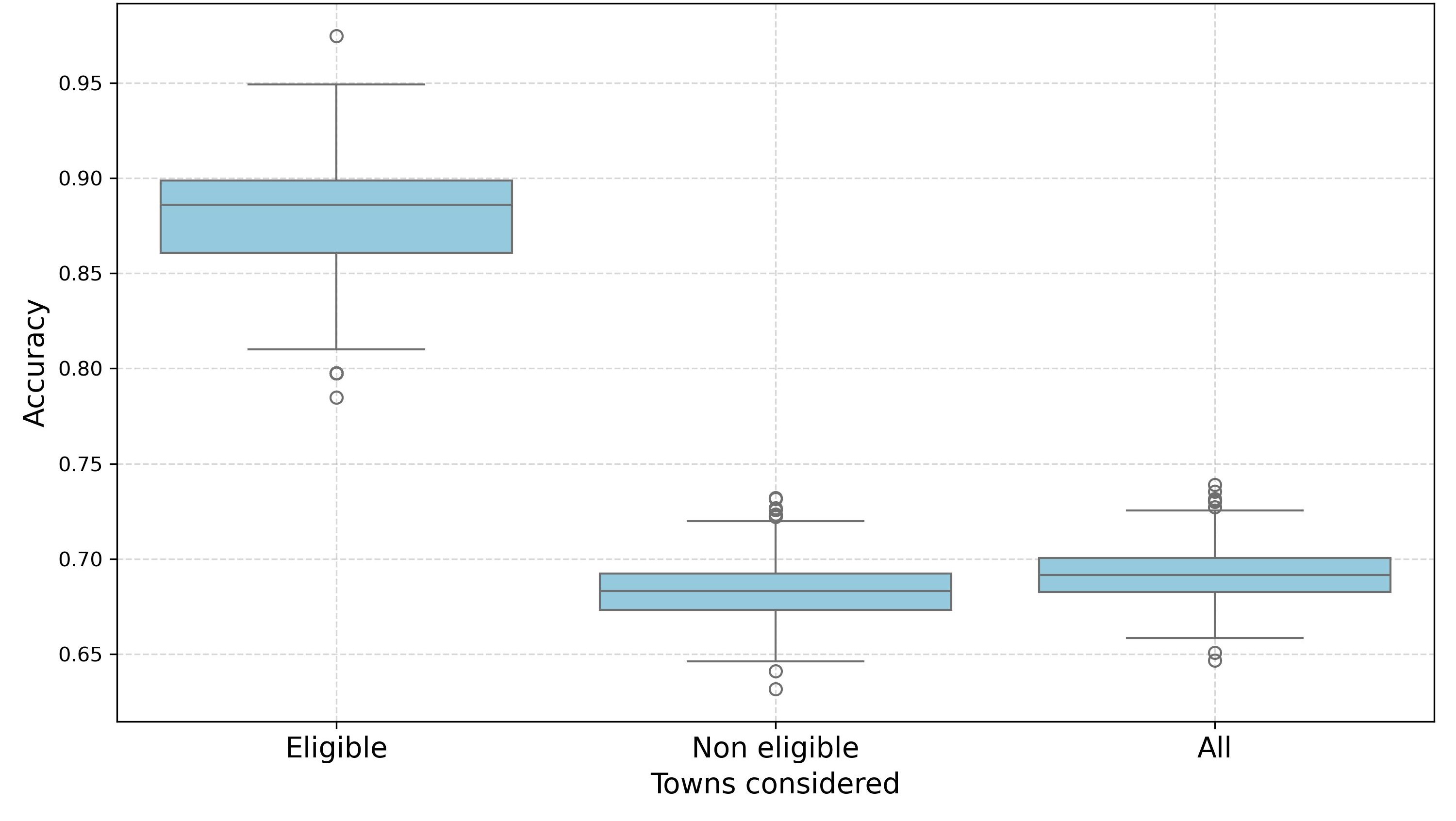}
    \caption{Accuracy of SwiGAN in predicting commune eligibility for the recognition of a state of natural disaster due to drought by the CatNat regime over the test period.}
    \label{fig:accuracy_catnat}
\end{figure}

\begin{figure}[h!]
    \centering
    \begin{subfigure}{0.49\textwidth}
        \centering
        \includegraphics[width=1\linewidth]{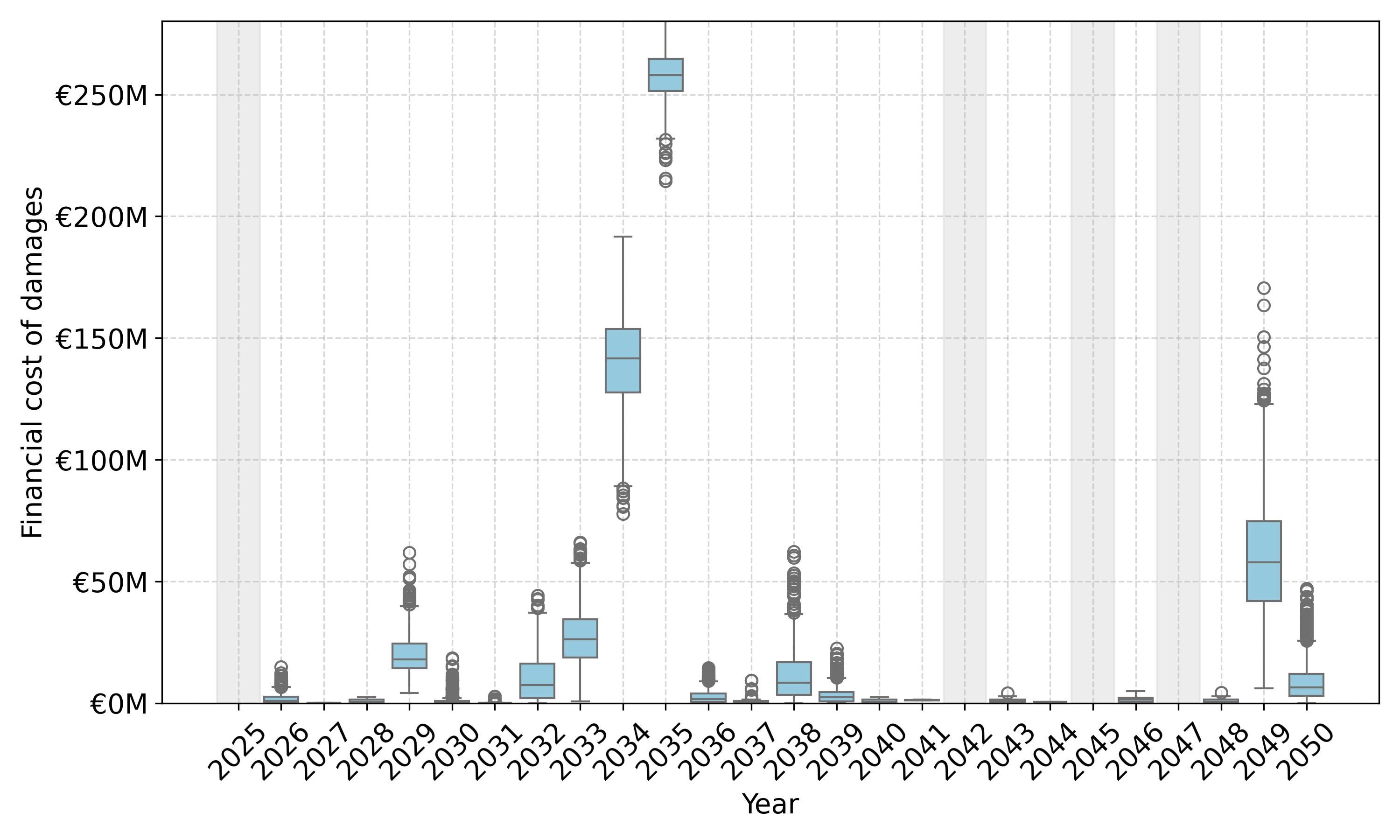}
        \captionsetup{font=small, justification=centering}
        \caption{Insurance damage costs}
    \end{subfigure}
    \begin{subfigure}{0.49\textwidth}
        \centering
        \includegraphics[width=1\linewidth]{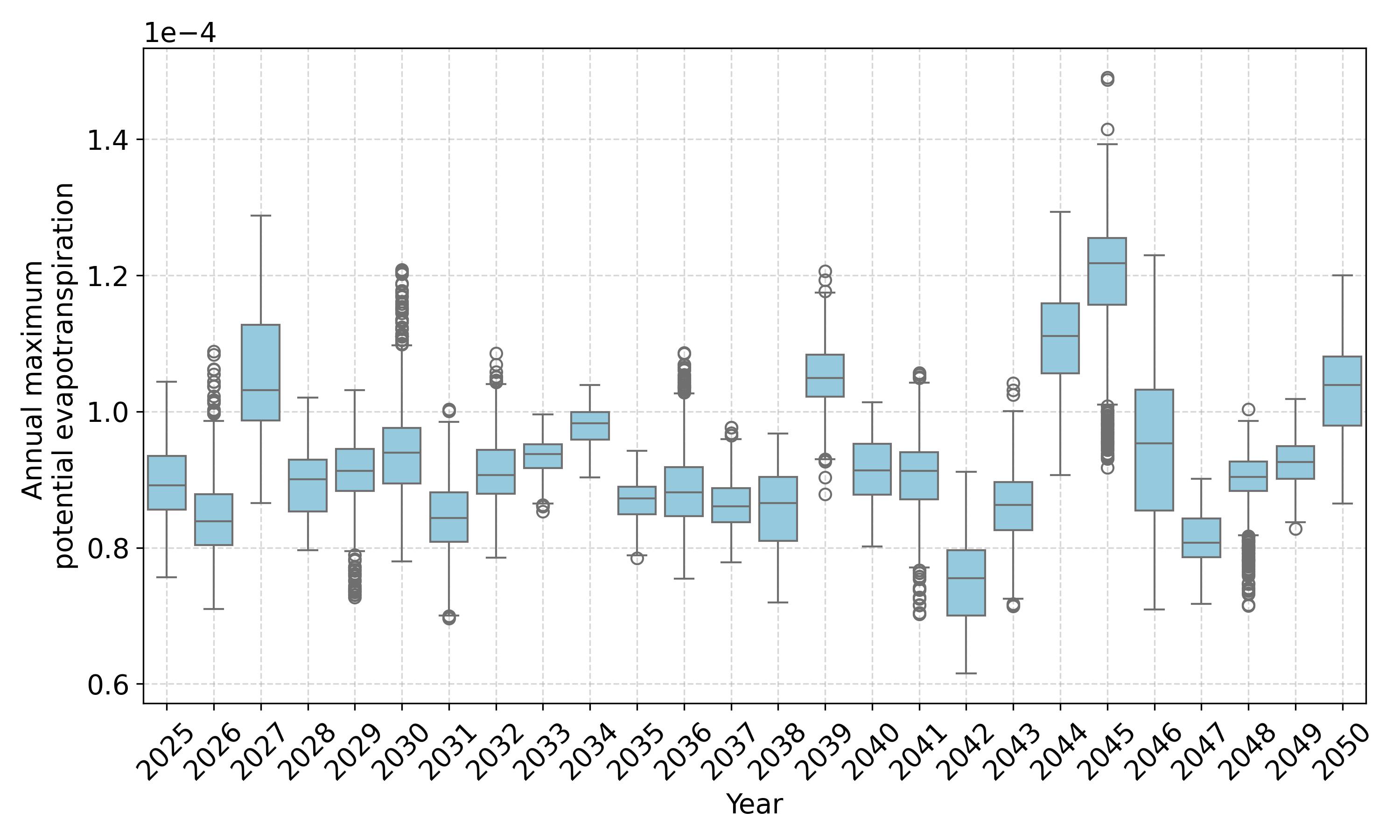}
        \captionsetup{font=small, justification=centering}
        \caption{Potential evapotranspiration}
    \end{subfigure}
    \caption{Future insurance damage costs due to drought-induced soil subsidence under RCP 8.5 (panel (a)). Panel (b) shows the annual maximum potential evapotranspiration, which, as shown in Section \ref{sec:var_importance}, is the covariate with the greatest impact on the variability of SWI.}
    \label{fig:costs_rga_rcp85}
\end{figure}

\begin{figure}
    \centering
    \includegraphics[width=1\textwidth]{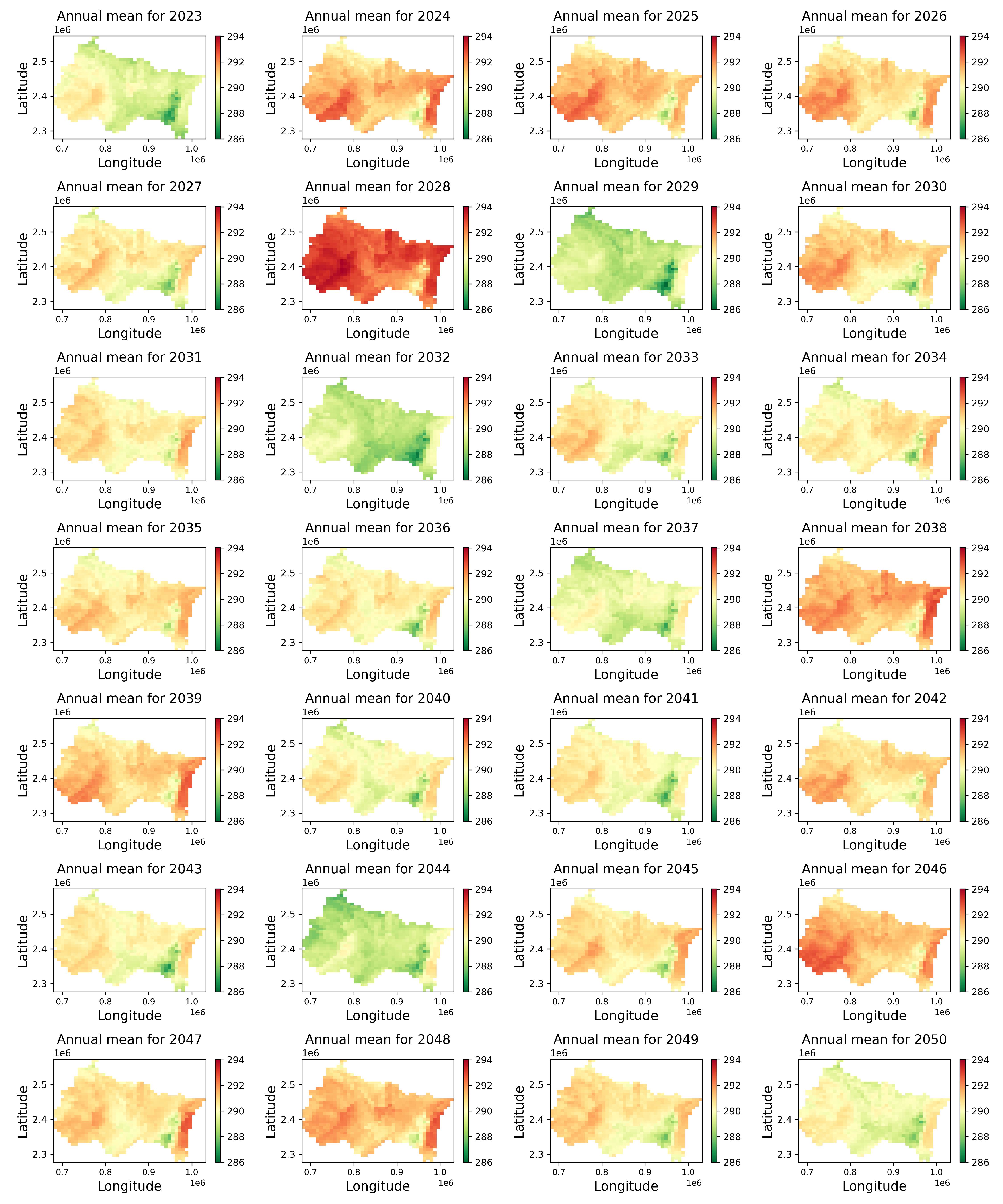}
    \caption{Evolution of mean annual temperature in the Grand Est region under the RCP 4.5 scenario.}
    \label{fig:temp_rcp45}
\end{figure}

\begin{figure}
    \centering
    \includegraphics[width=1\textwidth]{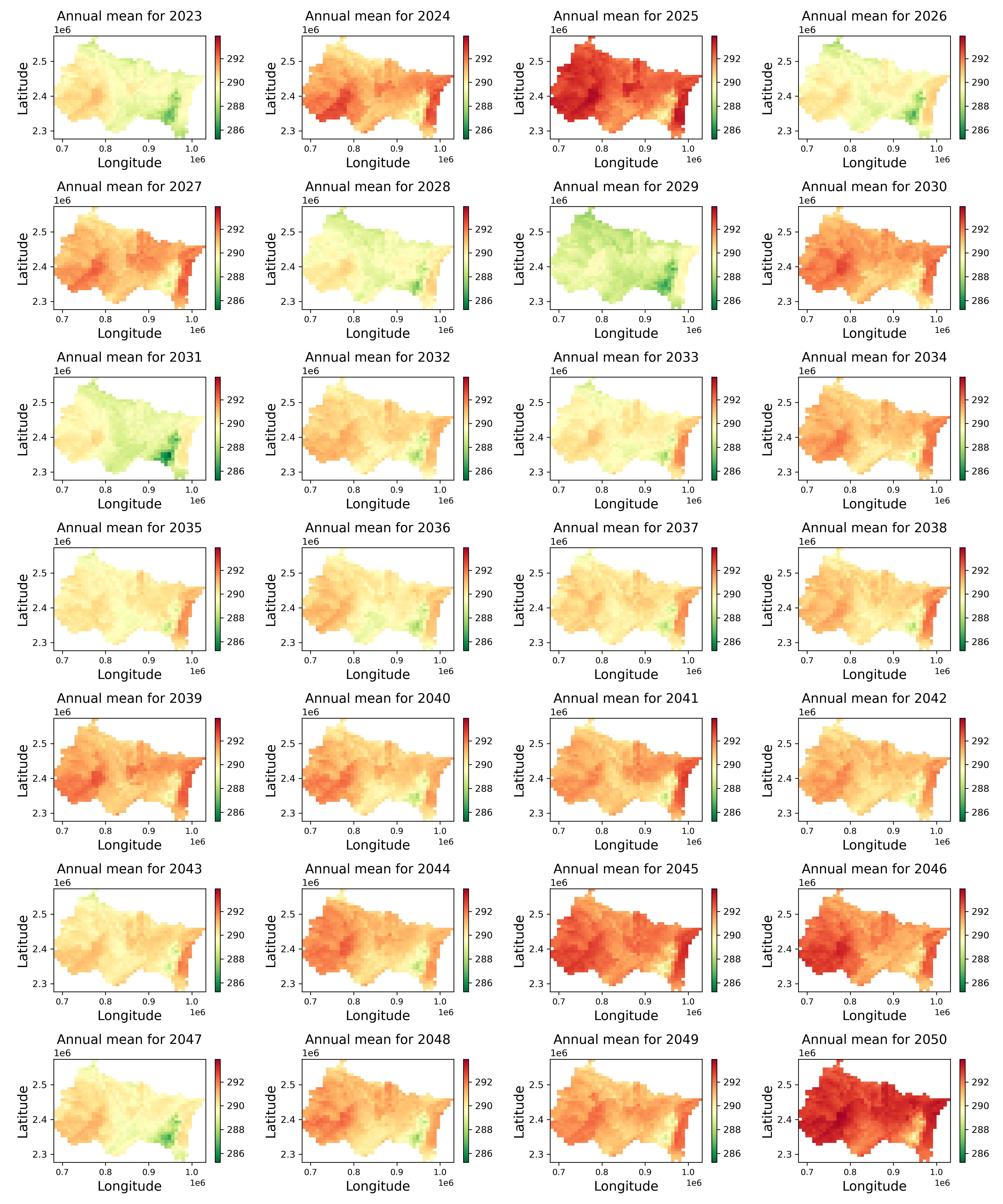}
    \caption{Evolution of mean annual temperature in the Grand Est region under the RCP 8.5 scenario.}
    \label{fig:temp_rcp85}
\end{figure}

\begin{figure}
    \centering
    \includegraphics[width=1\textwidth]{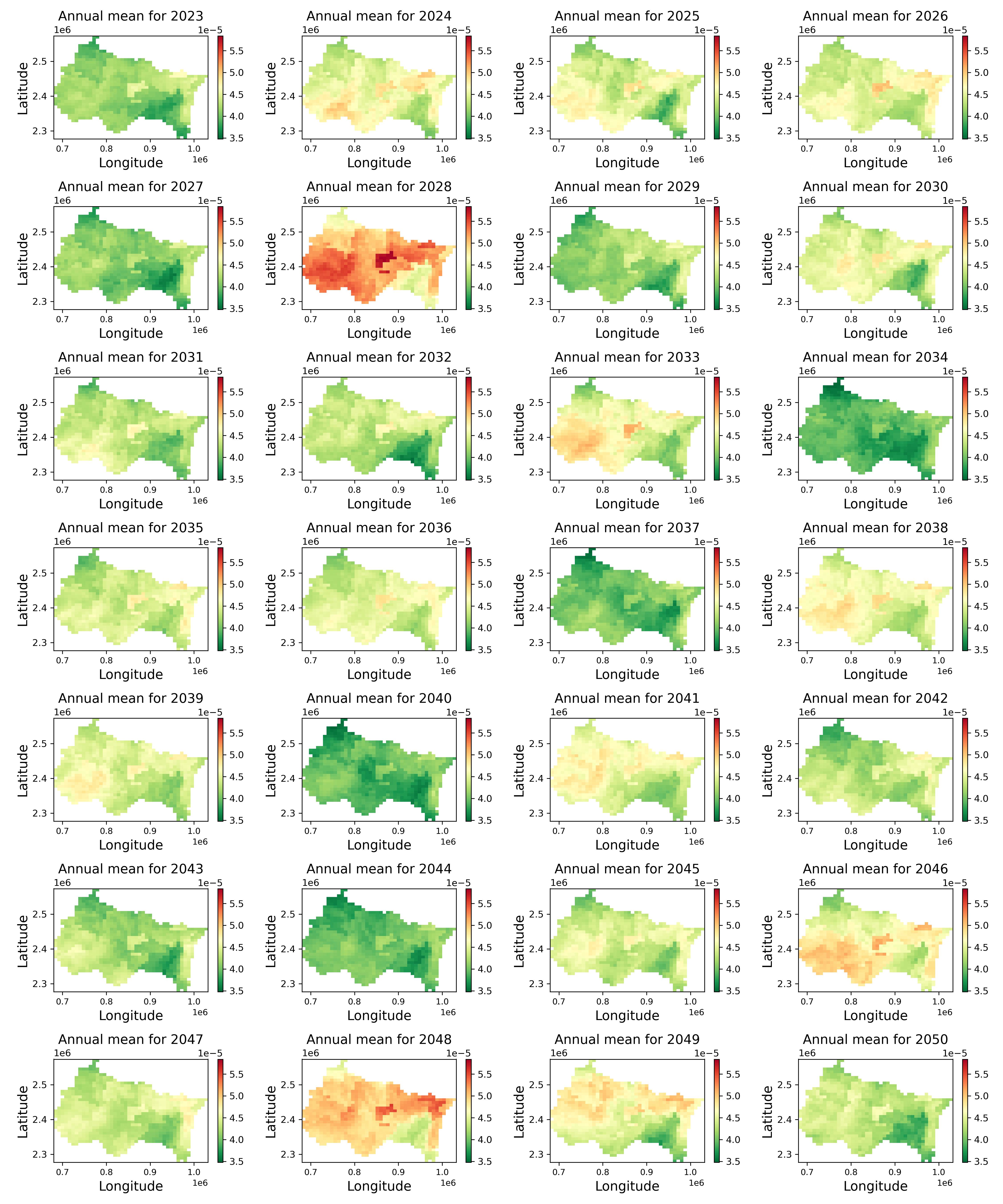}
    \caption{Evolution of mean annual evapotranspiration in the Grand Est region under the RCP 4.5 scenario.}
    \label{fig:evapotranspi_rcp45}
\end{figure}

\begin{figure}
    \centering
    \includegraphics[width=1\textwidth]{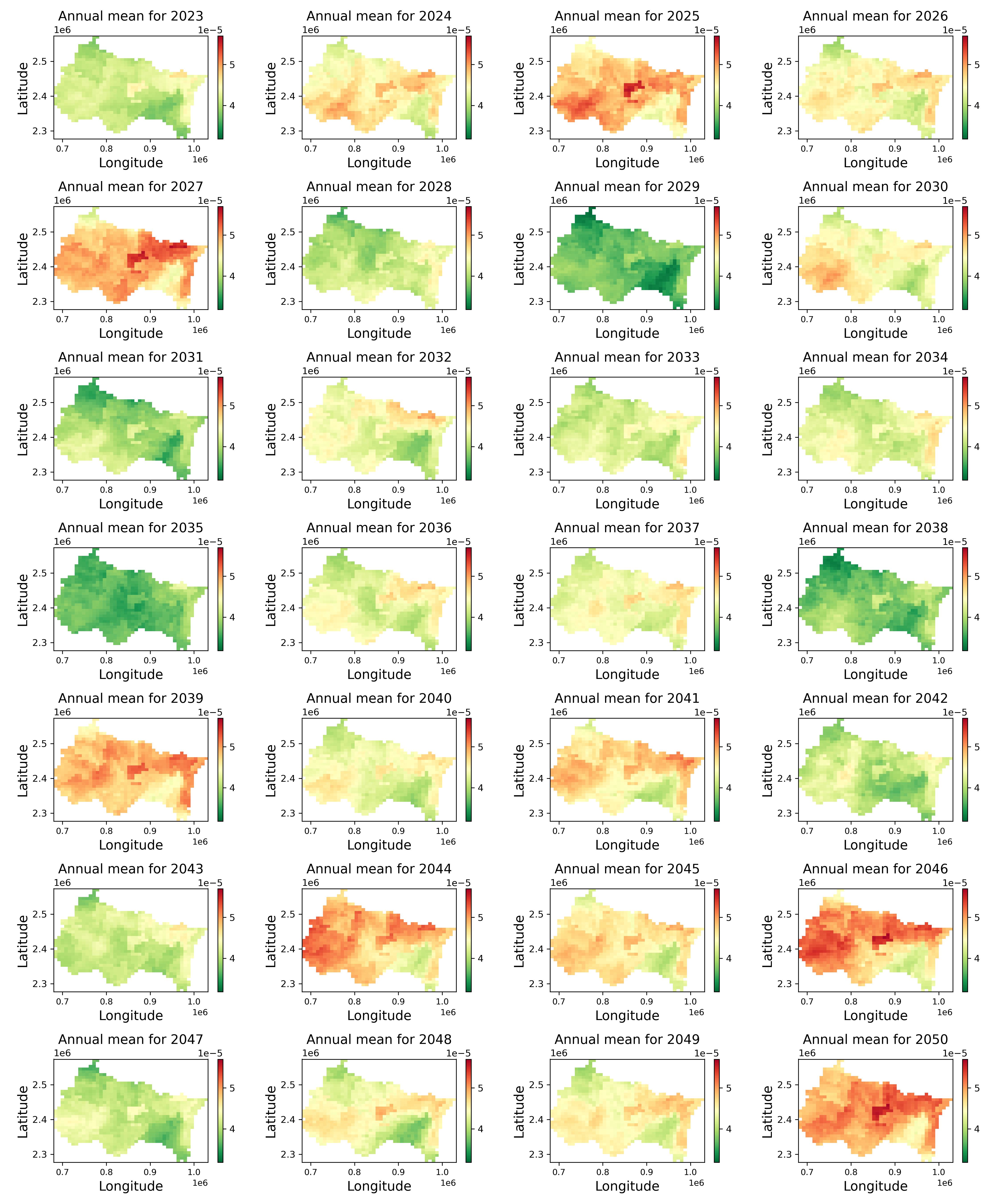}
    \caption{Evolution of mean annual evapotranspiration in the Grand Est region under the RCP 8.5 scenario.}
    \label{fig:evapotranspi_rcp85}
\end{figure}

\begin{figure}
    \centering
    \includegraphics[width=1\textwidth]{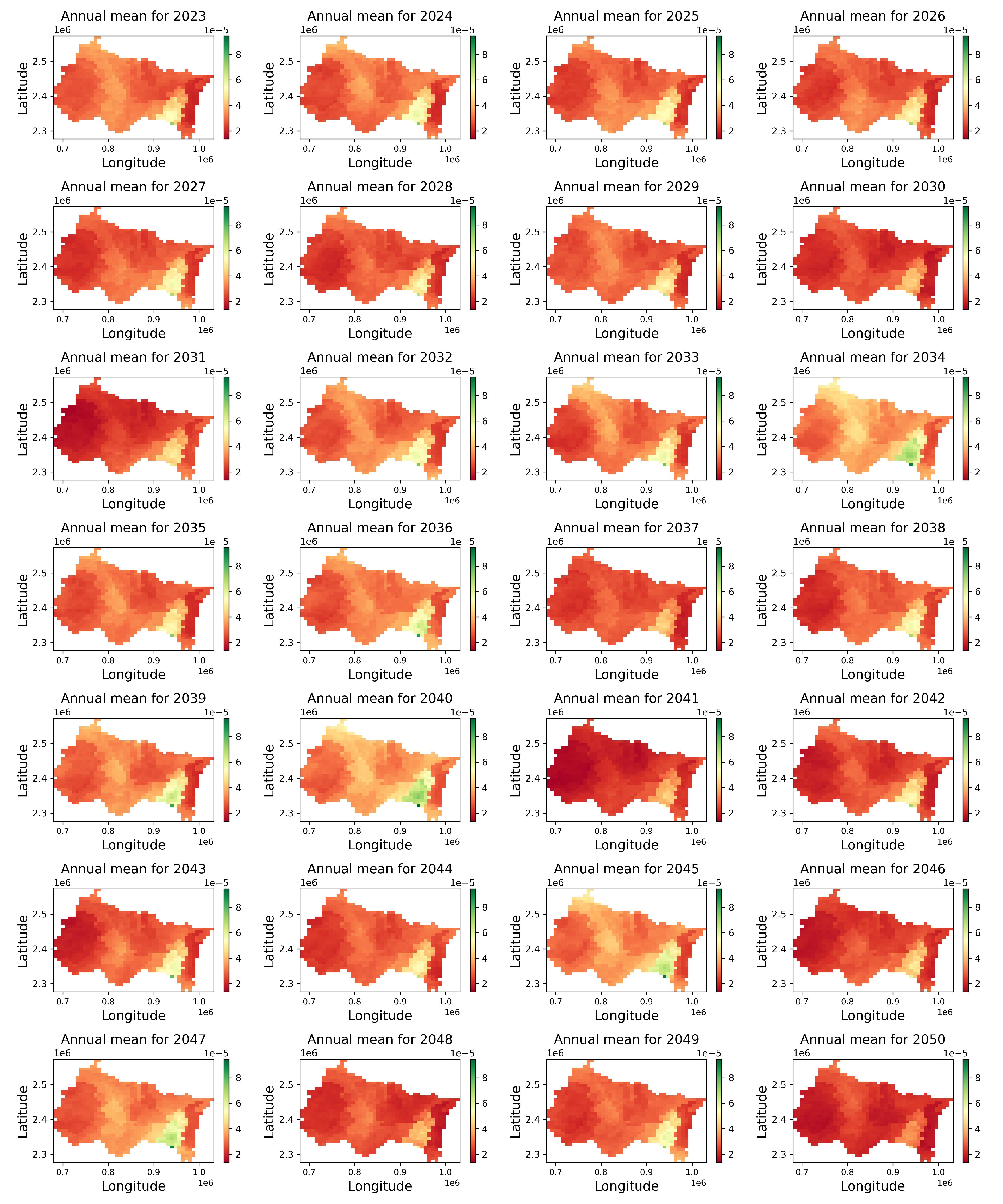}
    \caption{Evolution of mean annual precipitation in the Grand Est region under the RCP 4.5 scenario.}
    \label{fig:precip_rcp45}
\end{figure}

\begin{figure}
    \centering
    \includegraphics[width=1\textwidth]{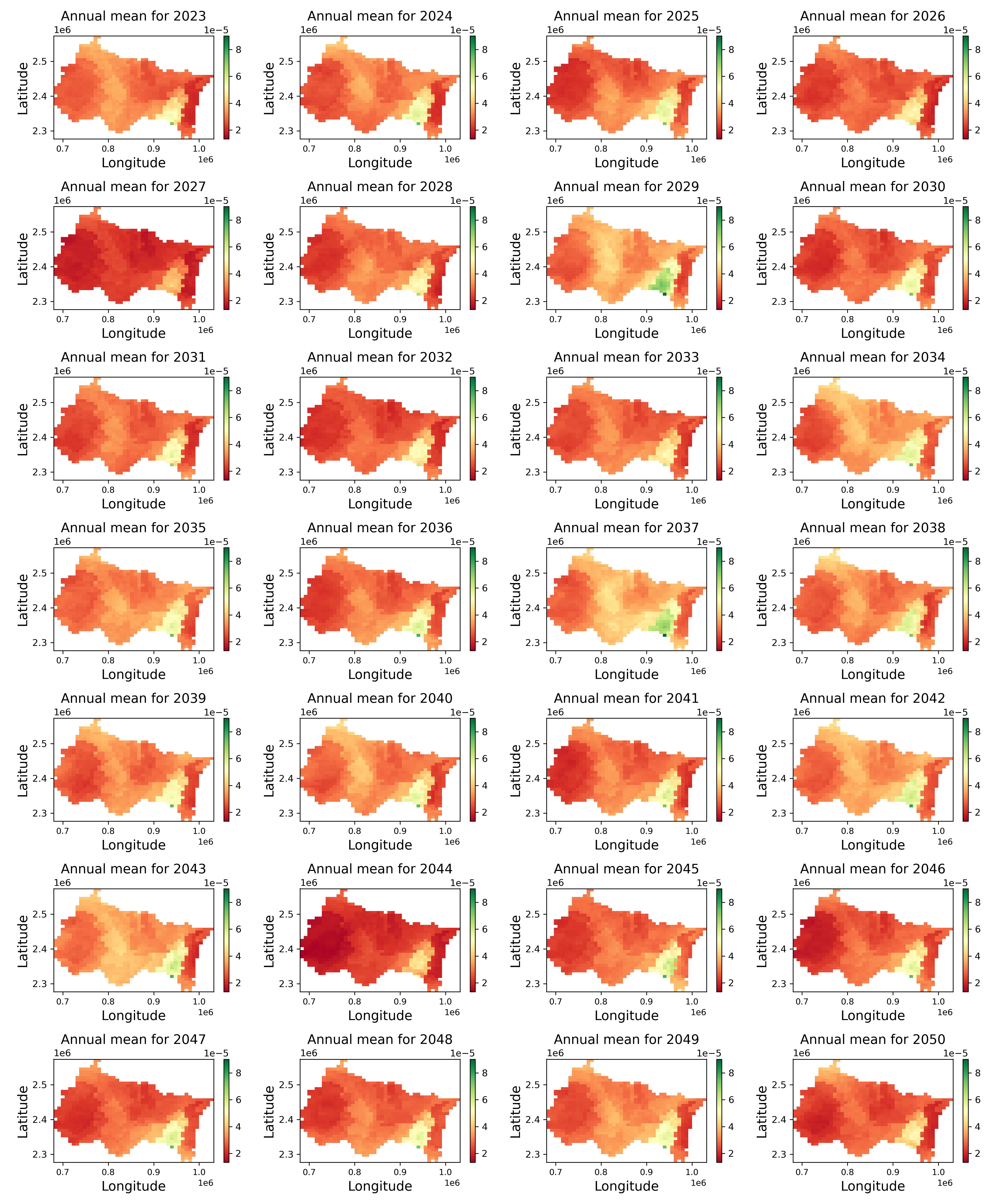}
    \caption{Evolution of mean annual precipitation in the Grand Est region under the RCP 8.5 scenario.}
    \label{fig:precip_rcp85}
\end{figure}

\subsection{Supplementary material on GANs, model training and hyperparameters}

This section is dedicated to providing a detailed description of the architectures of both the generator and the discriminator, as well as the motivations behind each of their components. It also presents the training procedure of the SwiGAN model and the choice of hyperparameters.

\subsubsection{The UNet generator} \label{subsec:sup_generator}

We design our generator following a UNet framework (see Figure \ref{fig:unet_generator}). The latter is composed of a downsampling phase with convolutional layers, also called the encoding phase, followed by an upsampling or decoding phase using transposed convolutions. During the downsampling phase, the resolution of the input maps is gradually reduced by a factor of two at each step and then recovered through the upsampling layers. Details on the architectures of the encoding and decoding phases are given in Tables \ref{tab:encoder_macro_architecture} and \ref{tab:decoder_macro_architecture}, respectively.

\begin{table}[h!]
    \centering
    \resizebox{\textwidth}{!}{\begin{tabular}{@{}l | llccc@{}}
        \toprule
        \textbf{Stage} & \textbf{Block} & \textbf{Layer} & \textbf{Kernel / Stride / Padding} & \textbf{Output shape} & \textbf{Activation} \\ 
        \midrule
        
        \multirow{5}{*}{Stage 1}& \multirow{4}{*}{Resblock 1} & Projection layer & $1\times1$ / 1 / 0             & \multirow{4}{*}{$48\times48\times64$}    & - \\
                                &                           & Conv.            &  $2\times$[$3\times3$ / 1 / 1] &                                           & LeakyReLU \\
                                &                           & scSE Attention Module & -                         &                                           & - \\ 
                                &                           & Residual Add        & -                         &                                           & - \\ 
                                \cline{2-6}
                                & Downsample                 & Conv.                 & $2\times2$ / 2 / 0          &         $24\times24\times64$            & LeakyReLU \\ 
        \midrule

        \multirow{5}{*}{Stage 2}& \multirow{4}{*}{Resblock 2} & Projection layer & $1\times1$ / 1 / 0             & \multirow{4}{*}{$24\times24\times128$}    & - \\
                                &                           & Conv.            &  $2\times$[$3\times3$ / 1 / 1] &                                           & LeakyReLU \\
                                &                           & scSE Attention Module & -                         &                                           & - \\ 
                                &                           & Residual Add        & -                         &                                           & - \\ 
                                \cline{2-6}
                                & Downsample                 & Conv.                 & $2\times2$ / 2 / 0           &         $12\times12\times128$            & LeakyReLU \\ 
        \midrule

        \multirow{5}{*}{Stage 3}& \multirow{4}{*}{Resblock 3} & Projection layer & $1\times1$ / 1 / 0             & \multirow{4}{*}{$12\times12\times256$}    & - \\
                                &                           & Conv.            &  $2\times$[$3\times3$ / 1 / 1] &                                           & LeakyReLU \\
                                &                           & scSE Attention Module & -                         &                                           & - \\ 
                                &                           & Residual Add        & -                         &                                           & - \\ 
                                \cline{2-6}
                                & Downsample                 & Conv.                 & $2\times2$ / 2 / 0           &         $6\times6\times256$            & LeakyReLU \\ 
        \midrule

        \multirow{5}{*}{Stage 4}& \multirow{4}{*}{Resblock 4} & Projection layer & $1\times1$ / 1 / 0             & \multirow{4}{*}{$6\times6\times256$}    & - \\
                                &                           & Conv.            &  $2\times$[$3\times3$ / 1 / 1] &                                           & LeakyReLU \\
                                &                           & scSE Attention Module & -                         &                                           & - \\ 
                                &                           & Residual Add        & -                         &                                           & - \\ 
                                \cline{2-6}
                                & Downsample                 & Conv.                 & $2\times2$ / 2 / 0           &         $3\times3\times256$            & LeakyReLU \\ 
        \midrule

        \multirow{5}{*}{Stage 5}& \multirow{4}{*}{Resblock 5} & Projection layer & $1\times1$ / 1 / 0             & \multirow{4}{*}{$3\times3\times256$}    & - \\
                                &                           & Conv.            &  $2\times$[$3\times3$ / 1 / 1] &                                           & LeakyReLU \\
                                &                           & scSE Attention Module & -                         &                                           & - \\ 
                                &                           & Residual Add        & -                         &                                           & - \\ 
                                \cline{2-6}
                                & Downsample                 & Conv.                 & $2\times2$ / 2 / 0           &         $1\times1\times256$            & LeakyReLU \\ 
        \midrule
        
        \multirow{2}{*}{CenterBlock} & \multirow{2}{*}{CenterBlock}            & Conv.                 & $1\times1$ / 1 / 0           &         $1\times1\times256$            & LeakyReLU \\   &            & Conv.                 & $1\times1$ / 1 / 0           &         $1\times1\times256$            & -\\ 
        \bottomrule
    \end{tabular}}
    \caption{Macro-architecture of the encoding phase. All convolutional layers are followed by a dropout layer and a batch normalization layer.}
    \label{tab:encoder_macro_architecture}
\end{table}

\begin{table}[h!]
    \centering
    \resizebox{\textwidth}{!}{\begin{tabular}{@{}l | llccc@{}}
        \toprule
        \textbf{Stage} & \textbf{Block} & \textbf{Layer} & \textbf{Kernel / Stride / Padding / Filters} & \textbf{Output shape} & \textbf{Activation} \\ 
        \midrule
        \multirow{7}{*}{Stage 1} & \multirow{2}{*}{Upscaling Block} & Transposed Conv.                 & $2\times2$ / 2 / 0 / 256&         $2\times2\times256$            & - \\
                                &                             & Zero Padding                 & -                               &         $3\times3\times256$            & - \\
                                \cline{2-6}
        
                                & \multirow{6}{*}{Resblock 1} & Transposed Conv.            & $2\times2$ / 2 / 0 / 256            & $2\times2\times256$   & LeakyReLU \\
                                &                             & Zero Padding                 & -                              & $3\times3\times256$  & - \\
                                &                           & Conv.                          & $2\times$[$3\times3$ / 1 / 1 / 256]  & $3\times3\times256$  & LeakyReLU \\
                                &                           & scSE Attention Module & -                         &                                           & - \\ 
                                &                           & Residual Add          & -                         &   $3\times3\times256$                     & - \\ 
        \midrule

        \multirow{7}{*}{Stage 2} & Upscaling Block & Transposed Conv.                 & $2\times2$ / 2 / 0 / 256         &         $6\times6\times256$            & - \\
                                \cline{2-6}
        
                                & \multirow{4}{*}{Resblock 2} & Transposed Conv.            & $2\times2$ / 2 / 0 / 256            & \multirow{4}{*}{$6\times6\times256$}   & LeakyReLU \\
                                &                           & Conv.                         & $2\times$[$3\times3$ / 1 / 1 / 256]  &                                         & LeakyReLU \\
                                &                           & scSE Attention Module         & -                              &                                                       & - \\ 
                                &                           & Residual Add                  & -                              &                                                        & - \\ 
        \midrule

        \multirow{7}{*}{Stage 3} & Upscaling Block & Transposed Conv.                 & $2\times2$ / 2 / 0 / 256          &         $12\times12\times256$            & - \\
                                \cline{2-6}
        
                                & \multirow{4}{*}{Resblock 3} & Transposed Conv.            & $2\times2$ / 2 / 0 / 256            & \multirow{4}{*}{$12\times12\times256$}   & LeakyReLU \\
                                &                           & Conv.                         & $2\times$[$3\times3$ / 1 / 1 / 256]  &                                         & LeakyReLU \\
                                &                           & scSE Attention Module         & -                              &                                                       & - \\ 
                                &                           & Residual Add                  & -                              &                                                        & - \\ 
        \midrule

        \multirow{7}{*}{Stage 4} & Upscaling Block & Transposed Conv.                 & $2\times2$ / 2 / 0 / 128         &         $24\times24\times128$            & - \\
                                \cline{2-6}
        
                                & \multirow{4}{*}{Resblock 4} & Transposed Conv.            & $2\times2$ / 2 / 0 / 128          & \multirow{4}{*}{$24\times24\times128$}   & LeakyReLU \\
                                &                           & Conv.                         & $2\times$[$3\times3$ / 1 / 1 / 128]  &                                         & LeakyReLU \\
                                &                           & scSE Attention Module         & -                              &                                                       & - \\ 
                                &                           & Residual Add                  & -                              &                                                        & - \\ 
        \midrule

        \multirow{7}{*}{Stage 5} & Upscaling Block & Transposed Conv.                 & $2\times2$ / 2 / 0 / 64        &         $48\times48\times64$            & - \\
                                \cline{2-6}
        
                                & \multirow{4}{*}{Resblock 5} & Transposed Conv.            & $2\times2$ / 2 / 0 / 64           & \multirow{4}{*}{$48\times48\times64$}   & LeakyReLU \\
                                &                           & Conv.                         & $2\times$[$3\times3$ / 1 / 1 / 64]  &                                         & LeakyReLU \\
                                &                           & scSE Attention Module         & -                              &                                                       & - \\ 
                                &                           & Residual Add                  & -                              &                                                        & - \\
                                \cline{2-6}
                                & \multirow{2}{*}{Head}     & Conv.                         & $1\times1$ / 1 / 0 / 1         & $48\times48\times1$   & - \\
                                &                           & Center crop                   & -                              & $37\times44\times1$                                       & - \\
        \bottomrule
    \end{tabular}}
    \caption{Macro-architecture of the decoding phase. Each stage $k$ receives the outputs of stage $5 - (k - 1)$ from the encoding phase. The purpose of the zero-padding layer in the first stage is to reestablish spatial dimension consistency between the encoding and decoding phases. All convolutional layers are followed by a dropout layer and a batch normalization layer, except for the head layers.}
    \label{tab:decoder_macro_architecture}
\end{table}

Downsampling allows the model to capture dependencies in the maps and detect patterns at various levels of resolution, while the upsampling phase restores the original resolution. 

Both the downsampling and upsampling blocks comprise six residual blocks (\cite{he2015deepresiduallearningimage}), to which we introduce skip connections between the input of each upsampling block and the output of the corresponding downsampling block at the same resolution. This is inspired by the original UNet model, and the objective is twofold. First, by reusing the encoded feature maps from each residual block during the decoding phase, information is propagated throughout the network with minimal loss, ensuring stable gradient flow similar to that of ResNet models. Second, the details captured by the downsampling block at different resolutions are reused during the upsampling step, allowing the network to recover fine-grained information from the input maps.

\begin{figure}[h!]
    \centering
    \includegraphics[width=1.0\linewidth]{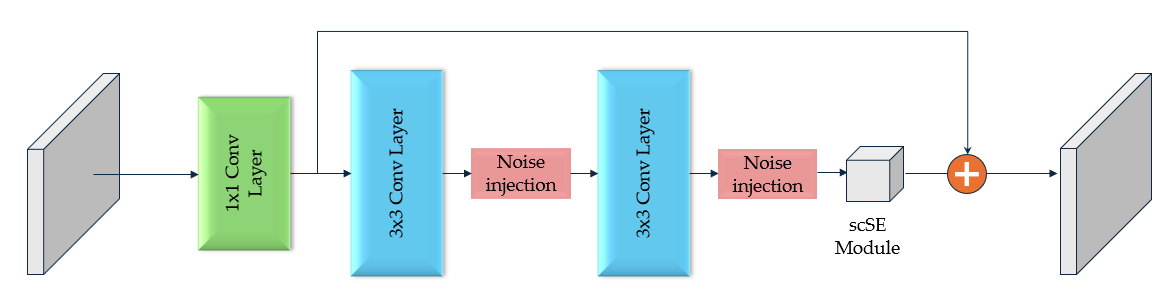}
    \caption{Illustration of a residual block. Noise is introduced at the output of each convolutional layer in the residual blocks using an approach similar to that of StyleGAN2 (see \cite{karras2020analyzingimprovingimagequality}). This helps mitigate overfitting and increases the diversity of the generated maps. Note that, during training, the model can bypass the main branch if it degrades the learning process.}
    \label{fig:residual_block}
\end{figure}

Each residual block (Figure \ref{fig:residual_block}) is composed of two $3\times3$ convolutional layers. Gaussian uncorrelated noise is added to the output of each convolutional block, in a manner similar to \cite{karras2020analyzingimprovingimagequality}. A $1\times1$ convolution without bias is applied to the input as a projection to match the dimensions of the residual block output.

To further reduce the probability of vanishing gradients during training and to improve efficiency and speed, we incorporate several techniques inspired by the literature into our UNet generator:

\begin{itemize}
    \item \textbf{Spatial and channel-wise attention:} Each residual block in both the encoding and decoding phases applies two successive convolutions, followed by a Concurrent Spatial and Channel-wise Squeeze and Excitation (scSE) module (see \cite{roy2018concurrentspatialchannelsqueeze}). This module extends the Squeeze-and-Excitation mechanism introduced by \cite{hu2019squeezeandexcitationnetworks} and enhances the model’s ability to focus on the most relevant features at both the pixel and channel levels. By reducing spatial dimensions to a single value per channel and aggregating channels into a spatial map, the module emphasizes both important locations and feature maps.

    \item \textbf{Stochastic depth:} Stochastic depth (\cite{huang2016deepnetworksstochasticdepth}) is used to facilitate the training of deep networks by randomly dropping residual layers with probability $1 - p_l$. At each training step, a given layer $l$ is applied with probability $p_l$, otherwise its input is directly passed to the next layer. The survival probability $p_l$ decreases linearly from 1 in the first layer to 0.5 in the last layer of the downsampling block, and symmetrically increases in the upsampling block. If $N$ is the total number of residual blocks, then:

    \begin{equation*}
        p_l=1 - \frac{l - 1}{2(N - 1)},\space for \quad l = 1,...,N 
    \end{equation*}

    During inference, all residual blocks are used.

    \item \textbf{Dropout:} As noted by \cite{isola2018imagetoimagetranslationconditionaladversarial}, UNet-based generators may ignore the noise vector and rely solely on input maps, producing similar outputs regardless of noise. To mitigate this issue, dropout layers are introduced in the last three layers of the downsampling phase and the first three layers of the upsampling phase. Applying dropout selectively avoids destabilizing the adversarial training process while promoting output diversity.
\end{itemize}

\subsubsection{The discriminators} \label{subsec:sup_discriminators}
The discriminator network consists of two sub-networks: a frame-level discriminator and a patch-level discriminator, both sharing a common set of layers referred to as the base discriminator (see Figure \ref{fig:patch_gan_discrim}). 

\begin{itemize}
    \item \textbf{The base discriminator:} This component processes the generated maps before passing them to the frame-level and patch-level discriminators. It consists of three convolutional layers, each followed by instance normalization (see \cite{ulyanov2017instancenormalizationmissingingredient}). Its purpose is to reduce computational complexity by applying shared transformations. Its outputs are fed into both discriminators. Table \ref{table:base_discriminator_layers} provides details.
    
    \begin{table}[h!]
    \begin{center}
    \begin{tabular}{ c c c c c c c }
    \hline
     Layer & Kernel & Filters & Stride & Padding & Normalization & Activation \\ 
    \hline
     Conv2D & 3 & 32 & 2 & 1 & InstanceNorm & LeakyReLU(0.2) \\ 
     Conv2D & 3 & 64 & 2 & 1 & InstanceNorm & LeakyReLU(0.2) \\
     Conv2D & 3 & 128 & 2 & 1 & InstanceNorm & LeakyReLU(0.2) \\
     \hline
    \end{tabular}
    \caption{Overview of the layers of the base discriminator}
    \label{table:base_discriminator_layers}
    \end{center}
    \end{table}

    \item \textbf{The frame discriminator:} This network evaluates whether an entire map is realistic. It receives feature maps of size $5 \times 6$ from the base discriminator and outputs a single scalar. Table \ref{table:frame_discriminator_layers} provides details.
    
    \begin{table}[h!]
    \begin{center}
    \begin{tabular}{ c c c c c c c }
    \hline
     Layer & Kernel & Filters & Stride & Padding & Normalization & Activation \\ 
    \hline
     Conv2D & 2 & 256 & 1 & 0 & InstanceNorm & LeakyReLU(0.2) \\ 
     Conv2D & 2 & 256 & 2 & 0 & InstanceNorm & LeakyReLU(0.2) \\
     Conv2D & 2 & 1 & 2 & 1 & None & None \\
     \hline
    \end{tabular}
    \caption{Overview of the layers of the frame discriminator}
    \label{table:frame_discriminator_layers}
    \end{center}
    \end{table}
    
    \item \textbf{The patch discriminator:} Inspired by PatchGAN (\cite{isola2018imagetoimagetranslationconditionaladversarial}), this network evaluates smaller regions of the input maps. It splits feature maps into patches and determines their realism. It receives feature maps of size $5 \times 6$ and outputs $4 \times 5$ values. Table \ref{table:patch_discriminator_layers} provides details.
    
    \begin{table}[h!]
    \begin{center}
    \begin{tabular}{ c c c c c c c }
    \hline
     Layer & Kernel & Filters & Stride & Padding & Normalization & Activation \\ 
    \hline
     Conv2D & 2 & 256 & 1 & 0 & InstanceNorm & LeakyReLU(0.2) \\ 
     Conv2D & 1 & 1 & 1 & 0 & None & None \\
     \hline
    \end{tabular}
    \caption{Overview of the layers of the patch discriminator}
    \label{table:patch_discriminator_layers}
    \end{center}
    \end{table}
\end{itemize}

As shown in Tables \ref{table:base_discriminator_layers}, \ref{table:frame_discriminator_layers}, and \ref{table:patch_discriminator_layers}, each convolutional layer is followed by instance normalization. Unlike batch normalization (\cite{ioffe2015batchnormalizationacceleratingdeep}), which computes statistics across a batch, instance normalization operates on each sample independently by normalizing each feature map across its spatial dimensions. \cite{gulrajani2017improved} show that this choice helps preserve the $1$-Lipschitz constraint discussed in Section \ref{subsec:wgan} while avoiding issues related to batch dependencies in gradient penalty computations.

\subsubsection{Model training} \label{sec:model_training}
As mentioned in Section \ref{subsec:wgan}, one of the main challenges in training GAN architectures is avoiding mode collapse (\cite{durall2020combatingmodecollapsegan}, \cite{thanhtung2020catastrophicforgettingmodecollapse}, \cite{kodali2017convergencestabilitygans}). While Wasserstein GANs help mitigate this issue, they still require careful experimentation and hyperparameter tuning. Another challenge, particularly in low-data regimes as in our case, is discriminator overfitting (\cite{karras2020traininggenerativeadversarialnetworks}, \cite{zhao2020differentiableaugmentationdataefficientgan}). This occurs when the discriminator is exposed to limited variability in the training data, leading it to memorize samples and causing training instability.

Before presenting the training procedure, we introduce two techniques to address overfitting and insufficient variability in generated samples, which are common issues in deep learning and especially in GAN models:

\begin{itemize}
    \item \textbf{Enhancing the discriminators:} In deep learning, data augmentation is a standard approach to mitigate overfitting (see \cite{perez2017effectivenessdataaugmentationimage, wang2025comprehensivesurveydataaugmentation}). It consists of applying random transformations such as rotation, shearing, mirroring, or degradations such as Gaussian blurring, CutMix (\cite{yun2019cutmixregularizationstrategytrain}), and Mixup (\cite{zhang2018mixupempiricalriskminimization}) to the input images. However, \cite{zhang2020consistencyregularizationgenerativeadversarial} and \cite{zhao2020improvedconsistencyregularizationgans} show that such augmentations may lead the generator to learn the distribution of augmented samples rather than the true data distribution.

    To address this issue, we apply the differentiable augmentation method (DiffAugment) proposed by \cite{zhao2020differentiableaugmentationdataefficientgan}. This approach applies differentiable transformations to both real and generated maps before feeding them to the discriminator. To ensure proper gradient propagation through the generator, these transformations must be differentiable. In our implementation, we use a combination of random translations and cutouts.

    We also employ spectral normalization (\cite{miyato2018spectralnormalizationgenerativeadversarial}) to stabilize gradients during training by normalizing weight matrices by their largest singular value, further reducing the risk of mode collapse.

    \item \textbf{Increasing variability in generated maps:} To enhance diversity in generated outputs, we introduce Gaussian noise after each activation during both training and inference. Specifically, for each residual block $k$, we generate a 2D noise matrix of size $(h_k, w_k)$ with normally distributed values, where $h_k$ and $w_k$ denote the spatial dimensions of the activation map. This noise is added independently to each channel after being scaled by a learnable coefficient. This approach is inspired by \cite{karras2019stylebasedgeneratorarchitecturegenerative} and \cite{karras2020analyzingimprovingimagequality}.
\end{itemize}

For the training procedure, we adopt the Wasserstein GAN framework with gradient penalty (see \cite{gulrajani2017improved}), with a penalty coefficient $\lambda_{pen}=10$. Let $\mathcal{L}_{adv}$ denote the adversarial loss described in Equation \ref{eq:adv_loss_wgan_gp}. Following \cite{isola2018imagetoimagetranslationconditionaladversarial} and \cite{foroumandi2024generative}, we incorporate a pixel reconstruction loss $\mathcal{L}_{rec}$ to stabilize training and guide the generator. We use the pixel-wise Mean Absolute Error (MAE) and set $\lambda_{rec}=100$. Higher values of $\lambda_{rec}$ emphasize prediction accuracy, while lower values favor adversarial learning.

To further stabilize the training process, we introduce feature matching (see \cite{salimans2016improvedtechniquestraininggans}). Let $v(x)$ denote the output of an intermediate discriminator layer for a real sample $x$, and $v(G(Z))$ the corresponding output for a generated sample. The feature matching loss is defined as:
$$\mathcal{L}_{feat} = \| E_{x \sim \mathbb{P}_{data}} v(x) - E_{Z \sim \mathbb{P}_Z} v(G(Z)) \|^2_2$$

We set $\lambda_{feat} = 10$. The final objective function is:
$$
\mathcal{L} = \mathcal{L}_{adv} + \lambda_{rec}\mathcal{L}_{rec} + \lambda_{feat}\mathcal{L}_{feat}
$$

The model is trained with a batch size of 64 and a 32-dimensional noise vector $Z$ for 1500 epochs.

Our covariates include a categorical variable representing the timestep $T$ at which generation is performed. We encode this variable using a lookup table mapping timestep indices to trainable 5-dimensional vectors. These vectors are duplicated and appended to the input maps as additional channels. The lag parameter is set to $u=8$, as increasing the number of lags did not significantly improve performance.

We use the AdamW optimizer with a learning rate of $10^{-5}$, a weight decay of 0.1, and $(\beta_1, \beta_2) = (0.5, 0.999)$. For each generator update, the discriminator is updated five times. A cosine annealing schedule is used for the learning rate. All hyperparameters were selected based on fine-tuning on the training and validation sets. Additionally, all linear and convolutional layers are initialized using the Glorot initialization scheme (see \cite{glorotinitialization}), which significantly improves training stability and performance.

\subsection{Supplementary material on variable importance}\label{subsec:var_importance_sup}

As highlighted by many authors in deep learning (\cite{lungnodules}, \cite{dosovitskiy2021imageworth16x16words}, \cite{pedersoli2017areasattentionimagecaptioning}), understanding which features influence model outputs is crucial for validating results and deriving meaningful insights.

Methods such as Shapley additive explanations quantify the contribution of each feature (or channel in image-based models) to the model output (see \cite{lundberg2017unifiedapproachinterpretingmodel}). The Shapley additive explanations method (SHAP method) is grounded in cooperative game theory and relies on Shapley values to assign contributions to each feature.

The SHAP method represents the model output as an additive function:
$$g(x') = \phi_0 + \sum_{i=1}^{M} \phi_i x'_i$$
where:
\begin{itemize}
    \item $M$ is the number of input features,
    \item $x' \in \{0,1\}^M$ is the coalition vector indicating feature presence,
    \item $\phi_i$ is the Shapley value representing the contribution of feature $i$,
    \item $\phi_0$ is the baseline value (average model output).
\end{itemize}

We estimate Shapley values using the permutation method, also known as permutation feature importance. This method evaluates the increase in prediction error after randomly permuting a feature’s values, thereby breaking its relationship with the target variable. Features with the largest absolute Shapley values are considered the most influential.

\bibliographystyle{abbrvnat}

\end{document}